# An Integrated Remote Sensing and Machine Learning Framework for Mapping Hazard Associated with Leached Metals

Salman Karim[1*], Bahareh Nojabaei [1], Hosein Foroutan[2]

{salmankarim, baharehn, hosein}@vt.edu

[1] Department of Mining & Minerals Engineering, Virginia Tech, Blacksburg, VA, USA.
[2] Department of Civil & Environmental Engineering, Virginia Tech, Blacksburg, VA, USA

## Abstract

This study presents an integrated remote sensing and machine learning framework for characterizing vegetation–environment patterns and leached-metal hazard at four coal combustion residual (CCR) sites in Virginia, USA. Groundwater concentrations were standardized against constituent-specific reference thresholds as Hazard Quotients (HQ) and summed into a composite Hazard Index (HI) representing the cumulative relative metal burden. The predictors were multi-temporal Sentinel-2 observations (2019–2025), including vegetation indices, biophysical parameters and spatio-temporal vegetation metrics (CSTV and rate of change), together with soil, climatic, hydrological, topographic and land-cover variables. Six supervised classifiers were evaluated on spatially separated test blocks: Classification and Regression Tree (CART), Random Forest (RF), Gradient Tree Boosting (GTB), Support Vector Machine (SVM), k-Nearest Neighbor (KNN) and Naïve Bayes (NB). The evaluation was supplemented by feature-set ablation and leave-one-site-out testing. Near the monitoring wells, higher metal HQ coincided with lower vegetation index values in 24 of 33 metal–site combinations (sign test $p = 0.007$). Well HI was negatively correlated with CIRE and CCC at three of the four sites ($\rho$ down to $-0.71$) and water and built surfaces near the wells masked the response at the fourth site. For composite HI classification within sites, RF reached accuracies of 0.768–0.924. Recall for the higher-HI class was 0.965–0.999. At the constituent level, RF classified the normalized HQ of 34 metal–site combinations with accuracies of 0.775–0.989. Feature-set ablation showed that the soil, climatic, topographic and land-cover covariates carried most of the predictive information. Models without vegetation predictors performed as well as or better than the full model, and at one site the distance to the CCR unit alone outperformed all remotely sensed predictor sets. Leave-one-site-out zero-shot transfer testing confirmed strong site dependence: RF accuracy fell to 0.734–0.749, no better than predicting the majority class. Overall, the proposed framework demonstrates the potential of integrating groundwater-derived HQ–HI information with satellite and environmental data to extend spatial screening beyond discrete monitoring wells. Rather than directly detecting dissolved metals, the framework identifies remotely observable vegetation–environment patterns associated with constituent-specific HQ and cumulative HI, providing a scalable approach to complement targeted sampling, long-term CCR surveillance, and remediation monitoring.

## 1 Introduction

The global reliance on coal-fired power has resulted in an immense environmental liability: millions of tons of coal combustion residuals (CCR) disposed of or stored in surface impoundments and landfills (Ruhl et al. 2012; Harkness et al. 2016). CCR contain elevated concentrations of organic pollutants and heavy metals/metalloids (HMs) which can be mobilized under variable hydrological and geochemical conditions and contaminate surrounding environment posing chronic risks to human health, ecological and aquatic systems (Dreesen et al. 1977; Meij and te Winkel 2007; Tchounwou et al. 2012; Izquierdo and Querol 2012; Swanson et al. 2013; Lauer et al. 2015; Wang et al. 2022). Studies at U.S. coal-ash impoundments and landfills have documented CCR leachates with metal concentrations frequently exceeding federal advisory guidelines (Rowe et al. 2002; Ruhl et al. 2010, 2012; Yang et al. 2015; Vengosh et al. 2019; Deonarine et al. 2023). Despite requiring lined landfills and leachate collection systems intended to ensure environmental safety, most of the U.S. coal-fired power plants show unsafe levels of HMs in monitoring wells, largely due to decades of waste stored in older unlined landfills (US EPA 2015; Russ et al. 2019). HMs contamination from CCRs generally progresses gradually through continuous leaching from unlined or unremedied impoundments, and because these metals are non-degradable, they persist in groundwater and soil for long periods after disposal (Adriano et al. 2004; Bolan et al. 2014; Brandt et al. 2017; Petrović and Fiket 2022; Padhye et al. 2023).

The established regulatory framework for monitoring CCR sites is based on a strategy of direct, point-based sampling which relies on a network of monitoring wells to detect and characterize contaminant releases (US EPA 2015). While these methods provide high analytical precision, they are inherently constrained by their limited spatial and temporal resolution, capturing only discrete locations and infrequent sampling intervals (Ling et al. 2003). The costs and labor associated with installing, maintaining, and regularly sampling a comprehensive network of wells are substantial. These limitations underscore the need for complementary cost-effective, scalable, and non-invasive methods capable of providing continuous environmental surveillance over large and often inaccessible areas, to better guide regulatory oversight, risk assessment, and remediation prioritization.

Vegetation serves as a powerful bioindicator for subsurface contamination, as it integrates and expresses biophysical and biochemical responses to environmental stressors (Adriano et al. 2004; Gholizadeh and Kopačková 2019). The uptake of heavy metals from contaminated soil and water interferes with critical physiological processes, including photosynthesis, pigment synthesis, and water transport and this interference results in observable stress responses such as chlorosis, reduced chlorophyll content, and structural damage to the canopy, all of which alter the plant's spectral reflectance properties in the visible, red-edge, and near-infrared (NIR) portions of the electromagnetic spectrum (Rout and Das 2009; Sperdouli 2022; Bonfiglio et al. 2024; Lovynska et al. 2024).

These spectral changes can be quantified using a variety of vegetation indices derived from satellite imagery (Huete 1988; Dash and Curran 2004; Mishra and Mishra 2012; D'Emilio et al. 2012; Li et al. 2015; Karim et al. 2026). In parallel, biophysical parameters such as Leaf Area Index (LAI) and Leaf Chlorophyll Content (LCC), often derived through radiative transfer modeling, provide direct metrics of canopy structure and physiological status.

The Sentinel-2 mission, with 13 spectral bands, 10–20 m spatial resolution and frequent revisit times, is well-suited for detecting subtle signals of vegetation stress (Zarco-Tejada et al. 2019; Shukla et al. 2019; Soydan et al. 2021; Plutalova et al. 2021; Yu et al. 2023a; Guo et al. 2024). Its strategically placed red-edge bands (centered around 705 nm, 740 nm, and 783 nm) are highly sensitive to variations in chlorophyll concentration and can reveal stress gradients that may be invisible in other spectral regions (Clevers et al. 2004; Eitel et al. 2011; Clevers and Gitelson 2013). Numerous studies have successfully demonstrated correlations between heavy metal concentrations in soil and vegetation spectral responses in mining and industrial landscapes (Götze et al. 2016; Jin et al. 2017; Zhang et al. 2018, 2019; Liu et al. 2018; Yang et al. 2022; Noor et al. 2022). However, these relationships are frequently site-specific and nonlinear, reflecting the complex interplay of factors like soil texture, moisture, topography, and climate that co-determine plant stress.

Machine learning (ML) provides a powerful, data-driven framework for modeling the complex, nonlinear relationships between environmental variables and vegetation health (Gao et al. 2020; van Klompenburg et al. 2020; Zhang et al. 2022; Wang et al. 2025). Supervised algorithms, including Random Forest (RF), Support Vector Machines (SVM), Gradient Boosting have been successfully applied to map heavy metal contamination in soils and sediments by integrating spectral information with environmental covariates to characterize soil and vegetation responses to toxic metals (Jerome H. Friedman 2001; Rumpf et al. 2010; Mountrakis et al. 2011; Rodriguez-Galiano et al. 2012; Belgiu and Drăguţ 2016; Thanh Noi and Kappas 2017; Tan et al. 2020; Fu et al. 2024). Moreover, Li et al. (2025) combined multi-source remote sensing (Landsat 8, Sentinel-1/2), terrain factors, and ensemble ML to develop multi-target inversion models for six metals (Cr, Zn, As, Cd, Hg, Pb) in both soils and crops. Yu et al. (2023) similarly integrated Sentinel-2A imagery with spectral, topographic, and spatial features to map Pb and Cd using ensemble learning. At the watershed scale, Wang et al. (2020) modeled ecological risk by linking NDVI and environmental drivers, while Wang et al. (2023) fused ground and airborne hyperspectral data with SVM and deep neural networks to predict soil heavy-metal concentrations. In agricultural landscapes, Zhang et al. (2022) demonstrated that Sentinel-2 time-series features especially red-edge indices combined with deep learning model can detect heavy-metal stress in rice. At finer scales, Salgado et al. (2023) used UAV multispectral imagery and random forest model to map Hg and As in soil–plant systems, showing substantial improvements over conventional assessment approaches. Together, these studies reflect a rapidly expanding body of work showing that combining remote sensing with advanced data-driven ML models enables reliable quantification and monitoring of heavy-metal contamination across diverse environmental contexts.

Critical knowledge gaps persist in linking vegetation stress, multi-year temporal dynamics, and subsurface leached-metal contamination across heterogeneous landscapes. These include the lack of reproducible multi-site frameworks, limited integration of environmental drivers such as soil, hydrology, topography and climate, and insufficient evaluation of temporal consistency needed to separate persistent contamination from transient environmental variability.

To address these challenges, this study proposes a comprehensive and reproducible supervised ML framework for characterizing and spatially mapping vegetation stress and environmental hazard patterns associated with CCR-leachate metals. The framework integrates groundwater monitoring observations of constituents mobilized through CCR leaching with multi-source geospatial information, including multi-temporal Sentinel-2 spectral bands, vegetation indices, biophysical parameters, spatio-temporal vegetation metrics, and environmental variables representing soil, climate, hydrology, topography, and land cover. Concentrations of individual metals are standardized relative to constituent-specific reference thresholds using Hazard Quotients (HQ) and subsequently integrated into a composite Hazard Index (HI), providing spatial reference information for supervised classification of vegetation stress conditions associated with lower and higher cumulative relative metal burden.

The study makes several methodological contributions. First, the proposed HQ–HI framework provides a unified representation of multiple co-occurring CCR-leachate-associated metals, rather than relying on a single constituent as the site-level classification reference. Second, the framework integrates CSTV and multi-year rate-of-change metrics with spectral, vegetation-index, biophysical, and environmental predictors to characterize both spatial and temporal variations in vegetation conditions. Third, multiple supervised ML algorithms are systematically evaluated to identify an effective classifier for mapping sitewide HI-associated patterns. The selected RF model is further evaluated at the individual-constituent level using normalized HQ to determine whether the framework can distinguish vegetation–environment patterns associated with individual metal hazard distributions in addition to the composite multi-constituent condition. Feature-importance analysis is subsequently used to evaluate the relative predictive contributions of different remote-sensing and environmental predictor domains. Finally, the framework evaluates geographic transferability through leave-one-site-out testing to determine how effectively relationships learned across CCR sites transfer to an unseen site characterized by different environmental and geochemical conditions. The proposed supervised ML framework connects discrete groundwater observations of CCR-leachate-associated constituents with spatially continuous earth observation and environmental information, extending spatial screening beyond monitoring-well locations. The framework does not represent direct satellite detection of dissolved metals; rather, it identifies remotely observable vegetation responses and environment patterns spatially associated with constituent-specific HQ and cumulative HI, providing a basis for targeted sampling, long-term CCR-site surveillance, and remediation monitoring. Collectively, this work establishes a scalable, low-cost, and operationally reproducible framework for large-scale detection of vegetation stress and spatial spread of hazard associated with metal contamination, providing valuable insights for groundwater monitoring, ecological risk assessment, and regulatory decision-making in legacy coal-ash landscapes.

## 2 Study Sites

This study focuses on four coal-fired power plants in Virginia, USA, that contain legacy CCR impoundments and landfills with documented histories of soil & groundwater contamination from leaching metals as shown in Figure 1. These sites represent diverse geomorphic, hydrological, and ecological settings, providing a robust basis for evaluating vegetation stress responses to leached-metal contamination and the transferability and effectiveness of the proposed framework.

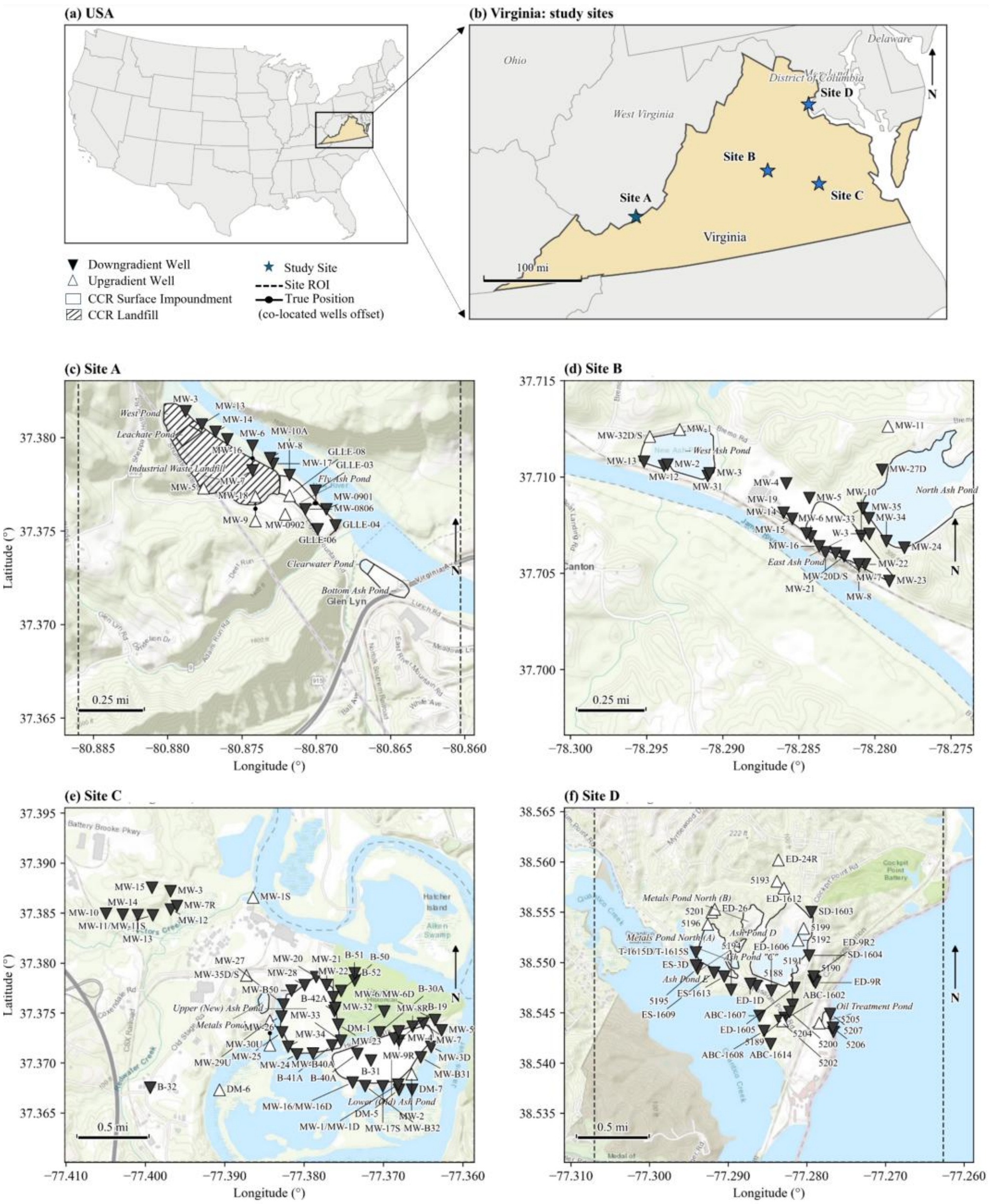

Figure 1: Location of the four CCR study sites in Virginia, USA. The top-left panel (a) shows the position of Virginia within the continental USA, while the top-right panel (b) highlights the distribution of the four power-station sites across the state. At bottom: each site with delineated CCR impoundments and landfills with up and downgradient monitoring wells at (c) Site A: Glen Lyn Power Station, (d)Site B: Bremo Power Station, (e) Site C: Chesterfield Power Station, and (f) Site D: Possum Point Power Station.

Site characteristics in Table 1 were compiled from publicly available regulatory and environmental datasets, including the Ashtracker database maintained by the Environmental Integrity Project (EIP 2024) which provides facility-specific information and documents groundwater monitoring infrastructure (number of wells), the subset of wells exceeding federal drinking-water standards, and the suite of monitored metals. Sampling years (2010–2019) reflect the periods during which groundwater quality assessments were performed under federal CCR regulations.

Table 1: Key characteristics of the four CCR sites analyzed in this study.

| Characteristic | Site A | Site B | Site C | Site D |
|---|---|---|---|---|
| Name | Glen Lyn Power Station | Bremo Power Station | Chesterfield Power Station | Possum Point Power Station |
| Location | Glen Lyn, VA | Bremo Bluff, VA | Chester, VA | Dumfries, VA |
| Activity status | Retired | Active | Active | Active |
| EPA Hazard rating | Low | Significant | Significant | Significant |
| Plant capacity (MW) | 338 | 254 | 1,353 | 1,235 |
| Year built | 1919 | 1951 | 1952 | 1956 |
| No. of CCR Unit | 6 | 3 | 4 | 5 |
| Total known area (acres) | 13 | 113 | 161 | 152 |
| Monitoring wells | 19 | 32 | 59 | 36 |
| Polluted wells | 17 | 30 | 51 | 18 |
| Monitored metals and metalloids | As, B, Co, Pb, Mn, Mo, Sr, Sb, Ba, Be, Cd, Cr, Cu, Hg, Ni, Se, Ag, Tl, U | As, B, Co, Fe, Pb, Li, Mn, Mo, Ra, Tl, Sb, Ba, Be, Cd, Cr, Cu, Hg, Se, Ag, Zn | As, Be, B, Cr, Co, Pb, Li, Mn, Mo, Ra, Sb, Ba, Cd, Cu, Hg, Se, Zn | As, B, Cd, Co, Mn, Ni, Ra, Mo, Sb, Ba, Be, Cr, Cu, Pb, Hg, Se, Ag, Tl |
| Sampling period | 2010–2014 | 2013–2019 | 2010–2019 | 2010–2019 |

# 3 Methods

## 3.1 Framework Overview

A comprehensive and reproducible framework (Figure 2) was developed to detect and map leached metal-induced vegetation stress patterns and CCR-leachate associated contamination hazard by integrating multi-source geospatial datasets with supervised machine-learning models.

The workflow begins with (1) multi-source data collection through Google Earth Engine, including Sentinel-2 surface-reflectance imagery, Dynamic World land-cover data, environmental covariates (soil, topography, hydrology, and climate), and groundwater monitoring-well leachate concentrations; (2) Data preprocessing includes cloud masking and quality filtering, spatial and temporal filtering, and derivation of relevant biophysical and spatiotemporal variables; (3) Feature engineering integrates Sentinel-2 spectral bands, derived vegetation and biophysical indices (VI & BPI), spatiotemporal variation metrics including CSTV, rate of change, and seasonal composites, environmental covariates, and land-cover probabilities; (4) Target feature labeling is performed by calculating a Hazard Index (HI) from monitoring-well measurements, spatially interpolating HI using ordinary kriging to generate a continuous HI surface, normalizing the interpolated HI to a 0–1 scale, and converting it into binary relatively higher-HI and lower-HI classes using a predefined reference threshold; (5) Feature stacking then combines the predictor variables into a unified multiband feature dataset. Pixel-level predictor and target values are subsequently extracted with (6) spatial block partitioning of 32 × 32 pixel blocks into training (70%), validation (15%) and test (15%) sets with a 8 pixel buffer between the partitions; (7) Multiple supervised machine learning classification models including Classification and Regression Tree (CART), Random Forest (RF), Gradient Tree Boosting (GTB), Support Vector

Machine (SVM), k-Nearest Neighbor (KNN), and Naïve Bayes (NB) are trained; (8) The trained models generate spatially explicit classified relatively higher-HI and lower-HI maps, which are (9) evaluated against reference labels using accuracy, Cohen's kappa, precision, recall, and F1-score and quantification of predictor-group contributions with impurity, group-permutation and SHAP importance is conducted; (10) Model selection identifies the best-performing model based on these evaluation results. The selected model can then undergo (11) optional cross-site transfer, in which the trained model is deployed at an independent site to assess spatial transferability. Finally, (12) mapping and visualization produce HI, classification, probability, and prediction-error maps, while (13) reporting and documentation support interpretation and communication of the results. Overall, the framework provides a scalable and reproduceable approach for screening and mapping vegetation-environment stress patterns and associated spatial contamination hazard patterns of CCR-derived leachate metals/metalloids across heterogeneous landscapes.

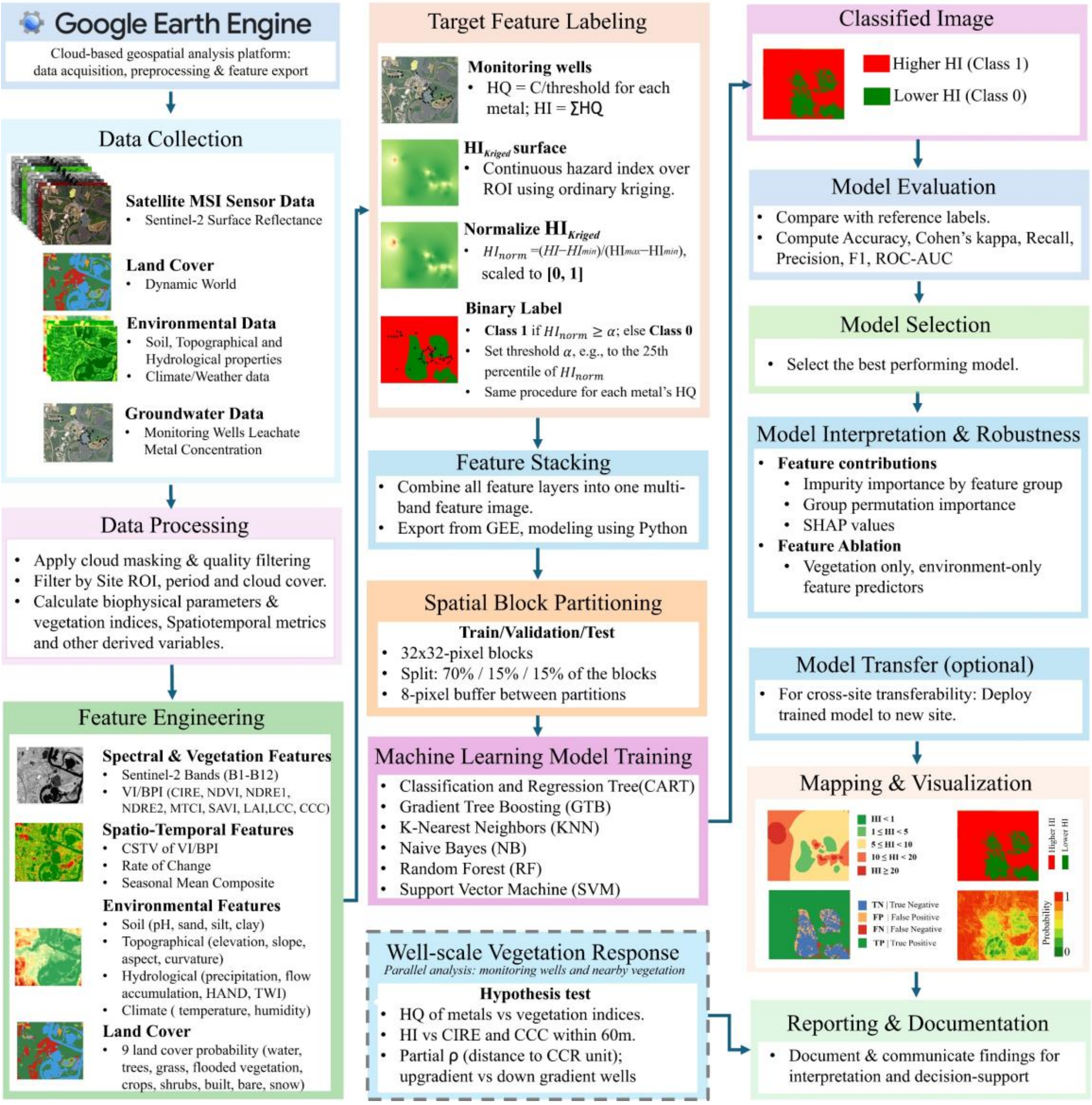

Figure 2: Simplified workflow of the proposed integrated remote sensing and machine learning driven framework for mapping CCR-associated leached metal hazard.

## 3.2 Data Collection

Multi-source datasets were acquired through Google Earth Engine (GEE) geospatial platform (Gorelick et al. 2017) to represent vegetation, soil, hydrological, and climatic conditions over the four CCR sites.

i. **Spectral Data:** Multispectral imagery was acquired from the Sentinel-2 Multispectral Instrument (MSI) constellation. We utilized Level-2A surface reflectance products, which are atmospherically corrected and provide data across 12 spectral bands at 10–20-meter spatial resolution with a nominal 5-day revisit cycle (ESA 2022).
ii. **Soil Data:** Key soil physicochemical properties were sourced from the SoilGrids v2.0 global soil information system. This dataset, produced by ISRIC–World Soil Information, provides global predictions for properties such as pH, texture fractions (sand, silt, and clay) at a 250-meter resolution (Turek et al. 2023). And soil moisture dataset is produced by NASA Global Land Data Assimilation System (Rodell et al. 2004)
iii. **Topographic Data:** Digital elevation data were obtained from the Shuttle Radar Topography Mission (SRTM), which provides near-global, high-resolution topographic information. The specific dataset used is the void-filled SRTM V3 global 1-arc-second product, which has been processed to remove data gaps and improve vertical accuracy (Farr et al. 2007). From this DEM, we derived primary topographic attributes, including elevation, slope, aspect and curvature.
iv. **Hydrological Data:** For hydrological analysis, we used the MERIT-Hydro global hydrography dataset, which combines elevation data with water body information to produce high-accuracy hydrographic features (Yamazaki et al. 2019). From this, we computed the Topographic Wetness Index (TWI) height above nearest drainage (HAND) and upstream drainage area.
v. **Climatic Data:** Daily daytime Land Surface Temperature was obtained from the Moderate Resolution Imaging Spectroradiometer (MODIS) aboard the Terra satellite, using the MOD11A1 Version 6.1 product. This dataset provides temperature data at a 1 km spatial resolution (Wan 2021). The values were converted to degrees Celsius for analysis. And near-surface specific humidity was sourced from the North American Land Data Assimilation System (NLDAS-2) (Parsons et al. 2010).
vi. **Land-cover Data:** For land-cover characterization, we used the Dynamic World global land-cover dataset, which provides near-real-time land-cover information derived from Sentinel-2 imagery at 10 m spatial resolution (Brown et al. 2022). Dynamic World provides class probabilities for nine land-cover categories: water, trees, grass, flooded vegetation, crops, shrub and scrub, built area, bare ground, and snow and ice. These land-cover probabilities were incorporated as ancillary features to characterize surface conditions surrounding the study sites and to help distinguish vegetation-stress signals from variability associated with different land-cover types.
vii. **Monitoring wells Data**: Ground-truth data for model training and validation consisted of in-situ measurements of metal and metalloid concentrations from groundwater monitoring wells. These records, which include concentrations of leaching metals and the publicly available Ashtracker database, which is maintained by the Environmental Integrity Project (EIP 2024).

### 3.3 Data preprocessing

Preprocessing involved cloud and cirrus masking using the QA60 band, and filtering by region of interest (ROI), date (2019–2023), and cloud cover (< 30 %). Median monthly composites were generated to reduce temporal noise. Each dataset was resampled to a uniform 20 m grid (WGS 84 / UTM Zone 17 N) and spatially aligned across layers.

### 3.4 Feature Engineering

A comprehensive set of feature variables was generated to characterize vegetation condition, temporal vegetation dynamics, land-cover variability, and environmental factors that may influence vegetation-environmental responses to subsurface leachate contamination. The feature space consisted of spectral, biophysical, temporal, soil, topographic, hydrological, climatic, and land-cover variables**.** Vegetation indices (VI) and biophysical parameters indices (BPI), and the Coefficient of Spatio-Temporal Variation (CSTV) were derived following the methodological framework established by Karim et al. (2026).

Table 2: Multi-source features and associated groups used as input to classification models.

| Feature group | Bands | n | Context |
|---|---|---|---|
| S2 | B1–B12, B8A (Sentinel-2 surface reflectance) | 12 | Canopy and surface optical properties across the visible, red-edge, NIR and SWIR regions; clouds masked with QA60 |
| VI/BPI | NDVI, MTCI, SAVI, CIRE, NDRE1, NDRE2, LAI, LCC, CCC | 9 | Greenness, chlorophyll, red-edge position, leaf area |
| CSTV | Spatiotemporal variations of the nine VI/BPI | 9 | Change in vegetation condition |
| RoC | Rates of change of the nine VI/BPI | 9 | Trend in vegetation condition |
| Land cover | Dynamic World class probabilities (water, trees, grass, flooded vegetation, crops, shrubs, built, bare, snow), land-cover label and vegetation probability | 11 | Land-cover composition |
| Topographic | DEM, slope, aspect, curvature | 4 | Terrain position and gradient |
| Soil | pH, sand, silt and clay content (0–5 cm) | 4 | Soil texture and chemistry |
| Climate | Land surface temperature, near-surface specific humidity, soil moisture, mean summer (JJA) precipitation | 4 | Thermal and moisture conditions |
| Hydrologic | Upstream area, height above nearest drainage, topographic wetness index | 3 | Hydrological position and drainage |

Together, these 65 features formed an integrated multi-source predictor space representing spectral response, vegetation biophysical condition, long-term vegetation dynamics, soil properties, terrain, hydrology, climate, and land cover**.** The combined feature stack was subsequently used as input to the machine-learning framework to identify spatial patterns of vegetation stress associated with leachate contamination while accounting for environmental factors that can independently influence vegetation condition.

## 3.5 Hazard Index from monitoring wells

### 3.5.1 Monitoring well data preparation

Groundwater monitoring-well data associated with coal combustion residuals at four power plant sites were compiled for the analysis. The datasets contained concentrations of metals and metalloids measured at individual monitoring wells over multiple sampling events spanning 2010–2019.

The raw datasets included metal constituent concentrations, monitoring-well identifiers, sampling dates, and geographic coordinates (latitude and longitude; WGS84). Prior to analysis, the datasets were processed using the following procedures:

i. Temporal aggregation: Where multiple sampling events were available for monitoring well, concentrations of each constituent were averaged across all valid sampling events to obtain a representative mean concentration for each well. Thus, the resulting values characterize average groundwater conditions over the available monitoring period rather than an individual sampling event.
ii. Non-detect handling: Concentrations reported below the analytical method detection limit (MDL) were substituted with one-half of the corresponding detection limit (MDL/2) prior to temporal aggregation.
iii. Unit standardization: Constituent concentrations were standardized to milligrams per liter (mg/L) prior to comparison with the selected reference concentrations. Constituents reported in other units were converted to the corresponding reference units before calculation.

Following preprocessing, each site was represented by a georeferenced monitoring-well dataset containing well locations and temporally aggregated concentrations for the available metals/metalloids.

### 3.5.2 Target constituent and reference thresholds

The four sites had 23 monitored leachate constituents. Because the magnitude and environmental significance of measured concentrations differ substantially among constituents, constituent-specific reference threshold concentrations were used to normalize the groundwater chemistry data for all sites.

Reference values in Table 3 were compiled from U.S. EPA standards and advisory values and, where applicable, WHO guidance. These values include Maximum Contaminant Levels (MCLs), action levels, Secondary Maximum Contaminant Levels (SMCLs), health-advisory values, and drinking-water guideline values. Because these benchmarks differ in their regulatory basis and purpose, they are collectively referred to in this study as reference thresholds and are used as normalization benchmarks within the proposed composite-index methodology.

Table 3: Reference threshold concentrations for metals/metalloids used in the study-specific Hazard Quotient (HQ) and composite Hazard Index (HI) calculations

| Constituent | Symbol | Threshold | Unit | Source |
|---|---|---|---|---|
| Antimony | Sb | 0.006 | mg/L | EPA MCL |
| Arsenic | As | 0.01 | mg/L | EPA MCL |
| Barium | Ba | 2 | mg/L | EPA MCL |
| Beryllium | Be | 0.004 | mg/L | EPA MCL |
| Boron | B | 3 | mg/L | WHO guideline |
| Cadmium | Cd | 0.005 | mg/L | EPA MCL |
| Chromium, total | Cr | 0.1 | mg/L | EPA MCL |
| Hexavalent chromium | Cr(VI) | 0.1 | mg/L | EPA HAL |
| Cobalt | Co | 0.006 | mg/L | EPA HAL |
| Copper | Cu | 1.3 | mg/L | EPA action level |
| Iron | Fe | 0.3 | mg/L | EPA SMCL |
| Lead | Pb | 0.015 | mg/L | EPA action level |
| Lithium | Li | 0.04 | mg/L | EPA HAL |
| Manganese | Mn | 0.3 | mg/L | EPA SMCL/HAL |
| Mercury | Hg | 0.002 | mg/L | EPA MCL |
| Molybdenum | Mo | 0.04 | mg/L | EPA HAL |
| Nickel[b] | Ni | 0.1 | mg/L | EPA (remanded) |
| Radium-226 + radium-228[b] | Ra | 5 | pCi/L | EPA MCL |
| Selenium | Se | 0.05 | mg/L | EPA MCL |
| Silver | Ag | 0.1 | mg/L | EPA SMCL |
| Strontium | Sr | 4 | mg/L | EPA lifetime HA |
| Thallium | Tl | 0.002 | mg/L | EPA MCL |
| Uranium | U | 0.03 | mg/L | EPA MCL |
| Zinc | Zn | 5 | mg/L | EPA SMCL |

[a] The former EPA MCL for nickel (0.1 mg/L) was remanded and is used here as a reference value only.
[b] Combined activity of radium-226 and radium-228; the HQ is computed from activities in pCi/L.

### 3.5.3 Hazard Quotient (HQ) computation

To account for the large differences in concentration magnitude and reference thresholds among the monitored constituents, for each monitoring well $j$ and each contaminant $i$, a dimensionless Hazard Quotient (HQ) was computed as:

$$HQ_{i,j} = \frac{C_{i,j}}{T_i} \quad (1)$$

where $C_{i,j}$ is the temporally aggregated concentration of constituent $i$ at well $j$, and $T_i$ is the corresponding reference threshold. This ratio is the basis of the hazard-quotient approach of the U.S. Environmental Protection Agency (US EPA 1989, 2000). The same concentration-to-standard ratio is used in groundwater quality indices for trace metals (Mohan et al. 1996; Backman et al. 1998; Edet and Offiong 2002). An $HQ_{i,j} > 1$ indicates that the mean concentration at well $j$ for contaminant $i$ exceeds its threshold. This normalization places all constituents on a common dimensionless scale regardless of their individual units or magnitudes, enabling direct inter-contaminant comparison and summation. For descriptive mapping and interpretation, HQ values were grouped into five classes in Table 4 which were defined for this study to characterize relative contaminant enrichment and threshold exceedance and are not intended to represent regulatory risk categories.

Table 4: Study-defined Hazard Quotient (HQ) classification and interpretation based on constituent-specific reference threshold exceedance

| HQ range | Class | Interpretation |
|---|---|---|
| HQ < 0.5 | Low | Below 50% of the reference threshold |
| 0.5 ≤ HQ < 1.0 | Moderate | 50–100% of the reference threshold |
| 1.0 ≤ HQ < 1.5 | Elevated | 1.0–1.5 times the reference threshold |
| 1.5 ≤ HQ < 2.0 | High | 1.5–2.0 times the reference threshold |
| HQ ≥ 2.0 | Very high | At least 2 times the reference threshold |

The site-level composite contamination burden at each well was quantified using the Hazard Index (HI), defined as the arithmetic sum of individual HQ values across all $n$ constituents:

$$HI_j = \sum_{i=1}^{n} HQ_{i,j} = \sum_{i=1}^{n} \frac{C_{i,j}}{T_i} \tag{2}$$

Summing HQ values follows the dose-additivity assumption that the U.S. EPA adopts for chemical mixtures (US EPA 1986, 2000). The heavy metal evaluation index (Edet and Offiong 2002) uses a similar summation. The HI therefore integrates information from multiple leachate constituents into a single dimensionless indicator. A constituent occurring at its reference concentration contributes 1.0 to the composite HI, while concentrations below or above the reference threshold contribute proportionally smaller or larger values, respectively. In this study, $n$ = 23 monitored constituents were considered across sites. These constituents are characteristic of CCR leachate. Multi-constituent analyses of this kind have previously been used to document groundwater impacts from CCR disposal (Ruhl et al. 2012; Harkness et al. 2016) . For descriptive analysis, HI values were grouped into five study-specific classes (Table 5).

Table 5: Study-defined classes of the composite Hazard Index (HI).

| HI range | Class | Interpretation |
|---|---|---|
| HI < 1 | Below threshold | Combined burden below the reference thresholds |
| 1 ≤ HI < 5 | Low | One or more constituents slightly exceed their thresholds |
| 5 ≤ HI < 10 | Moderate | Exceedance by several constituents, or substantial exceedance by one |
| 10 ≤ HI < 20 | High | Substantial multi-constituent exceedance |
| HI ≥ 20 | Very high | Severe multi-constituent exceedance |

### 3.5.4 Spatial Interpolation of the Composite HI

The HI values were assigned to the coordinates of their monitoring wells to create a georeferenced point dataset. Because the wells sample groundwater only at discrete locations, ordinary kriging (Matheron 1963; Webster and Oliver 2007) was used to estimate HI between wells and produce a continuous surface. Kriging has been widely used to map groundwater contamination and health-risk indices from monitoring-well data (Belkhiri et al. 2017).

The kriged estimate at an unsampled location $s_0$ can be expressed as:

$$\hat{HI}(s_0) = \sum_{j=1}^{N} \lambda_j \, HI(s_j) \tag{3}$$

where $HI(s_j)$ is the calculated composite HI at monitoring well $j$, $\lambda_j$ represents its kriging weight, and $N$ is the number of monitoring wells contributing to the prediction.

An exponential semivariogram model was used to characterize spatial dependence, with model parameters specified for each site according to the spatial configuration and variability of monitoring-well HI values. Kriging was performed separately for each site so that the resulting surfaces reflected the site-specific distribution of groundwater contamination.

The resulting interpolated HI surface was clipped to the corresponding study region of interest (ROI) and resampled/exported at the spatial resolution required for integration with the satellite-derived predictor variables.

### 3.5.5 Min–Max Normalization of the Kriged HI

Because the absolute magnitude of the composite HI can vary substantially among sites, the interpolated HI surface was transformed to a bounded scale using min–max normalization:

$$HI_{\text{norm}}(s) = \frac{HI_{\text{kriged}}(s) - HI_{\min}}{HI_{\max} - HI_{\min}} \tag{4}$$

where $HI_{\text{kriged}}(s)$ is the interpolated HI at pixel $s$ and $HI_{\min}$ and $HI_{\max}$ represent the minimum and maximum interpolated HI values within the normalization domain.

The transformation produces values ranging from 0 to 1:

$$0 \le HI_{\text{norm}} \le 1$$

where values approaching 0 represent areas with relatively lower composite metals contamination and values approaching 1 represent areas with relatively higher composite contamination. The normalized HI in this study therefore represents the spatial intensity of multi-metal groundwater contamination relative to the observed contamination range, rather than the concentration of any individual constituent.

### 3.6 Generation of target Classes for machine learning models

For subsequent remote-sensing classification, the continuous normalized HI surface was converted into binary target classes representing lower-HI and higher-HI conditions. A classification threshold α was applied as:

$$Y(s) = \begin{cases} 0, & HI_{\text{norm}}(s) < \alpha \\ 1, & HI_{\text{norm}}(s) \geq \alpha \end{cases} \quad (5)$$

where α is the 25th percentile of the $HI_{\text{norm}}$ distribution within the site ROI. Class 0 therefore contains the quarter of the area with the lowest composite burden, and Class 1 contains the remaining three quarters. Converting continuous groundwater-quality data into threshold-based classes is common in machine-learning studies of groundwater contamination (Rodriguez-Galiano et al. 2014; Podgorski and Berg 2020). A percentile threshold was used instead of a fixed regulatory value for three reasons. First, the composite HI exceeded 1 throughout the ROI, so the conventional HI = 1 cut-off would have assigned every pixel to the same class. Second, $HI_{\text{norm}}$ is scaled relative to each site's own range, so a fixed absolute value of α would have a different meaning at each site. Third, the 25th percentile separates the lowest-burden areas from the rest of the site. The resulting class imbalance of about 25:75 was addressed by evaluating the models with class-balanced metrics.

The binary raster served as the target for the supervised models. Vegetation reflectance has previously been used as an indirect indicator of metal contamination (Kooistra et al. 2004; Zhang et al. 2018; Karim et al. 2026), and spatial predictors have been used to predict groundwater contaminant concentrations at regional scale (Knoll et al. 2019).

The same procedure was applied to individual constituents for the constituent-level analysis. For each metal with at least one well exceeding its reference threshold (HQ ≥ 1) and at least three distinct well values at a site, the well HQ values were kriged with an exponential semivariogram (sill equal to the variance of the well HQ, no nugget, and the range used for the composite HI), normalized to 0–1 within the site ROI, and converted to binary classes at the 25th percentile of the normalized HQ. This produced 34 metal–site targets: seven at Site A, eight at Site B, twelve at Site C and seven at Site D. The framework presented here therefore links monitoring-well geochemistry, the interpolated multi-metal HI and remotely sensed vegetation response.

### 3.7 Feature stacking, data sampling and train-validation-test split

All predictor layers (Table 2) were stacked into one 65-band composite per site on the resampled 10 m grid, and the binary HI class raster was added as the target, so that each pixel record contained its predictor values and its reference class. Because neighbouring pixels have similar predictor and label values, a random pixel split would place near-identical pixels in the training and test sets and overestimate accuracy (Roberts et al. 2017; Ploton et al. 2020). The pixels were therefore partitioned with a spatial block design. The site grid was divided into blocks of 32 × 32 pixels (640 m × 640 m), and whole blocks were assigned to the training (70%), validation (15%) and test (15%) sets. Of 500 random block assignments, the one whose partition sizes and class-1 fractions were closest to the target shares and to the site-wide class-1 fraction was retained. Training pixels within 8 pixels (160 m) of a validation or test block, and validation pixels within 8 pixels of a test block, were excluded as a buffer. The test sets contained 9,216 pixels at Site A, 9,600 at Site B, 34,240 at Site C and 28,960 at Site D, and the training sets 30,828, 30,119, 113,416 and 96,160 pixels, respectively. The same partition was used for every model and seed, so that differences among models were not confounded with differences in the test data.

### 3.8 Machine learning models training

To classify vegetation stress associated with leached metal hazard, six supervised machine learning algorithms were implemented in Python with scikit-learn (Pedregosa et al. 2011), with settings that follow the corresponding GEE classifiers: CART (Breiman et al. 1984), RF (Breiman 2001), GTB (Jerome H. Friedman 2001), SVM (Cortes and Vapnik 1995; Chang and Lin 2011), KNN (Cover and Hart 1967), and NB (Domingos and Pazzani 1997). They cover tree-based, ensemble, probabilistic, distance-based and margin-based learning. Predictors were standardized to z-scores with the mean and standard deviation of the training pixels, and class weights inversely proportional to class frequency were applied in all models to offset the 25:75 class ratio. RF used 100 trees with $\sqrt{p}$ candidate features per split (p = number of predictors); CART was grown without pruning (minimum leaf size 1); GTB was implemented as histogram-based gradient boosting (Ke et al. 2017) with 300 iterations, a learning rate of 0.05 and up to 256 leaves per tree; SVM used a radial basis function kernel ($C = 1$, $\gamma = 1/p$) trained on a stratified subsample of 10,000 training pixels; KNN used $k = 15$; and NB was a categorical naïve Bayes classifier applied to the standardized predictors clipped to ±4 and discretized into 64 bins. A pixel was assigned to class 1 when its predicted class-1 probability was at least 0.5 (for SVM, when its decision value was at least 0). Each model was trained with three random seeds (42, 43 and 44) on the same partition, and the resulting binary maps (class 1 = higher HI, class 0 = lower HI) formed the basis for model evaluation, comparison and spatial prediction.

### 3.9 Model evaluation

Model performance was evaluated using five complementary classification metrics that characterize both overall agreement and class-specific predictive performance: Overall Accuracy, Cohen's Kappa coefficient (κ), precision, recall, and $F_1$-score. All metrics were calculated from the confusion matrix using the held-out test dataset.

Accuracy represents the proportion of correctly classified samples relative to the total number of evaluated samples and was calculated as:

$$\text{Accuracy} = \frac{TP+TN}{TP+TN+FP+FN} \quad (6)$$

where *TP*, *TN*, *FP*, *FN* are true positives, true negatives, false positives, and false negatives, respectively. Accuracy provides an overall measure of classification performance; however, it can be influenced by class imbalance when one class is substantially more prevalent than the other.

To account for agreement that could occur by chance, Cohen's Kappa coefficient $\kappa$ was also calculated as:

$$\kappa = \frac{P_o - P_e}{1 - P_e} \tag{7}$$

where $P_o$ is the observed agreement, equivalent to overall accuracy, and $P_e$ is the expected agreement by chance. For a binary confusion matrix,

$$P_o = \frac{TP+TN}{N};\ P_e = \frac{(TP+FP)(TP+FN)+(TN+FN)(TN+FP)}{N^2};\ N = TP + TN + FP + FN \tag{8}$$

a $\kappa$ value approaching 1 indicates stronger agreement between predicted and reference classes beyond that expected by chance, whereas values near 0 indicate agreement comparable to chance.

To assess how effectively the models distinguish between lower HI (class 0) and higher HI (class 1) conditions, we derived precision, recall, and the $F_1$-score for each class using the confusion matrix. Precision represents the proportion of predicted positives that are truly positive:

$$\text{Precision} = \frac{TP}{TP+FP} \tag{9}$$

High precision indicates that the model rarely mislabels lower hazard areas as higher.

Recall measures the ability of the classifier to correctly identify actual hazardous pixels:

$$\text{Recall} = \frac{TP}{TP+FN} \tag{10}$$

A high recall value indicates that the model effectively captures hazard signatures and minimizes missed detections.

The $F_1$-score, defined as the harmonic mean of precision and recall,

$$F_1 - \text{score} = 2 \times \frac{(\text{Precision}\times\text{Recall})}{(\text{Precision}+\text{Recall})} \tag{11}$$

$F_1$-score provides a single, balanced metric reflecting both false positives and false negatives. This is particularly important for contamination-mapping applications, where misclassification of hazard areas may pose environmental and human-health risks.

Together, these metrics offer a comprehensive evaluation of model performance, ensuring that the selected classifier is not only accurate overall but also reliable in distinguishing subtle spectral stress patterns associated with leached metal hazard.

### 3.10 Model selection and transfer

The model used for the maps, the feature-importance analysis and the constituent-level analysis was selected from the confusion matrices: because the framework is intended for screening, missing a higher-HI area (a false negative) was considered more costly than a false alarm, and the model with the fewest false negatives at every site was preferred among the models with high overall accuracy. Cross-site transferability was evaluated with leave-one-site-out testing: each model was trained on pixels from three sites and tested on the fourth site, which contributed no training data. To reduce differences in predictor scale among sites, the predictors of each site were standardized with that site's own mean and standard deviation, which uses no label information. A stratified sample of 20,000 pixels per training site (60,000 pixels in total; 10,000 for SVM) was used for training, and a stratified sample of 50,000 pixels of the held-out site for testing. The model settings and seeds were the same as in the within-site analysis.

### 3.11 Feature-set ablation

To test whether the within-site skill comes from the vegetation signal or from environmental, RF was retrained on the same spatial blocks and seeds with four reduced predictor sets: vegetation only (Sentinel-2 bands, vegetation indices and biophysical parameters, CSTV and rate of change; 39 predictors), environment only (topographic, soil, climatic, hydrologic and land-cover variables; 26 predictors) and the distance of each pixel to the nearest CCR unit (one predictor).

### 3.12 Correlation between well HQ and vegetation indices

To test the hypothesis that higher groundwater metal hazard is associated with lower vegetation index values, the summer (growing-season) value of each of the nine vegetation indices and biophysical parameters was averaged within a 30 m buffer around each monitoring well and related to the well HQ of the metals that exceeded their reference thresholds at all four sites, using Spearman's rank correlation coefficient (ρ) separately for each metal and site (Spearman 1904). A negative ρ indicates lower index values, i.e. less greenness, chlorophyll or leaf area, at wells with higher HQ. For each correlation, the number of wells and the p-value are reported, and p-values were adjusted with the Benjamini–Hochberg procedure (Benjamini and Hochberg 1995) within each site. Because wells with high HQ tend to lie close to the CCR units, a partial Spearman correlation controlling for the distance of each well to the nearest unit was also computed. Upgradient wells, which sample groundwater before it passes the units, served as a control: the index values at downgradient and upgradient wells were compared with a one-sided Mann–Whitney U test (Mann and Whitney 1947) . A sign test over the metal–site combinations summarized the direction of the associations. To relate the composite hazard directly to vegetation condition, the HI at each well was also plotted against the CIRE and CCC level averaged within 60 m of the well (about 28 pixels of the 20 m grid, compared with about 7 within 30 m), with a Theil–Sen trend line fitted to log10(HI) (Sen 1968; Theil 1992). CIRE and CCC were chosen because they have a direct physical meaning, leaf and canopy chlorophyll content, and were among the most sensitive indicators of vegetation stress at a CCR site in our earlier work (Karim et al. 2026a).

### 3.13 Mapping and visualization

The classified maps were visualized using color-coded symbology to highlight gradients of hazard. Additionally, classified hazard regions were overlaid with kriged metal-concentration surfaces to enable spatial validation and assess correspondence between vegetation stress patterns and underlying contamination hazard levels. This visualization workflow provided an interactive and reproducible environment for examining spatial patterns across sites and evaluating model performance.

## 4 Results & Discussions

### 4.1 Spatiotemporal variation patterns across all sites

CSTV of BPI and VI analysis, computed from 2019 and 2025 Sentinel-2 growing-season composites, revealed distinct spatio-temporal vegetation stress patterns across the four study sites. By applying the CSTV thresholds defined in Karim et al. (2026) where CSTV ≥ 2 indicates abrupt or high stress, CSTV values close to 1 denotes persistent but stable stress and CSTV = 0 indicates no stress. Figure 3(bottom-left) plots the percentage of vegetation cover in each site and the bar chart in Figure 3(bottom-right) quantifies the percentage of vegetation stress detected from 2019 to 2025 exhibiting multi-year stress, while the CSTV maps in Figure 3(top) illustrate the spatial distribution of these stress responses. Figures of vegetation indices and rate of change of the corresponding indices can be found in supplementary material.

Across all sites, biophysical parameters (CCC, LCC, LAI) consistently exhibit the highest proportions of stressed vegetation ranging from 68% to 84%, indicating strong temporal fluctuations in chlorophyll content and canopy structure from 2019 to 2025. This alignment confirms that chlorophyll degradation and canopy thinning are the most temporally sensitive and reliable indicators of contamination-driven stress.

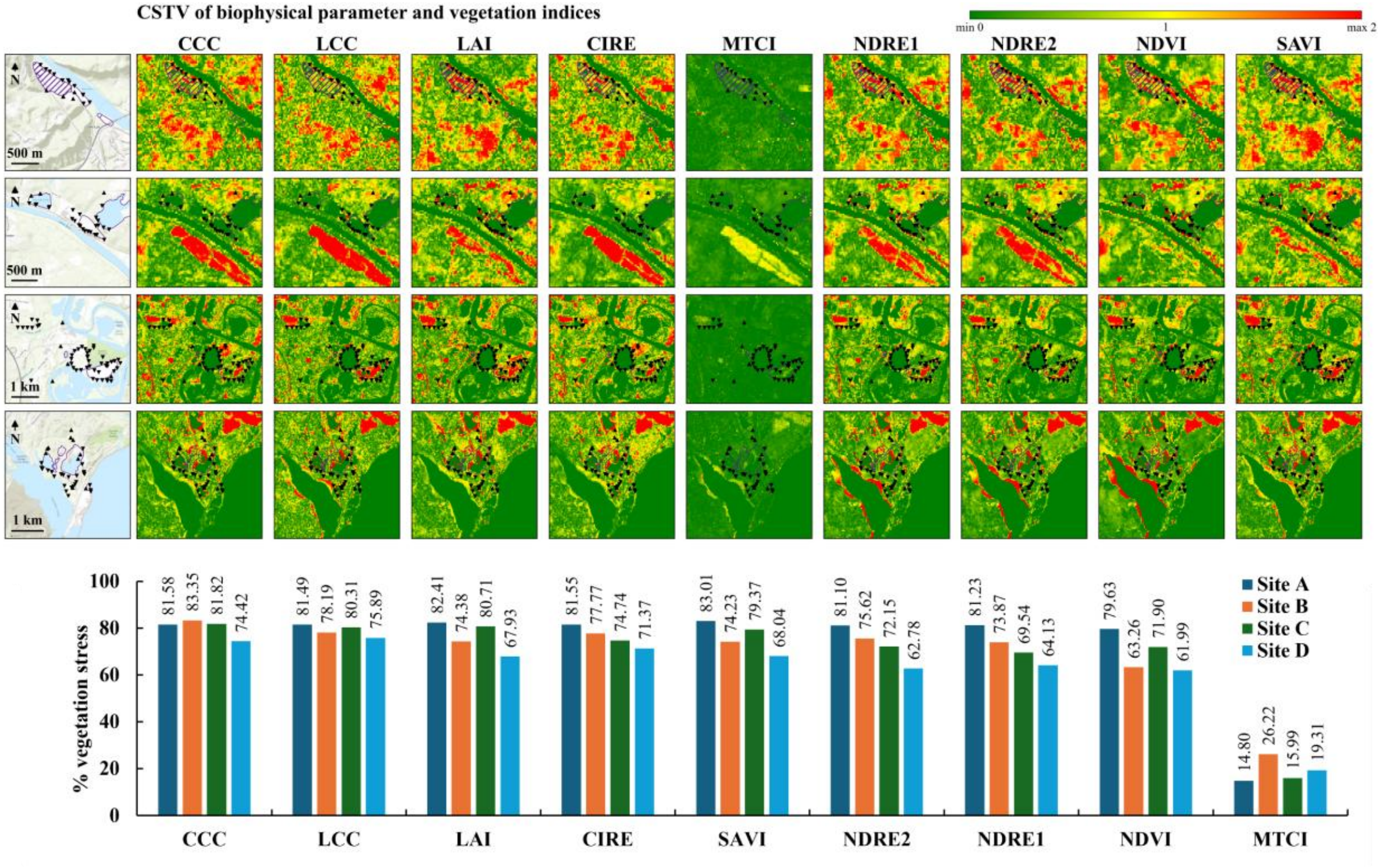

Figure 3: The CSTV maps (top) visualize multi-year spatiotemporal variability, where higher values (yellow to red) indicate persistent or abrupt vegetation stress potentially associated with leached-metal contamination. The bar chart (bottom left) quantifies the proportion of stressed vegetation detected by each VI/BPI and the bar chart (bottom right) summarizes overall vegetation cover within each site.

Among vegetation indices, CIRE, SAVI, NDRE1&2 and NDVI show strong CSTV-derived stress proportions (generally in 62% to 84%) and spatial patterns that closely mirror those of BPI. Their CSTV maps reveal well-defined stress patches capturing multi-year physiological responses. However, MTCI, consistent with its low sensitivity in correlation analyses, displays minimal CSTV response across all sites (<26%) and its CSTV maps remain largely green, indicating little multi-year variation and reinforcing its limited diagnostic value for gradual, chronic stress typically associated with leached-metal exposure.

The CSTV maps thus identify both abrupt and persistent multi-year stress at all four sites; its association with groundwater metal hazard is tested at the monitoring wells below.

## 4.2 Spatial and site-level patterns of Hazard Quotients

The HQ differed markedly in magnitude and spatial distribution among the four sites (Figure 4 and Figure 5). The kriged HQ surfaces showed hotspots around subsets of monitoring wells rather than uniform contamination, and the magnitude and extent of these hotspots differed among constituents.

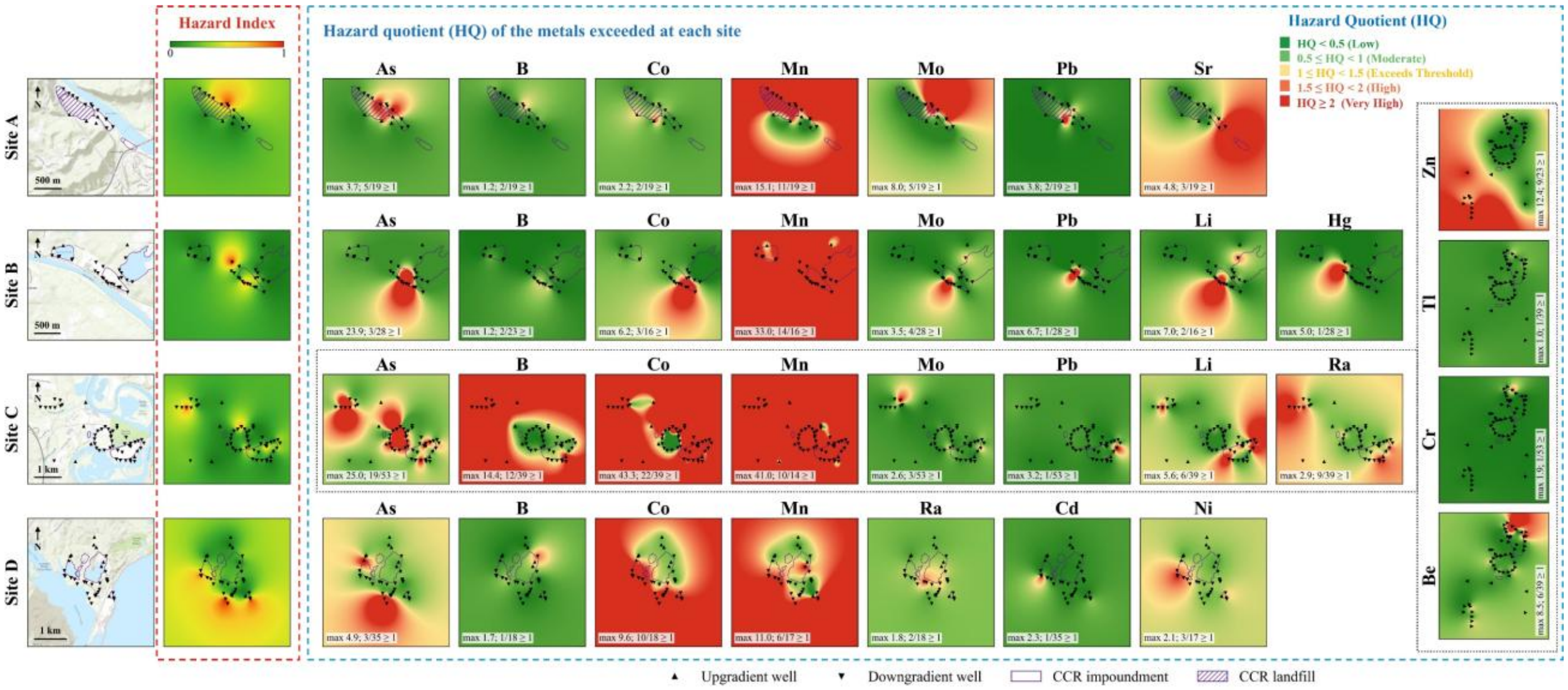

Figure 4: Kriging-interpolated Hazard Quotient (HQ) maps for site-specific metals across four CCR sites. The leftmost panels show the site basemaps, while subsequent panels show spatial HI (normalised) and HQ distributions derived from groundwater monitoring wells. Only constituents with recorded threshold exceedances at each site are shown.

Mn exhibited the most widespread HQ exceedances across the four study sites (Figure 5). Its site HQ exceeded 1 at all four sites (11.0–41.0), and 35–88% of the monitoring wells exceeded the threshold, the largest share being at Site B (14 of 16 wells). The spatial HQ distributions (Figure 4) further supported the extensive Mn hazard observed at Sites B and C.

Co reached the highest site HQ of all constituents (43.3 at Site C) and exceeded the threshold at more than half of the wells at Sites C (22 of 39) and D (10 of 18), whereas exceedances at Sites A and B were limited to two and three wells. The interpolated HQ (Figure 4) separates localized areas of elevated Co from less affected areas.

As exceeded the reference threshold at all four sites (site HQ 3.7–25.0). The high site HQ at Site B (23.9) arose from only three of 28 wells, whereas at Site C 19 of 53 wells exceeded the threshold, which shows that a high maximum HQ does not necessarily indicate widespread exceedance.

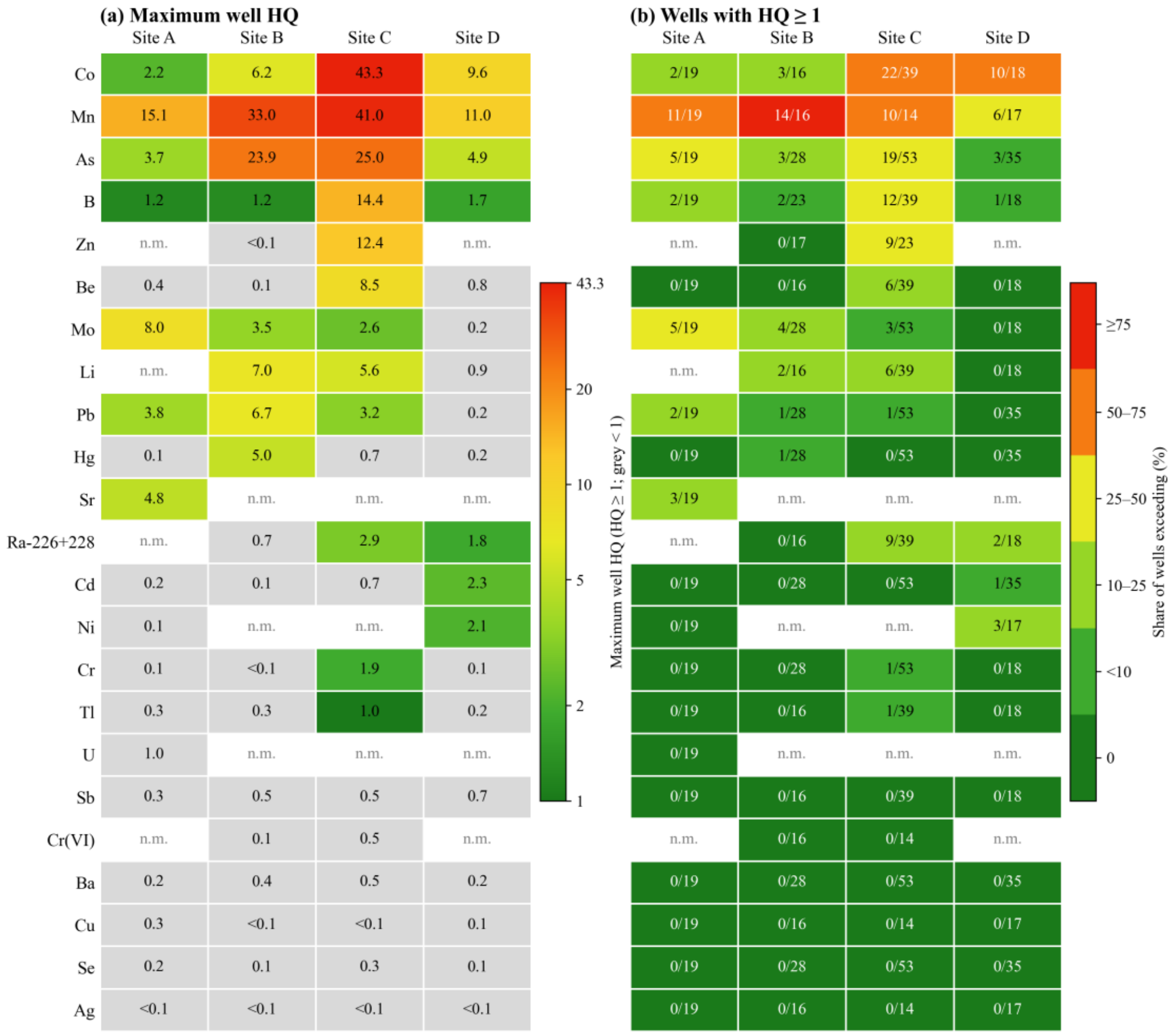

Figure 5: Distribution of groundwater metal HQ across four study sites. (a) Site-level HQ values for 23 constituents (n.m., not measured). HQ ≥ 1 indicate exceedance of the reference threshold and are displayed using a logarithmic green-to-red color scale, whereas values below 1 are shown in gray. (b) Number of monitoring wells with HQ ≥ 1 relative to the total number of wells analyzed for each constituent, with cell colors representing the percentage of wells exceeding the reference threshold.

Further constituents exceeded their thresholds at individual sites (Figure 5): Mo, Pb, Sr and B at Site A; Li, Pb, Hg, Mo and B at Site B; B, Zn, Be, Li, Ra-226+228, Pb, Mo, Cr and Tl at Site C; and Ni, Cd, B and Ra-226+228 at Site D. Ba, Sb, Cr(VI), Cu, Se, Ag and U remained below their thresholds at all sites where they were measured. Several constituents had a high site HQ although only one or a few wells exceeded the threshold, e.g. Pb (HQ = 6.7) and Hg (5.0) at Site B, Cd (2.3) at Site D, and Pb (3.2), Cr (1.9) and Tl (1.0) at Site C (Figure 4, Figure 5), which indicates localized exceedances. In contrast, Mn exceeded the reference threshold at more than half of the wells at Sites A, B and C, and Co at more than half of the wells at Sites C and D.

Overall, Site C had the strongest multi-constituent signature, with 12 constituents at HQ ≥ 1 and the highest HQ for Mn, Co, As and B. Site B had eight constituents at HQ ≥ 1 and the most widespread Mn exceedance, and Sites A and D had seven each. The relative severity of groundwater hazard therefore depends on whether the HQ magnitude, the number of constituents or the share of exceeding wells is considered.

Groundwater metal contamination at the four sites is therefore multi-constituent, spatially heterogeneous and site dependent. Although Mn, As, Co and B exceeded their reference thresholds at all four sites, no single constituent characterized the overall pattern, which supports combining the HQ into a composite HI while retaining the constituent-level maps.

## 4.3 Association between leached metal hazard and vegetation indices near monitoring wells

The composite HI at the monitoring wells was compared with CIRE and CCC (Figure 6). In this study, CIRE among the vegetation indices and CCC among the biophysical parameters showed comparatively higher vegetation stress near the CCR units. This agrees with our earlier work at a CCR site, where were the indices most strongly correlated with higher vegetation stress along with groundwater metal hazard (Karim et al. 2026). These two indices track leaf and canopy chlorophyll content. At three of the four sites, both indices declined as HI increased. The association was strongest and significant at Site B: CIRE $\rho = -0.67$, CCC $\rho = -0.71$ ($n = 28$, $p < 0.001$). It was unchanged after controlling for the distance to the nearest CCR unit (partial $\rho = -0.68$ and $-0.71$, $p < 0.001$). Sites A ($n = 19$) and D ($n = 35$) showed weaker negative trends that were not significant ($\rho = -0.25$ to $-0.38$, $p = 0.085$–$0.20$). At Site A, the CCC association became significant once distance was controlled (partial $\rho = -0.48$, $p = 0.042$). Upgradient wells plot at low HI and high index values and indicate the vegetation condition outside the influence of the CCR units. The relationships between HI and all nine vegetation indices and biophysical parameters are given in Figure S0 of the Supplementary Material. At Sites A, B and D, all nine had negative $\rho$.

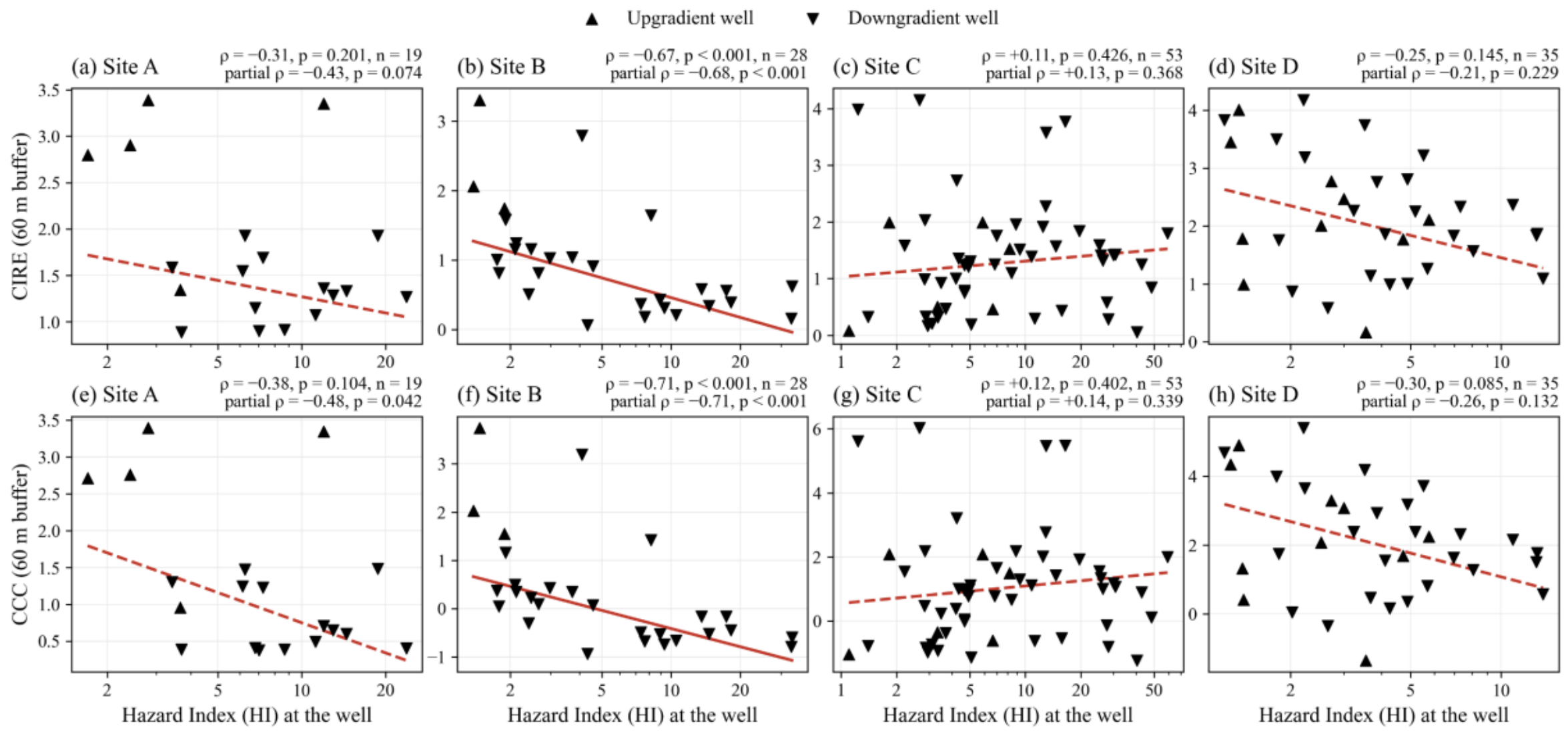

Figure 6: Hazard Index (HI) at each monitoring well against (top) CIRE and (bottom) CCC averaged within 60 m of the upgradient and downgradient well at Sites A–D; HI on a logarithmic axis. Lines: Theil–Sen trend on log10(HI), solid where Spearman's $\rho$ is significant ($p < 0.05$). $\rho$, p and n, and the partial $\rho$ controlling for the distance to the nearest CCR unit, are given above each panel.

Site C did not follow this pattern (CIRE ρ = +0.11, CCC ρ = +0.12, n = 53, $p > 0.40$). Site C lies along a river. Of its wells, 20 are surrounded mainly by water (13) or built and bare surfaces (7) within 60 m. The other sites have at most six such wells each (Figure 7). At these wells the index values reflect the surface type, not vegetation condition. The median CIRE was 0.43–0.44, against 1.82 at wells surrounded mainly by trees. The water-dominated wells also had the lowest HI (median 4.7), which produces the weak positive slope. With only the 32 wells surrounded mainly by vegetation, the positive slope disappeared (ρ ≈ −0.07), but no decline appeared either. Near the Site C wells, land cover rather than metal stress therefore controls the index values.

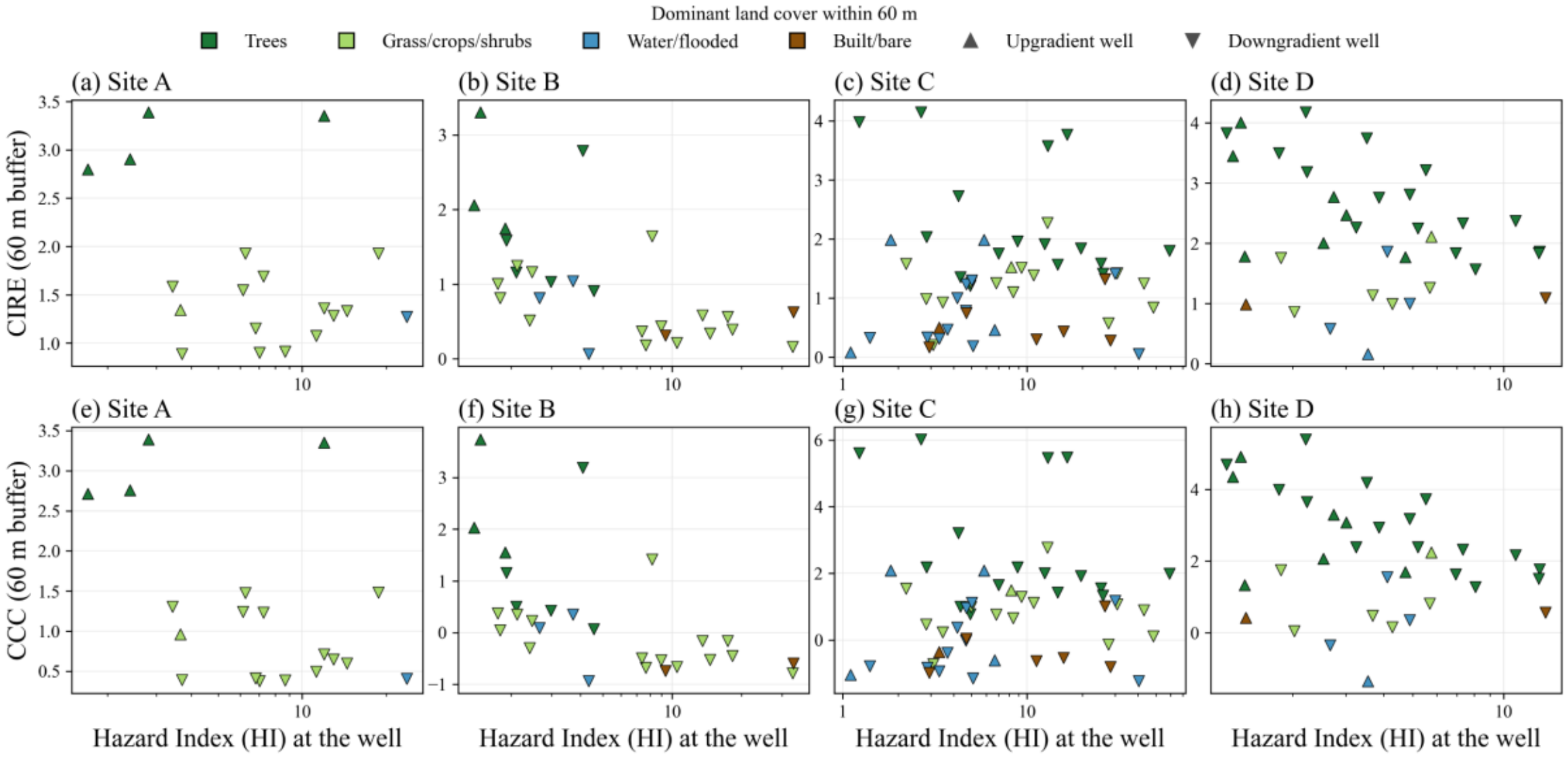

Figure 7: Well HI against CIRE and CCC within 60 m of the wells, coloured by the dominant Dynamic World land cover of the buffer.

The individual metals showed the same direction more consistently (Table S1 in the Supplementary Material). Across all 33 metal–site combinations with HQ above the reference threshold, the mean Spearman ρ over the nine vegetation indices was negative in 24. In 20 of these 24, all nine indices were negative. With no association, 24 or more negative signs out of 33 would occur with a probability of 0.007 (one-sided sign test). The value is only indicative, because the metals at a site share the same wells and are therefore not independent. The sign was negative in 7 of 7 combinations at Site A, 7 of 8 at Site B and 7 of 7 at Site D. The only exception outside Site C was Pb at Site B (mean ρ = 0.10). For the four metals that exceeded their thresholds at all four sites (Mn, As, Co and B), the mean ρ was negative in 14 of the 16 combinations ($p = 0.002$). The strongest associations were Co at Sites B and D (mean ρ = −0.60 and −0.49); B at Sites A, B and D (−0.59, −0.47 and −0.35); Mo at Site A (−0.51); Mn at Sites B and D (−0.49 and −0.41); Li at Site B (−0.34; As was weaker but negative at all four sites (−0.04 to −0.29). Site C was the exception. There the mean ρ was positive for 8 of 11 metals (Co, B, Mo, Ra, Li, Be, Cr and Tl; 0.04 to 0.43), with all nine indices positive. Only Mn, As and Pb were negative (−0.16, −0.20 and −0.10). This is consistent with the land-cover effect described above. Each coefficient is based on 14–53 wells and is uncertain on its own. The evidence lies in the consistency of the negative sign across indices, metals and sites. Near the wells, a vegetation response to the leached metals is detectable at three of the four sites. Across a whole site, it is weaker than the environmental gradients that dominate the pixel-level classification (see the feature-importance analysis in section 4.7).

### 4.4 Classification performance of machine learning models across study sites

The six classifiers differed clearly in their ability to separate the lower- and higher-HI classes (Table 6). Mean accuracy over the four sites ranged from 0.746 (NB) to 0.856 (GTB), and mean κ from 0.369 (NB) to 0.559 (GTB). The tree ensembles GTB and RF achieved the highest mean accuracy (0.856 and 0.852), consistent with the effectiveness of ensemble methods for high-dimensional, nonlinear, multi-source remote-sensing data (Breiman 2001; Gislason et al. 2006; Belgiu and Drăguţ 2016).

Table 6: Performance of the six supervised machine learning models at the four sites on the held-out test blocks (mean of three seeds; SD ≤ 0.03): overall accuracy (OA), balanced accuracy (BA), Matthew's correlation coefficient (MCC), Cohen's κ, class-wise precision, recall and F1-score, and false negatives (FN, missed higher-HI pixels; mean count). All class 1: majority-class baseline that predicts every pixel as higher HI. Bold: best model value per site (lowest for FN).

| | | | | | Class 0 (lower HI) | | | Class 1 (higher HI) | | | |
|---|---|---|---|---|---|---|---|---|---|---|---|
| Model | OA | BA | MCC | κ | Precision | Recall | F1 | Precision | Recall | F1 | FN |
| *Site A* | | | | | | | | | | | |
| *All class 1* | 0.743 | 0.500 | 0.000 | 0.000 | – | 0.000 | 0.000 | 0.743 | 1.000 | 0.853 | 0 |
| RF | 0.924 | 0.866 | 0.797 | 0.787 | 0.950 | 0.745 | 0.835 | 0.918 | 0.986 | 0.951 | **93** |
| GTB | 0.870 | 0.789 | 0.641 | 0.631 | 0.831 | 0.623 | 0.712 | 0.880 | 0.956 | 0.916 | 301 |
| CART | 0.842 | 0.778 | 0.574 | 0.572 | 0.711 | 0.647 | 0.677 | 0.882 | 0.909 | 0.895 | 623 |
| SVM | 0.922 | 0.912 | 0.802 | 0.801 | 0.820 | 0.891 | 0.854 | 0.961 | 0.932 | 0.946 | 465 |
| KNN | 0.860 | 0.805 | 0.624 | 0.623 | 0.744 | 0.692 | 0.717 | 0.896 | 0.917 | 0.907 | 565 |
| NB | 0.777 | 0.766 | 0.488 | 0.477 | 0.549 | 0.742 | 0.631 | 0.899 | 0.789 | 0.840 | 1,443 |
| *Site B* | | | | | | | | | | | |
| *All class 1* | 0.737 | 0.500 | 0.000 | 0.000 | – | 0.000 | 0.000 | 0.737 | 1.000 | 0.848 | 0 |
| RF | 0.768 | 0.560 | 0.297 | 0.167 | 0.978 | 0.121 | 0.215 | 0.761 | 0.999 | 0.864 | **7** |
| GTB | 0.792 | 0.610 | 0.397 | 0.291 | 0.933 | 0.226 | 0.364 | 0.782 | 0.994 | 0.876 | 41 |
| CART | 0.780 | 0.608 | 0.341 | 0.277 | 0.755 | 0.246 | 0.370 | 0.783 | 0.971 | 0.867 | 202 |
| SVM | 0.807 | 0.639 | 0.455 | 0.359 | 0.949 | 0.284 | 0.437 | 0.795 | 0.995 | 0.884 | 38 |
| KNN | 0.727 | 0.552 | 0.149 | 0.128 | 0.456 | 0.180 | 0.258 | 0.759 | 0.923 | 0.833 | 545 |
| NB | 0.753 | 0.570 | 0.228 | 0.181 | 0.602 | 0.184 | 0.282 | 0.766 | 0.957 | 0.851 | 307 |
| *Site C* | | | | | | | | | | | |
| *All class 1* | 0.756 | 0.500 | 0.000 | 0.000 | – | 0.000 | 0.000 | 0.756 | 1.000 | 0.861 | 0 |
| RF | 0.828 | 0.671 | 0.478 | 0.423 | 0.840 | 0.365 | 0.509 | 0.826 | 0.978 | 0.896 | **580** |
| GTB | 0.871 | 0.780 | 0.627 | 0.615 | 0.819 | 0.603 | 0.695 | 0.882 | 0.957 | 0.918 | 1,113 |
| CART | 0.841 | 0.733 | 0.535 | 0.521 | 0.753 | 0.522 | 0.617 | 0.860 | 0.945 | 0.900 | 1,431 |
| SVM | 0.760 | 0.723 | 0.414 | 0.408 | 0.508 | 0.650 | 0.570 | 0.876 | 0.796 | 0.834 | 5,277 |
| KNN | 0.770 | 0.651 | 0.331 | 0.326 | 0.538 | 0.417 | 0.470 | 0.824 | 0.885 | 0.853 | 2,986 |
| NB | 0.732 | 0.741 | 0.424 | 0.400 | 0.470 | 0.759 | 0.581 | 0.903 | 0.724 | 0.803 | 7,143 |
| *Site D* | | | | | | | | | | | |
| *All class 1* | 0.753 | 0.500 | 0.000 | 0.000 | – | 0.000 | 0.000 | 0.753 | 1.000 | 0.859 | 0 |
| RF | 0.888 | 0.808 | 0.682 | 0.671 | 0.860 | 0.651 | 0.741 | 0.894 | 0.965 | 0.928 | **758** |
| GTB | 0.892 | 0.840 | 0.701 | 0.700 | 0.807 | 0.737 | 0.770 | 0.916 | 0.942 | 0.929 | 1,263 |
| CART | 0.844 | 0.774 | 0.567 | 0.566 | 0.703 | 0.636 | 0.667 | 0.884 | 0.912 | 0.898 | 1,916 |
| SVM | 0.785 | 0.763 | 0.485 | 0.477 | 0.549 | 0.719 | 0.623 | 0.898 | 0.807 | 0.850 | 4,213 |
| KNN | 0.793 | 0.726 | 0.448 | 0.448 | 0.578 | 0.593 | 0.585 | 0.866 | 0.858 | 0.862 | 3,087 |
| NB | 0.723 | 0.771 | 0.469 | 0.421 | 0.466 | 0.866 | 0.606 | 0.939 | 0.676 | 0.786 | 7,079 |
| *Mean of the four sites* | | | | | | | | | | | |
| *All class 1* | 0.751 | 0.500 | 0.000 | 0.000 | – | 0.000 | 0.000 | 0.751 | 1.000 | 0.858 | 0 |
| RF | 0.852 | 0.726 | 0.563 | 0.512 | 0.907 | 0.471 | 0.575 | 0.850 | 0.982 | 0.910 | **360** |
| GTB | 0.856 | 0.755 | 0.591 | 0.559 | 0.847 | 0.547 | 0.635 | 0.865 | 0.962 | 0.910 | 680 |

| | | | | | Class 0 (lower HI) | | | Class 1 (higher HI) | | | |
|---|---|---|---|---|---|---|---|---|---|---|---|
| CART | 0.827 | 0.723 | 0.504 | 0.484 | 0.730 | 0.512 | 0.583 | 0.852 | 0.934 | 0.890 | 1,043 |
| SVM | 0.819 | 0.759 | 0.539 | 0.511 | 0.706 | 0.636 | 0.621 | 0.882 | 0.882 | 0.879 | 2,498 |
| KNN | 0.788 | 0.683 | 0.388 | 0.381 | 0.579 | 0.470 | 0.508 | 0.836 | 0.896 | 0.864 | 1,796 |
| NB | 0.746 | 0.712 | 0.402 | 0.369 | 0.522 | 0.638 | 0.525 | 0.877 | 0.786 | 0.820 | 3,993 |

RF achieved the highest accuracy at Site A (0.924) and 0.768–0.888 at the other sites (mean 0.852); its κ of 0.787 at Site A contrasts with 0.167 at Site B, where RF assigned almost all test pixels to the majority class. Tree ensembles are robust to noise, multicollinearity and nonlinear interactions among predictors (Rodriguez-Galiano et al. 2012; Maxwell et al. 2018). GTB achieved the highest mean accuracy (0.856) and κ (0.559) and the highest accuracy at Sites C (0.871) and D (0.892) and identified the minority lower-HI class better than RF at Sites B and C (class-0 recall 0.226 vs. 0.121 and 0.603 vs. 0.365). CART reached 0.780–0.844 (mean 0.827); single trees are interpretable but more sensitive to the training sample than ensembles (Quinlan 1993). KNN reached 0.727–0.860 (mean 0.788), in line with its sensitivity to feature scaling and mixed land cover (Cover and Hart 1967); boosting and single trees can also overfit heterogeneous feature distributions (Jerome H. Friedman 2001; Natekin and Knoll 2013). SVM was competitive at Sites A and B (0.922 and 0.807, the highest accuracy and κ at Site B) but weaker at Sites C and D (0.760 and 0.785; mean 0.819). SVMs suit high-dimensional classification (Cortes and Vapnik 1995) but are sensitive to parameter settings, class imbalance and correlated features (Mountrakis et al. 2011), and here the SVM was trained on a subsample of 10,000 pixels with fixed parameters. Naïve Bayes (NB) had the lowest mean accuracy (0.746) and κ (0.369), as expected from its assumption of feature independence (Langley et al. 1992), which is violated by correlated spectral, biophysical, soil and topographic predictors.

The tree ensembles (GTB and RF) were thus the most reliable classifiers of the HI classes, consistent with prior remote-sensing studies of vegetation and pollution (Belgiu and Drăguţ 2016; Maxwell et al. 2018); CART and SVM were competitive at individual sites, and KNN and NB were less suited to the heterogeneous sites. All models identified the majority higher-HI class more reliably than the lower-HI class, and the difference between the best models (0.004 in mean accuracy between GTB and RF) was small compared with the differences among sites.

### 4.5 Performance of the Random Forest model across study sites

After evaluating all six supervised machine learning models, the RF classifier was selected for the maps and for the feature-importance and constituent-level analyses because it produced the fewest false negatives of all six models at every site, i.e. it missed the fewest higher-HI pixels. RF produced on average 93, 7, 580 and 758 false negatives at Sites A–D, compared with 301 (GTB), 38 (SVM), 1,113 (GTB) and 1,263 (GTB) for the next best model, and 1,438 false negatives in total compared with 2,718 for GTB; its class-1 recall (0.965–0.999) was the highest at every site. For a screening tool, a missed higher-HI area is the more costly error, because it would not be prioritized for sampling. RF also achieved a mean accuracy within 0.004 of GTB and the highest accuracy at Site A. The price of this choice is a larger number of false positives, i.e. lower class-0 recall (0.471 on average, compared with 0.547 for GTB).

Overall accuracy ranged from 0.768 to 0.924 across the four sites (SD over seeds ≤ 0.003). Site A achieved the highest accuracy (0.924), followed by Site D (0.888), Site C (0.828) and Site B (0.768). Cohen's κ followed the same order, with values of 0.787, 0.671, 0.423 and 0.167 for Sites A, D, C and B. Accuracy must be read against the class proportions of the test sets: class 1 made up 73.7–75.6% of the test pixels, so a classifier that assigns every pixel to class 1 would reach an accuracy of 0.737–0.756. The RF accuracy exceeded this baseline by 0.181 at Site A and 0.135 at Site D, but only by 0.072 at Site C and 0.031 at Site B. Agreement beyond chance was therefore substantial at Sites A and D, moderate at Site C and slight at Site B. Balanced accuracy, which is 0.5 for a classifier that predicts only the majority class, ranged from 0.560 (Site B) to 0.866 (Site A) for RF, and MCC from 0.297 to 0.797 (Table 6). In balanced accuracy, SVM (mean 0.759) and GTB (0.755) exceeded RF (0.726), because RF assigned more lower-HI pixels to the majority class; RF was nevertheless retained because of its lower number of false negatives.

Class-specific metrics show that the differences among sites arise from the lower-HI class. For class 0, precision was high at every site (0.840–0.978), but recall ranged from 0.745 at Site A to 0.651 at Site D, 0.365 at Site C and 0.121 at Site B, with F1-scores of 0.835, 0.741, 0.509 and 0.215, respectively. High class-0 precision combined with low recall indicates that RF predicted class 0 only where the predictors were unambiguous and assigned most other lower-HI pixels to class 1. For class 1 (higher-HI conditions), recall was consistently high (0.965–0.999) and precision ranged from 0.761 (Site B) to 0.918 (Site A), giving F1-scores of 0.864–0.951. At Site B, a class-1 recall of 0.999 combined with a precision of 0.761, close to the class-1 share of 0.737, shows that nearly all pixels were predicted as class 1.

Site A produced on average 1,765 true negatives (TN), 603 false positives (FP), 93 false negatives (FN) and 6,755 true positives (TP), i.e. 696 misclassified pixels of 9,216 (7.6%). Site B produced 305 TN, 2,224 FP, 7 FN and 7,064 TP: 88% of its 2,529 class-0 test pixels were classified as class 1, whereas only 7 of 7,071 class-1 pixels were missed. Site C produced 3,050 TN, 5,312 FP, 580 FN and 25,298 TP. Of its 5,892 misclassified pixels, 90% were false positives, which explains the low class-0 recall (0.365) and the moderate κ (0.423). Site D showed an intermediate pattern, with 4,651 TN, 2,489 FP, 758 FN and 21,062 TP (3,247 misclassified pixels, 77% of them false positives). False positives exceeded false negatives at every site; in total, the four test sets contained 10,628 false positives and 1,438 false negatives. This asymmetry is reflected in the consistently high recall for class 1 (≥ 0.965) and the lower recall for class 0. Thus, the RF classifier rarely missed higher-HI reference pixels but frequently assigned lower-HI pixels to the higher-HI class; this tendency was strongest at Sites B and C.

RF thus captured relationships between the HI classes and the predictors at Sites A and D, and to a lesser extent at Site C, but added little beyond the majority class at Site B. It was highly sensitive to the higher-HI class at all sites, whereas the lower-HI areas were poorly separated at two of the four sites. The classes represent vegetation and environmental patterns associated with lower and higher HI, not metal concentrations or direct evidence of metal-induced physiological stress.

### 4.6 Spatial distribution of composite HI and Random Forest classification

The composite HI varied strongly within and among the sites (Figure 8). Before normalization, HI ranged from 1.70 to 23.7 at Site A, 1.38 to 33.51 at Site B, 1.10 to 59.15 at Site C and 1.21 to 13.79 at Site D; Site C had the highest maximum and widest range, and Site D the lowest maximum and narrowest range.

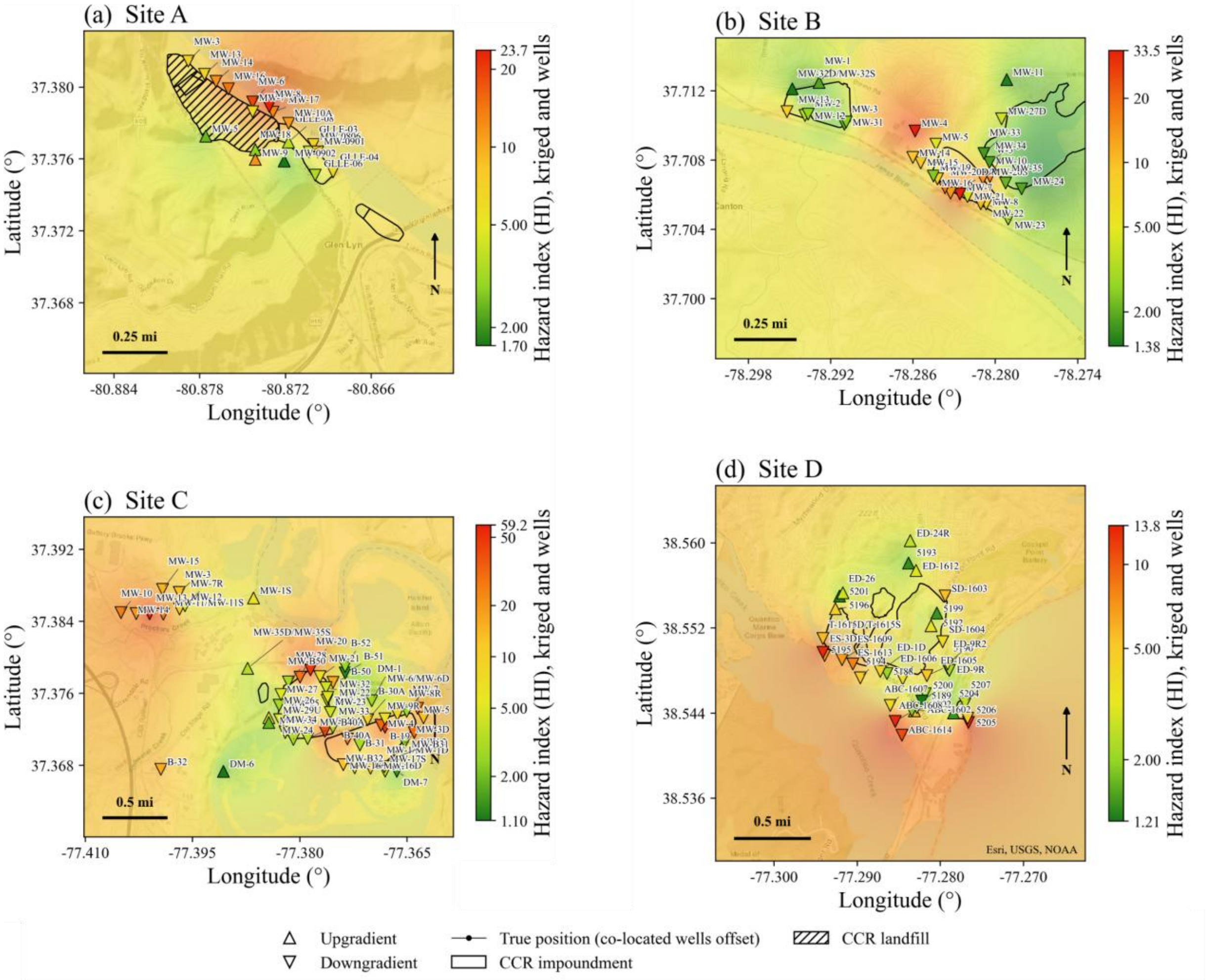

Figure 8: Hazard index (HI) of the groundwater monitoring wells at (a) Site A, (b) Site B, (c) Site C and (d) Site D. For each well, HI is the sum of the hazard quotients (HQ = constituent concentration / screening threshold) of the constituents measured there.

The normalized HI surfaces showed distinct gradients and localized areas of elevated burden at all four sites (Figure 9). At Site A, higher HI was concentrated in one part of the site, and the binary reference map contained a distinct lower-HI region surrounded by higher HI; RF reproduced this configuration, with disagreement mainly at the transition between the classes.

At Site B, a localized zone of high HI was surrounded by progressively lower values, and the lower-HI class was concentrated in one part of the site. RF recovered only a small part of this region (class-0 recall 0.121) and predicted most of it as higher HI. Site C showed the greatest heterogeneity: besides the highest maximum and widest range of HI, its surface contained several gradients and local maxima, and the reference map had several small lower-HI areas within a predominantly higher-HI landscape. RF reproduced the broad distribution of the higher-HI class but disagreed around the small lower-HI areas. At Site D, the HI range was narrower, but the normalized surface still showed clear within-site variation, with a coherent lower-HI region within a broader higher-HI area that RF reproduced, again with disagreement mainly at the transition between the classes.

The RF probability surfaces show the continuous probability of the higher-HI class. High and low probabilities matched the predicted classes, and intermediate probabilities occurred mainly in transition zones, where the predictors were less distinctly associated with either class.

The error maps show that disagreement was spatially structured. FP and FN pixels occurred mainly near the boundaries between the reference classes, particularly at Site C, where the classes were fragmented. Errors were localized at Site A and intermediate at Site D, whereas at Site B the false positives covered most of the lower-HI reference region.

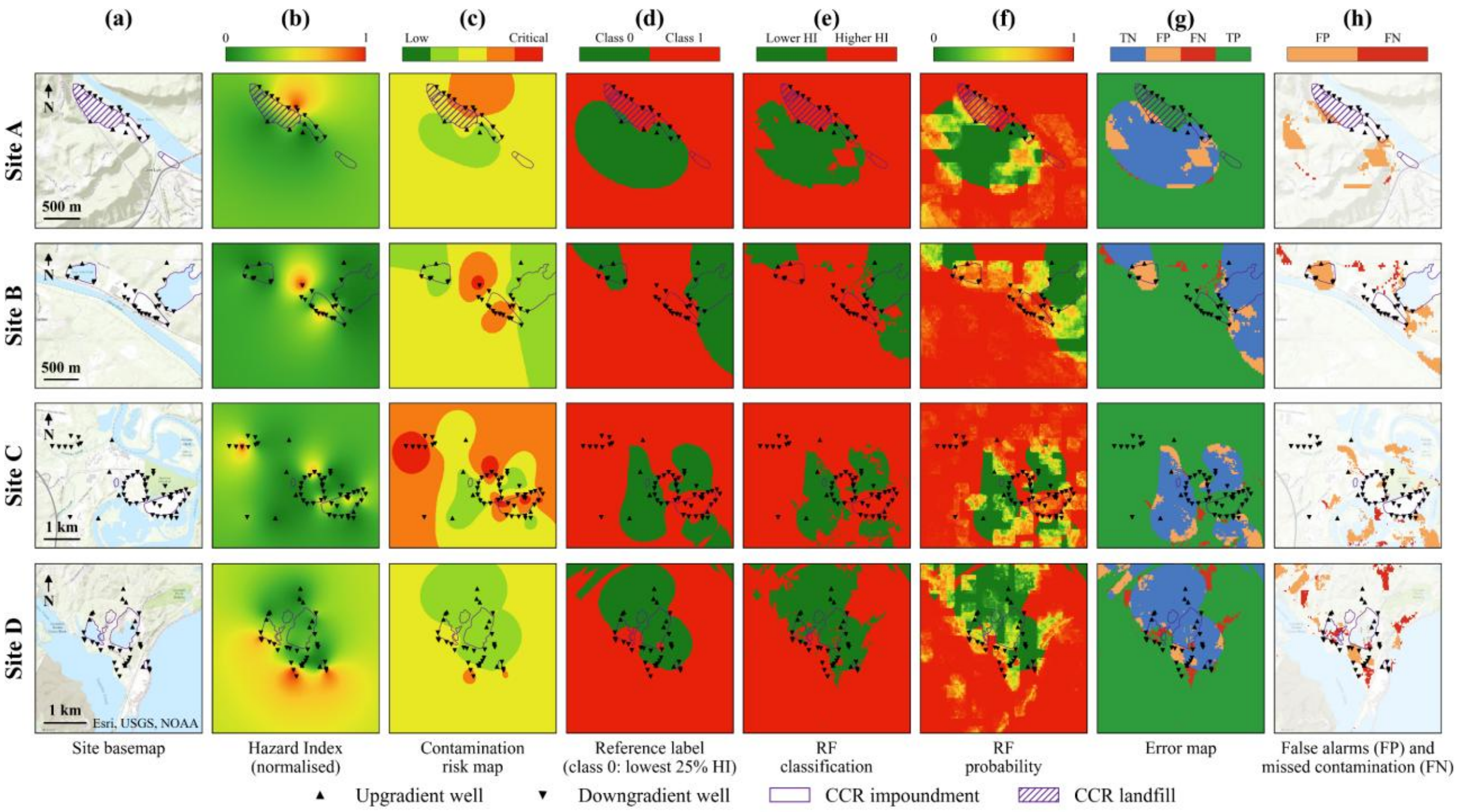

Figure 9: Spatial distribution of composite Hazard Index (HI) and Random Forest (RF) classification outputs across four CCR study sites. Each row represents one study site. From left to right: (a) site basemap; (b) kriged composite HI normalized to 0–1, representing relative within-site cumulative metal burden; (c) categorical HI risk map ranging from Low to Critical; (d) binary HI-derived reference classes representing lower-HI (class 0, green) and higher-HI (class 1, red) conditions; (e) RF-predicted classes; (f) predicted probability of the higher-HI class; (g) spatial classification outcomes as true negative (TN), false positive (FP), false negative (FN), and true positive (TP); and (h) false-positive and false-negative locations overlaid on the satellite basemap.

Overall, Figure 9 shows the progression from well HI to the classified maps. The agreement between the reference classes and the RF predictions shows that the predictor space contains information associated with the spatial variation of HI; the outputs represent vegetation and environmental characteristics associated with lower and higher HI, not measured metal concentrations.

## 4.7 Feature importance and environmental controls on HI classification

Impurity importance, group permutation importance and SHAP values were used to quantify the contribution of each predictor group to the RF classification (Figure 10 and Figure 11). Importance varied among sites, and no single predictor or group controlled the classification at all sites, consistent with the use of RF for high-dimensional, multi-source data with nonlinear interactions among correlated predictors (Gislason et al. 2006; Belgiu and Drăguţ 2016).

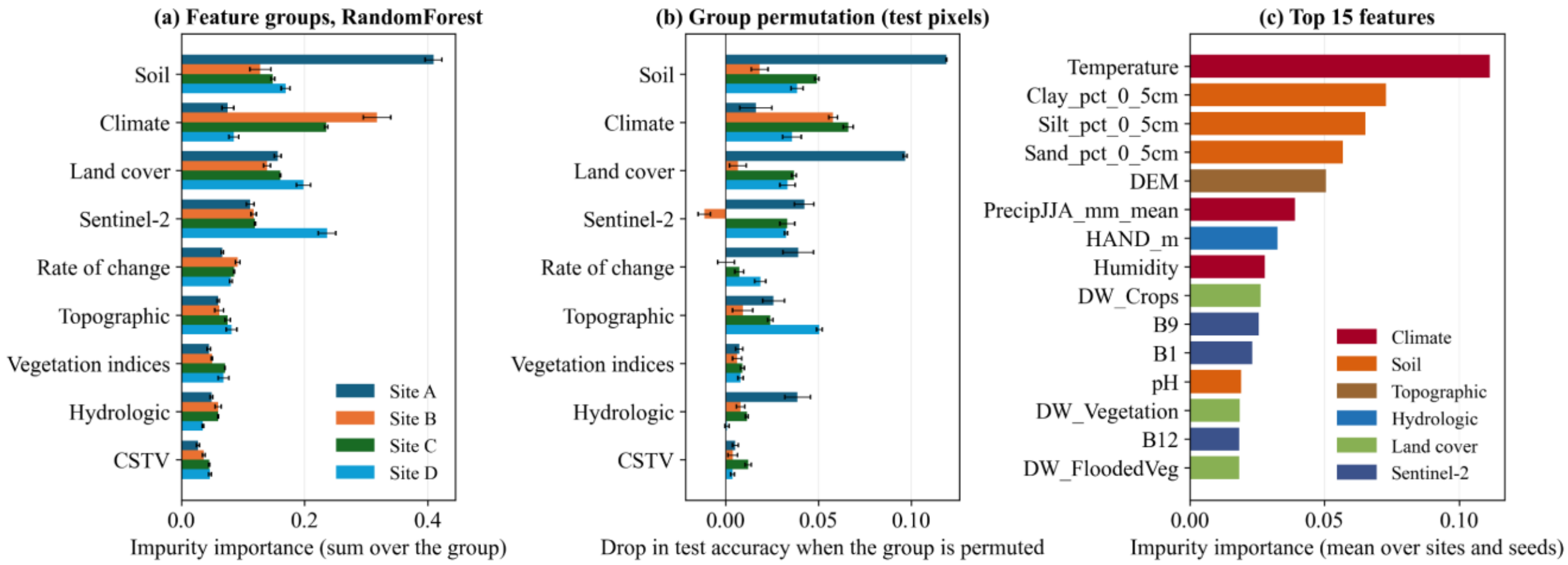

Figure 10: Random Forest feature importance across the four CCR study sites (mean of three seeds). (a) Impurity importance summed over the features of each group; (b) decrease in test accuracy when all features of a group are permuted jointly; error bars show the SD over seeds. (c) The 15 individual features with the highest impurity importance, averaged over sites and seeds and coloured by group.

The individual-feature analysis (Figure 10) showed that the most important individual features were coarse-resolution environmental covariates rather than vegetation indices. By impurity importance, soil texture dominated at Site A (silt 0.165, clay 0.127 and sand 0.090), land surface temperature at Site B (0.269) and Site C (0.104, followed by humidity 0.073 and elevation 0.062), and clay content at Site D (0.103, followed by the Dynamic World crop probability 0.066 and elevation 0.055). Elevation ranked third or fourth at every site. The 15 most important features over all sites (Figure 10c) were climate, soil, topographic, hydrologic, Sentinel-2 and land-cover variables; no vegetation index, CSTV or rate-of-change feature was among them.

The group-level importance (Figure 10a, b) and the SHAP analysis (Figure 11) show the same site-specific differences. At Site A, soil was the most important group by all three measures (impurity 0.410, permutation decrease in accuracy 0.119, summed mean |SHAP| 0.188), followed by land cover (0.157, 0.097 and 0.111). At Site B, climate dominated (0.318, 0.058 and 0.192), driven almost entirely by land surface temperature (mean |SHAP| 0.172), followed by land cover and soil. At Site C, climate was again the leading group (0.235, 0.066 and 0.140), followed by land cover and soil by impurity and by soil and land cover by permutation and SHAP. At Site D, Sentinel-2 bands were the most important group by impurity (0.236) and second by SHAP (0.111, after land cover with 0.114), with the shortwave-infrared bands B11 and B12 among the leading features, whereas topographic variables caused the largest permutation decrease in accuracy (0.051). The vegetation-related groups contributed little at every site. Vegetation indices and biophysical parameters accounted for 0.025–0.028 of the summed mean |SHAP|, CSTV for 0.015–0.025 and rate-of-change metrics for 0.038–0.046; together they made up 13.8–17.4% of the total SHAP importance, and vegetation indices and CSTV alone 7.1–10.0%. Their permutation decreases in accuracy did not exceed 0.012, except for the rate-of-change metrics at Site A (0.039). Multi-year Sentinel-2 vegetation metrics have been used to monitor vegetation condition and dynamics (Karim et al. 2026; Mulualem et al. 2026), but in the present feature space their contribution was small relative to that of the environmental covariates.

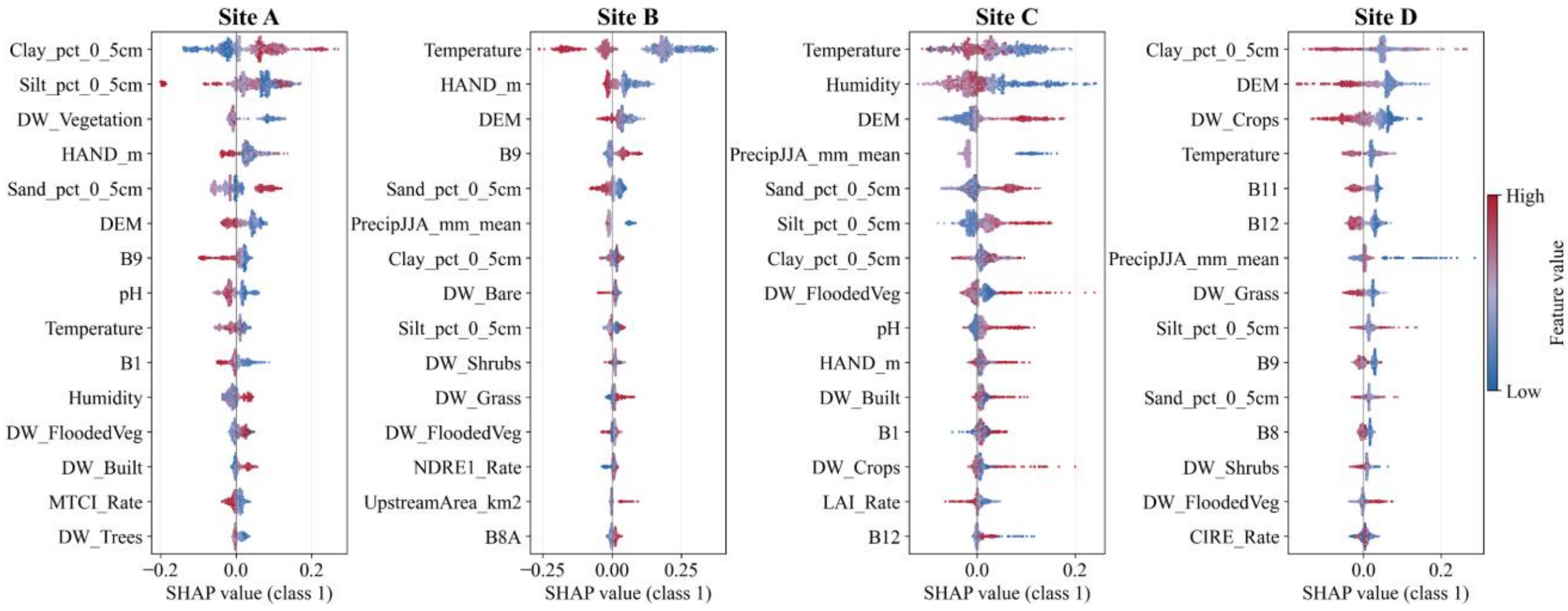

Figure 11: SHAP summary of the Random Forest classifier at Sites A–D, computed with TreeSHAP for 1,000 held-out test pixels per site. Each point is one pixel; its horizontal position is the SHAP value of the feature (positive values increase the predicted probability of the higher-HI class) and its colour the feature value (low to high). Features are ordered by mean absolute SHAP value; the 15 most important features are shown.

Land-cover probabilities ranked second at every site by impurity importance (0.139–0.198) and among the three most important groups by SHAP. This indicates that spatial differences in vegetation and surface-cover composition were associated with the HI-derived classes and highlights the need to account for surface type when interpreting remotely sensed vegetation responses.

The Sentinel-2 red-edge region is sensitive to chlorophyll and canopy characteristics, and the NIR and SWIR regions to vegetation structure and water status (Clevers and Gitelson 2013), although the sensitivity of vegetation indices varies with their formulation, canopy properties and environmental conditions (Yan et al. 2025; Karim et al. 2026). The weak contribution of the vegetation indices in this study therefore does not show that they are insensitive to metal stress; it indicates that, at 20 m resolution and with a target derived from interpolated well data, the vegetation signal was weaker than the large-scale environmental gradients that co-varied with the kriged HI.

The dominance of soil, climate, topography, hydrology and land cover is central to the interpretation of the RF predictions. Vegetation spectra respond to many environmental controls unrelated to metals, so the model captured a combined vegetation–environment signature associated with HI rather than groundwater contaminants. Several of these covariates have a much coarser resolution than the 20 m grid (soil 250 m, land surface temperature 1 km, NLDAS humidity about 12 km) and vary smoothly across a site, as does the kriged HI. Within a site, such smooth covariates can act as proxies for location and reproduce a smooth target surface without a causal relation to groundwater chemistry (Ploton et al. 2020). Soil moisture and, at Sites B and D, humidity were constant within the site and received no SHAP importance.

RF importance measures also quantify predictive contribution within the fitted model, not causality, and Strobl et al. (2007) showed that they can be biased under particular predictor structures, which is relevant here because spectral bands, vegetation indices, biophysical parameters and their temporal derivatives are interrelated.

The feature-importance analysis demonstrates that classification of the HI-derived classes was multivariate and site specific but dominated by environmental covariates. Soil texture led at Site A, climate at Sites B and C, and Sentinel-2 reflectance, land cover and soil at Site D, whereas vegetation indices, biophysical parameters, CSTV and rate-of-change metrics contributed least at all sites for the RF classification model. The remotely sensed classification therefore represents vegetation and environmental conditions associated with composite HI, rather than direct detection of metals or attribution of vegetation change to a single contaminant.

### 4.8 Environmental and vegetation feature-set ablation

To test the feature-importance results directly, RF was retrained on the same spatial blocks and seeds with three reduced predictor sets: vegetation only, environment only, and the distance to the CCR unit as a location-only baseline (Table 7). The higher-HI class makes up about 75% of the test pixels at every site. Cohen's κ and the F1 score of the lower-HI class therefore separate the models better than accuracy does.

Table 7: Random Forest test performance with different predictor sets and a location-only baseline (held-out test blocks; mean of three seeds). κ: Cohen's kappa; precision; recall; F1: F1 score. Majority-class accuracy: accuracy when every pixel is predicted as higher HI. Precision and F1 were derived from the mean class recalls and the test class shares.

| | | | Class 0 (lower HI) | | | Class 1 (higher HI) | | |
|---|---|---|---|---|---|---|---|---|
| Predictor set | Accuracy | κ | Precision | Recall | F1 | Precision | Recall | F1 |
| ***Site A*** *(majority-class accuracy 0.743; n test = 9,216)* | | | | | | | | |
| All features | 0.924 | 0.787 | 0.950 | 0.745 | 0.835 | 0.918 | 0.986 | 0.951 |
| Vegetation only | 0.894 | 0.698 | 0.896 | 0.666 | 0.764 | 0.894 | 0.973 | 0.932 |
| Environment only | 0.918 | 0.778 | 0.872 | 0.796 | 0.832 | 0.931 | 0.960 | 0.945 |
| Distance to CCR unit | 0.650 | 0.082 | 0.318 | 0.316 | 0.317 | 0.764 | 0.765 | 0.765 |
| ***Site B*** *(majority-class accuracy 0.737; n test = 9,600)* | | | | | | | | |
| All features | 0.768 | 0.167 | 0.980 | 0.121 | 0.215 | 0.761 | 0.999 | 0.864 |
| Vegetation only | 0.721 | 0.021 | 0.331 | 0.057 | 0.097 | 0.740 | 0.959 | 0.835 |
| Environment only | 0.800 | 0.329 | 0.931 | 0.260 | 0.406 | 0.790 | 0.993 | 0.880 |
| Distance to CCR unit | 0.787 | 0.456 | 0.594 | 0.609 | 0.601 | 0.859 | 0.851 | 0.855 |
| ***Site C*** *(majority-class accuracy 0.756; n test = 34,240)* | | | | | | | | |
| All features | 0.828 | 0.423 | 0.840 | 0.365 | 0.509 | 0.826 | 0.978 | 0.896 |
| Vegetation only | 0.764 | 0.198 | 0.541 | 0.218 | 0.310 | 0.788 | 0.940 | 0.858 |
| Environment only | 0.853 | 0.535 | 0.841 | 0.489 | 0.618 | 0.855 | 0.970 | 0.909 |
| Distance to CCR unit | 0.659 | 0.163 | 0.348 | 0.453 | 0.394 | 0.804 | 0.726 | 0.763 |
| ***Site D*** *(majority-class accuracy 0.753; n test = 28,960)* | | | | | | | | |
| All features | 0.888 | 0.671 | 0.860 | 0.651 | 0.741 | 0.894 | 0.965 | 0.928 |
| Vegetation only | 0.844 | 0.549 | 0.729 | 0.583 | 0.648 | 0.872 | 0.929 | 0.900 |
| Environment only | 0.884 | 0.670 | 0.815 | 0.684 | 0.744 | 0.902 | 0.949 | 0.925 |
| Distance to CCR unit | 0.718 | 0.262 | 0.433 | 0.470 | 0.451 | 0.822 | 0.799 | 0.810 |

Removing the vegetation indices did not reduce performance. The environment-only model matched the full model at Sites A and D (κ = 0.778 vs 0.787 and 0.670 vs 0.671). At Sites B and C, it exceeded the full model (κ = 0.329 vs 0.167 and 0.535 vs 0.423).

The vegetation-only model was clearly more accurate than the majority-class baseline at Sites A and D (accuracy 0.894 and 0.844 vs 0.743 and 0.754; κ = 0.698 and 0.549). At Site C it was close to the baseline (κ = 0.198). At Site B it was below the baseline and had no skill (κ = 0.021; lower-HI F1 = 0.097).

The distance to the CCR unit alone was much weaker than the full model at Sites A, C and D (κ = 0.082–0.262). At Site B, however, it outperformed every remotely sensed and environmental predictor set (κ = 0.456 vs 0.021–0.329), so the Site B classification mainly reflects proximity to the CCR unit.

Overall, the environmental covariates carry most of the predictive information. The vegetation predictors add little beyond them at the pixel scale, even though a vegetation response is detectable near the monitoring wells.

### 4.9 Constituent-level evaluation of the Random Forest model across all sites

The RF model was also applied to every metal that exceeded its reference threshold (HQ ≥ 1) at one or more wells of a site: seven metals at Site A (Mn, As, Mo, Sr, B, Co and Pb), eight at Site B (Mn, Mo, As, Co, B, Li, Pb and Hg), twelve at Site C (Co, As, B, Mn, Ra-226+228, Zn, Be, Li, Mo, Cr, Pb and Tl) and seven at Site D (Co, Mn, As, Ni, Ra-226+228, B and Cd). For each metal, the kriged well HQ was normalized within the site and converted to binary classes at its 25th percentile, and RF models were trained and tested with the same spatial partition and seeds as the HI models.

Figure 12 shows the HQ, normalized HQ, RF predictions and classification errors for the seven metals at Site A. Figures S1-S3 for Site B, C and D are added in Supplementary material. The metals had distinct spatial patterns; RF reproduced the main structure of each reference map, with disagreement mainly at the transitions between lower- and higher-HQ regions.

The constituent-level models (Table 8) achieved accuracies of 0.775–0.989 (mean 0.868) and κ of 0.281–0.969 (mean 0.612) over the 34 metal–site combinations. The mean accuracy per site was 0.866 at Site A, 0.873 at Site B, 0.853 at Site C and 0.890 at Site D. The highest agreement was obtained for Pb (accuracy 0.989, κ 0.969) and Hg (0.957, 0.879) at Site B, where a single exceeding well produces a compact high-HQ zone, and the lowest κ for Mo at Site C (0.281), Li at Site C (0.364) and Co at Site A (0.383). At Site A, accuracy ranged from 0.790 (Co) to 0.913 (Sr) and κ from 0.383 (Co) to 0.727 (Sr and B).

Notably, at Site B, where the composite-HI model reached a κ of only 0.167, κ for the individual metals ranged from 0.391 to 0.969. Class-specific performance again showed high sensitivity to the higher-HQ class: class-1 recall ranged from 0.804 (Co, Site B) to 1.000 (Sr, Site A; mean 0.951), with F1-scores of 0.841–0.992. Recall for the lower-HQ class was lower and more variable, from 0.238 (Mo, Site C) to 0.968 (Pb, Site B; mean 0.619). Class-0 recall was below 0.5 for eight of the 34 combinations, all with accuracy of 0.839 or less.

Table 8: Random Forest performance for constituent-level classification of the normalized HQ of each metal exceeding its reference threshold at Sites A–D (held-out test blocks; mean of three seeds; SD over seeds ≤ 0.009 for accuracy, ≤ 0.022 for κ and ≤ 0.039 for the class-wise metrics). Wells HQ ≥ 1: wells exceeding the threshold / wells with the metal. Class 0 (lower HQ): lowest 25% of the normalized HQ. P: precision; R: recall; F1: F1 score.

| | | | | Class 0 (lower HQ) | | | Class 1 (higher HQ) | | |
|---|---|---|---|---|---|---|---|---|---|
| Metal | Wells HQ ≥ 1 | Accuracy | κ | P | R | F1 | P | R | F1 |
| *Site A* | | | | | | | | | |
| Mn | 11/19 | 0.901 | 0.693 | 0.961 | 0.617 | 0.751 | 0.891 | 0.992 | 0.939 |
| As | 5/19 | 0.810 | 0.440 | 0.634 | 0.502 | 0.560 | 0.851 | 0.908 | 0.878 |
| Mo | 5/19 | 0.889 | 0.674 | 0.921 | 0.621 | 0.742 | 0.882 | 0.982 | 0.929 |
| Sr | 3/19 | 0.913 | 0.727 | 1.000 | 0.636 | 0.777 | 0.898 | 1.000 | 0.946 |
| B | 2/19 | 0.905 | 0.727 | 0.916 | 0.688 | 0.786 | 0.903 | 0.979 | 0.939 |
| Co | 2/19 | 0.790 | 0.383 | 0.615 | 0.439 | 0.512 | 0.828 | 0.908 | 0.866 |
| Pb | 2/19 | 0.857 | 0.596 | 0.767 | 0.624 | 0.688 | 0.880 | 0.936 | 0.907 |
| *Site B* | | | | | | | | | |
| Mn | 14/16 | 0.798 | 0.391 | 0.619 | 0.442 | 0.515 | 0.836 | 0.912 | 0.872 |
| Mo | 4/28 | 0.830 | 0.417 | 0.996 | 0.324 | 0.489 | 0.815 | 1.000 | 0.898 |
| As | 3/28 | 0.922 | 0.784 | 0.892 | 0.785 | 0.835 | 0.931 | 0.968 | 0.949 |
| Co | 3/16 | 0.775 | 0.457 | 0.551 | 0.691 | 0.613 | 0.882 | 0.804 | 0.841 |
| B | 2/23 | 0.866 | 0.677 | 0.681 | 0.883 | 0.769 | 0.956 | 0.861 | 0.906 |
| Li | 2/16 | 0.848 | 0.590 | 0.687 | 0.695 | 0.691 | 0.901 | 0.897 | 0.899 |
| Pb | 1/28 | 0.989 | 0.969 | 0.986 | 0.968 | 0.977 | 0.989 | 0.995 | 0.992 |
| Hg | 1/28 | 0.957 | 0.879 | 0.975 | 0.849 | 0.907 | 0.951 | 0.993 | 0.972 |
| *Site C* | | | | | | | | | |
| Co | 22/39 | 0.857 | 0.590 | 0.779 | 0.605 | 0.681 | 0.876 | 0.942 | 0.908 |
| As | 19/53 | 0.839 | 0.459 | 0.930 | 0.377 | 0.536 | 0.829 | 0.991 | 0.903 |
| B | 12/39 | 0.906 | 0.726 | 0.922 | 0.682 | 0.784 | 0.903 | 0.981 | 0.940 |
| Mn | 10/14 | 0.856 | 0.598 | 0.767 | 0.628 | 0.691 | 0.880 | 0.935 | 0.906 |
| Ra | 9/39 | 0.878 | 0.648 | 0.825 | 0.647 | 0.725 | 0.891 | 0.954 | 0.922 |
| Zn | 9/23 | 0.863 | 0.578 | 0.836 | 0.544 | 0.659 | 0.869 | 0.966 | 0.915 |
| Be | 6/39 | 0.885 | 0.668 | 0.842 | 0.660 | 0.740 | 0.896 | 0.959 | 0.926 |
| Li | 6/39 | 0.792 | 0.364 | 0.645 | 0.388 | 0.485 | 0.818 | 0.928 | 0.870 |
| Mo | 3/53 | 0.793 | 0.281 | 0.780 | 0.238 | 0.365 | 0.794 | 0.978 | 0.876 |
| Cr | 1/53 | 0.876 | 0.625 | 0.880 | 0.579 | 0.698 | 0.875 | 0.974 | 0.922 |
| Pb | 1/53 | 0.855 | 0.613 | 0.704 | 0.714 | 0.709 | 0.905 | 0.901 | 0.903 |
| Tl | 1/39 | 0.832 | 0.473 | 0.748 | 0.461 | 0.570 | 0.846 | 0.950 | 0.895 |
| *Site D* | | | | | | | | | |
| Co | 10/18 | 0.887 | 0.693 | 0.815 | 0.724 | 0.767 | 0.908 | 0.943 | 0.925 |
| Mn | 6/17 | 0.908 | 0.733 | 0.941 | 0.682 | 0.791 | 0.900 | 0.985 | 0.941 |
| As | 3/35 | 0.815 | 0.413 | 0.752 | 0.390 | 0.513 | 0.824 | 0.957 | 0.886 |
| Ni | 3/17 | 0.903 | 0.710 | 0.958 | 0.641 | 0.768 | 0.892 | 0.991 | 0.939 |
| Ra | 2/18 | 0.924 | 0.790 | 0.875 | 0.808 | 0.840 | 0.939 | 0.962 | 0.950 |
| B | 1/18 | 0.908 | 0.742 | 0.862 | 0.750 | 0.802 | 0.920 | 0.960 | 0.940 |
| Cd | 1/35 | 0.883 | 0.687 | 0.775 | 0.755 | 0.765 | 0.918 | 0.926 | 0.922 |

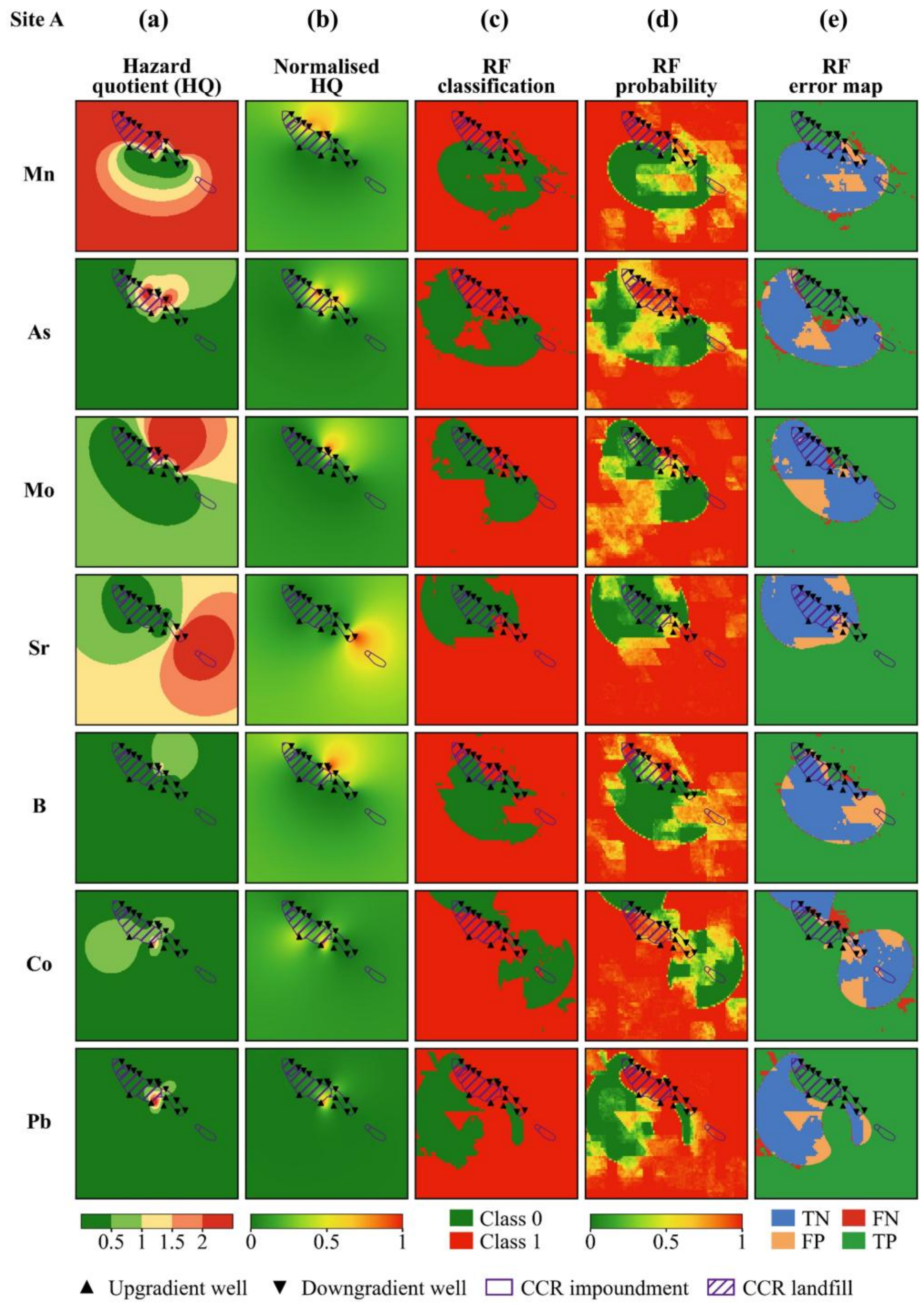

Figure 12: Constituent-level Random Forest classification at Site A for the seven metals with HQ ≥ 1 (Mn, As, Mo, Sr, B, Co and Pb). Each row shows one metal; the columns show (a) the kriged Hazard Quotient (HQ), (b) the normalized HQ, (c) the RF classification, (d) the RF probability of the higher-HQ class and (e) the classification outcome (TN, FP, FN, TP). The corresponding figures for Sites B–D are provided in the Supplementary Material. Class 0 and Class 1 represent lower- and higher-HQ conditions, respectively.

RF thus separated the normalized HQ classes of individual metals at all four sites, with performance comparable to or higher than for the composite HI, and the metals had distinct spatial signatures. Because each metal map is derived from the same well network and kriging procedure, the same caveat applies as for HI: the classes reflect the spatial structure of the interpolated surfaces, and their agreement with the remotely sensed predictors does not establish a physiological response to individual metals.

### 4.10 Cross-Site transferability and spatial generalization

In the leave-one-site-out experiment, models were trained on three sites and tested on the fourth, which contributed no training data (Table 9 & The held-out RF accuracy (0.734–0.749) did not exceed the share of the majority class in the test samples (0.739–0.756), except marginally at Site B (0.749 vs 0.748). Recall for the higher-HI class was 0.979–0.999, whereas recall for the lower-HI class was 0.000–0.027. Of 12,198–13,031 lower-HI test pixels per site, only 5 (Site A) to 332 (Site D) were classified correctly.

Table 10). Leave-one-site-out testing lowered the performance of every classifier at every site (Figure S5). For RF: accuracy fell from 0.77–0.92 within sites to 0.73–0.75 at the held-out sites; macro F1 fell from 0.54–0.89 to 0.42–0.45.

Table 9 : Cross-site transfer performance of the RF model using leave-one-site-out training (mean of three seeds; 50,000 test pixels per held-out site).

| | | | Class 0 (lower HI) | | | Class 1 (higher HI) | | |
|---|---|---|---|---|---|---|---|---|
| Target site | Training sites | Accuracy | Precision | Recall | F1 | Precision | Recall | F1 |
| Site A | Sites B, C, D | 0.734 | 0.023 | 0.000 | 0.001 | 0.738 | 0.992 | 0.846 |
| Site B | Sites A, C, D | 0.749 | 0.665 | 0.007 | 0.013 | 0.749 | 0.999 | 0.856 |
| Site C | Sites A, B, D | 0.742 | 0.093 | 0.008 | 0.014 | 0.753 | 0.979 | 0.851 |
| Site D | Sites A, B, C | 0.745 | 0.301 | 0.027 | 0.050 | 0.755 | 0.980 | 0.853 |

Table 10: Confusion matrices for leave-one-site-out RF transfer testing (mean counts of three seeds).

| Target site | TN | FP | FN | TP | Test N |
|---|---|---|---|---|---|
| Site A | 5 | 13,026 | 279 | 36,690 | 50,000 |
| Site B | 86 | 12,515 | 50 | 37,349 | 50,000 |
| Site C | 96 | 12,102 | 810 | 36,992 | 50,000 |
| Site D | 332 | 11,967 | 766 | 36,935 | 50,000 |

This failure is consistent with the site-specific feature importance. Differences among sites in land cover, vegetation, soil, topography, hydrology, climate and HI distribution change the relationships between predictors and classes, a geographic domain shift that is a recognized challenge when remote-sensing models are transferred between regions (Meyer and Pebesma 2021) . Here, the dominance of coarse soil and climate covariates contributes directly to the shift: soil texture, land surface temperature and humidity vary smoothly within each site but differ in level and range among sites, so the relationships learned at the training sites do not apply to the held-out site, even after per-site standardization.

Strong within-site performance therefore does not imply performance at an independent site; the models learn, at least partly, site-specific relationships. Poor transfer does not mean that vegetation–HI associations are absent at the target site, only that the relationships learned at three sites are not invariant. Broader multi-site training data, or a few target-site observations for local calibration, may be needed for generalization.

### 4.11 Limitations and Implications

While the results demonstrate considerable potential for detecting leached-metal contamination hazard using integrated multispectral vegetation metrics, environmental covariates, and supervised machine learning, a few limitations should be acknowledged. First, although groundwater monitoring records span 2010–2019 while Sentinel-2 imagery begins in 2019, creating a partial temporal mismatch, this does not compromise the overall validity of the approach. Leachate metals from CCR migrate slowly and persist in groundwater and soils because they continue to leach from unlined impoundments and do not readily degrade (Adriano et al. 2004; Bolan et al. 2014). Prior research consistently shows that these contaminants remain stable for long periods, forming gradually moving plumes that can persist in subsurface environments for years to decades (Ruhl et al. 2012; Brandt et al. 2017; Petrović and Fiket 2022). Thus, the kriged concentration maps used here represent chronic contamination conditions appropriately aligned with vegetation stress signals captured during 2019–2025. A second limitation concerns the reliance on interpolated surfaces rather than dense point-based sampling, which may smooth fine-scale heterogeneity in contamination patterns. Third, the cross-site transfer analysis demonstrated substantial limitations in geographic generalization. When models trained on three sites were applied to the excluded fourth site, overall accuracy decreased below the corresponding within-site performance. The site-specific feature-importance analysis similarly showed that different combinations of soil, land-cover, climatic, hydrological, topographic, spectral, biophysical, and temporal variables contributed to classification at different sites. These findings indicate that environmental heterogeneity can produce geographic domain shift, limiting the transferability of relationships learned at one set of locations to an environmentally distinct site. Accordingly, application of the framework to new CCR sites may require geographically diverse training data, local calibration, or domain-adaptation approaches rather than direct application of a model trained elsewhere. Fourth, the framework does not explicitly incorporate vegetation species, community composition, phenological differences, or species-specific physiological responses. Vegetation responses to leaching metal exposure can vary according to species, contaminant type and concentration, soil properties, water availability, nutrient status, and other environmental stresses (Nagajyoti et al. 2010; Emamverdian et al. 2015). Consequently, similar spectral or biophysical responses may arise from multiple environmental stressors, while some metal-related physiological changes may not produce detectable canopy-level reflectance responses. The remotely sensed classes should therefore be interpreted as vegetation and environmental conditions associated with lower or higher HI, rather than as direct observations of metal-induced physiological stress. Additional uncertainty arises from the limited hydrogeological information available for the study sites. Additional limitations include the lack of hydrogeological details in publicly available datasets (e.g., aquifer stratigraphy, hydraulic head and conductivity).

Despite these limitations, the study has important implications for CCR environmental monitoring and spatial screening. The results demonstrate that combining groundwater-derived composite HI with Sentinel-2 spectral information, vegetation indices, biophysical parameters, temporal-change metrics, and environmental covariates can identify spatial vegetation–environment patterns and hazard associated with differences in cumulative hazard from CCR leachates. Importantly, these associations should not be interpreted as evidence that satellite imagery directly detects dissolved metals or that groundwater contamination is the sole cause of the observed vegetation patterns.

The comparison among supervised classification algorithms further demonstrated the value of integrating multiple predictor domains. Tree-based models, particularly RF and GTB, provided strong within-site discrimination of the HI-derived classes, while feature-importance analysis showed that classification depended on combinations of vegetation, spectral, temporal, land-cover, soil, climatic, hydrological, and topographic information. The contribution of these environmental covariates is particularly important because it demonstrates that vegetation condition cannot be interpreted solely from individual vegetation indices such as NDVI or CIRE.

At the same time, the leave-one-site-out analysis establishes an important practical boundary for implementation. Strong performance within individual sites did not translate directly to geographically independent sites, indicating that a universally transferable classifier cannot yet be inferred from the four-site analysis. Future development should therefore emphasize multi-site calibration, independent-site validation, domain adaptation, and integration of hydrogeological information to improve geographic generalization.

The proposed framework offers a scalable, low-cost, and non-invasive alternative to traditional point-based monitoring, and can be considered a screening and decision-support approach that complements, rather than replaces, conventional groundwater monitoring. Satellite observations provide repeated and spatially continuous information over areas that cannot practically be covered by dense monitoring-well networks. When integrated with groundwater measurements, the framework could help identify areas exhibiting vegetation–environment characteristics associated with elevated HI, prioritize locations for additional groundwater, soil, or vegetation sampling, and support evaluation of spatial and temporal changes around CCR facilities. Such targeted use could improve the spatial efficiency of field investigations while retaining direct environmental measurements as the basis for confirmation and regulatory assessment.

Collectively, the findings demonstrate the potential of integrating groundwater monitoring, geospatial interpolation, multispectral remote sensing, temporal vegetation analysis, environmental covariates, and supervised machine learning for large-area assessment of CCR sites. Rather than providing direct detection of leached metals, the framework establishes a pathway for identifying remotely observable vegetation and environmental patterns associated with spatial variation in leaching metal contamination. With additional field validation, hydrogeological integration, and improved cross-site generalization, this approach could provide a complementary tool for long-term CCR surveillance, targeted environmental sampling, remediation monitoring, and prioritization of areas requiring further investigation.

## 5 Conclusion

This study combined groundwater monitoring with satellite and geospatial data to identify vegetation and environmental patterns associated with leached-metal hazard at four CCR sites. Concentrations of multiple metals/metalloids were standardized as Hazard Quotients (HQ), summed into a composite Hazard Index (HI) and related to multi-temporal Sentinel-2 observations, vegetation indices, biophysical parameters, CSTV and rate-of-change metrics, and soil, climate, hydrology, topography and land-cover variables.

Maximum composite HI ranged from 13.79 to 59.15 among the sites, and constituent composition differed, so a composite HI represented groundwater conditions more completely than a single metal. Near the monitoring wells, higher HQ of Mn, As, Co and B was associated with lower values of all nine vegetation indices for 14 of the 16 metal–site combinations (mean Spearman $\rho$ down to −0.60), supporting the hypothesis that groundwater metal hazard is expressed in vegetation condition.

Among the six supervised algorithms evaluated on spatially separated test blocks, Gradient Tree Boosting (GTB) and Random Forest (RF) achieved the highest mean accuracy (0.856 and 0.852). RF accuracy ranged from 0.768 to 0.924 and $\kappa$ from 0.167 to 0.787 among sites; the higher-HI class was identified with a recall of 0.965–0.999, whereas lower-HI recall ranged from 0.121 to 0.745, so that at Site B the model added little beyond the majority class. RF was selected for mapping because it produced the fewest false negatives of all models at every site (1,438 in total, compared with 2,718 for GTB). Feature-importance and SHAP analyses showed that classification was dominated by soil texture, climate, elevation and land cover, whereas vegetation indices, biophysical parameters, CSTV and rate-of-change metrics contributed 13.8–17.4% of the total SHAP importance.

At the constituent level, RF classified the normalized HQ of the 34 metal–site combinations with HQ ≥ 1 with an accuracy of 0.775–0.989 (mean 0.868), $\kappa$ of 0.281–0.969 and higher-HQ recall of 0.804–1.000, as well as or better than the composite HI, including at Site B ($\kappa$ 0.391–0.969 vs. 0.167 for HI).

Cross-site testing identified a fundamental limitation in geographic generalization. In leave-one-site-out tests, RF predicted almost all pixels of the held-out site as higher HI (accuracy 0.734–0.749, equal to or below the majority-class share; $\kappa$ −0.020 to 0.010), and no classifier exceeded $\kappa$ = 0.143. The predictor–HI relationships learned within sites are therefore site specific and, because they rely largely on coarse environmental covariates, partly reflect the spatial structure of the interpolated HI rather than a transferable vegetation response. Broader multi-site training, site-specific calibration, finer-resolution covariates and domain-adaptation approaches will be required before the framework can be applied to sites without monitoring data.

Taken together, the results lead to three conclusions: (i) Near the monitoring wells, vegetation responds to the leached metals. Higher metal HQ coincided with lower vegetation index values in 24 of 33 metal–site combinations, including 14 of 16 for Mn, As, Co and B. Sites A, B and D show this consistently. At Site C, the index values near the wells reflect land cover (water and built surfaces) rather than metal stress; (ii) Within a site, the HI pattern can be reproduced, but mainly from environmental features including soil, climatic, topographic and land-cover covariates. Removing the vegetation indices did not reduce performance, and at Site B the distance to the CCR unit alone outperformed all remotely sensed predictor sets. The maps therefore interpolate the monitored hazard rather than detect it independently; (iii) Across sites, the learned relationships do not transfer ($\kappa = -0.020$ to 0.010 for RF). Satellite data can therefore support groundwater monitoring in two ways: by showing where vegetation near CCR units responds to leached metals, and by interpolating the monitored hazard within a site. They cannot yet replace monitoring wells or map leached-metal hazard at sites without monitoring well data.

The primary contributions of this study are fivefold: (i) development of a multi-constituent HQ–HI framework that converts groundwater monitoring data into spatial reference information representing cumulative relative metals burden; (ii) integration of spectral, vegetation-index, biophysical, CSTV, multi-year rate-of-change, and environmental feature predictors into a unified machine-learning framework; (iii) demonstration that supervised ML models such as RF can effectively classify and spatially map vegetation–environment conditions associated with lower and higher composite HI within individual CCR sites; (iv) demonstration at Site A that the RF framework can be extended from composite site-level hazard classification to individual constituent-specific classification and (v) quantitative evaluation of geographic transferability through leave-one-site-out testing, demonstrating both the potential and current limitations of transferring remotely sensed HI associations among environmentally heterogeneous sites.

Overall, the study establishes a framework for using groundwater observations together with satellite-derived vegetation dynamics and environmental information to provide spatial screening beyond discrete monitoring wells. The approach should not be interpreted as direct remote detection of groundwater metals. Instead, it identifies remotely observable vegetation–environment patterns associated with spatial variation in composite groundwater HI. With additional field validation, hydrogeological integration, and improved cross-site generalization, the framework could complement conventional groundwater monitoring by supporting targeted sampling, long-term CCR surveillance, and remediation monitoring**.**

## 6 Credit Authorship Contribution Statement

Salman Karim: Methodology, Investigation, Formal Analysis, Visualization, Software, Writing – original draft.

Bahareh Nojabaei: Conceptualization, Supervision, Writing – review & editing.

Hosein Foroutan: Conceptualization, Writing – review & editing.

## 7 Declaration of Competing Interests

The authors declare that there is no known competing interests or relationships that could influence the research results reported in this paper.

## 8 Declaration of Ethical Responsibilities

All authors have read, understood, and have complied as applicable with the statement on "Ethical responsibilities of Authors" as found in the Instructions for Authors.

## 9 Funding Declaration

This work was supported by the U.S. Department of Energy, National Energy Technology Laboratory, under Award Number DEFE0032184.

## 10 Data Availability Statement

The multisource remote sensing dataset used in this study is openly available on Zenodo. It contains the Sentinel-2 vegetation indices, biophysical parameters, spatio-temporal metrics, and the environmental and land-cover predictors for the four CCR sites (2019–2025) ((Karim 2026), https://doi.org/10.5281/zenodo.23823223). The code for feature extraction in Google Earth Engine, Hazard Index labelling, model training, spatial evaluation and figure generation is available at https://github.com/salmansust/hazard_rs_ml.

## 12 Supplementary Material

**Table S1.** Spearman's ρ between the well HQ of each metal exceeding its reference threshold and the summer value of nine vegetation indices within 30 m of the monitoring wells at Sites A–D. Max HQ: highest well HQ at the site; Mean: mean ρ of the nine indices. Negative values indicate lower index values at wells with higher HQ.

| Metal | Max HQ | NDVI | SAVI | CIRE | NDRE1 | NDRE2 | MTCI | LAI | LCC | CCC | Mean |
|---|---|---|---|---|---|---|---|---|---|---|---|
| *Site A* | | | | | | | | | | | |
| Mn | 15.1 | −0.15 | −0.16 | −0.11 | −0.09 | −0.08 | −0.05 | −0.13 | −0.64 | −0.23 | −0.18 |
| As | 3.7 | −0.04 | −0.07 | −0.04 | −0.03 | −0.01 | −0.03 | −0.07 | −0.02 | −0.01 | −0.04 |
| Co | 2.2 | −0.06 | −0.07 | −0.09 | −0.10 | −0.07 | −0.07 | −0.10 | −0.13 | −0.04 | −0.08 |
| B | 1.2 | −0.59 | −0.59 | −0.57 | −0.58 | −0.55 | −0.52 | −0.55 | −0.76 | −0.60 | −0.59 |
| Mo | 8.0 | −0.57 | −0.54 | −0.53 | −0.54 | −0.54 | −0.50 | −0.49 | −0.34 | −0.52 | −0.51 |
| Pb | 3.8 | −0.03 | −0.03 | −0.09 | −0.02 | −0.05 | 0.00 | −0.03 | 0.01 | 0.01 | −0.03 |
| Sr | 4.8 | −0.21 | −0.15 | −0.18 | −0.22 | −0.25 | −0.29 | −0.16 | 0.13 | −0.13 | −0.16 |
| *Site B* | | | | | | | | | | | |
| Mn | 33.0 | −0.53 | −0.53 | −0.48 | −0.49 | −0.49 | −0.57 | −0.54 | −0.29 | −0.48 | −0.49 |
| As | 23.9 | −0.33 | −0.34 | −0.30 | −0.30 | −0.31 | −0.32 | −0.40 | −0.06 | −0.23 | −0.29 |
| Co | 6.2 | −0.62 | −0.70 | −0.58 | −0.59 | −0.59 | −0.57 | −0.65 | −0.47 | −0.59 | −0.60 |
| B | 1.2 | −0.52 | −0.47 | −0.51 | −0.50 | −0.51 | −0.57 | −0.57 | −0.21 | −0.41 | −0.47 |
| Mo | 3.5 | −0.14 | −0.03 | −0.14 | −0.12 | −0.14 | −0.11 | −0.12 | 0.17 | −0.01 | −0.07 |
| Pb | 6.7 | 0.10 | 0.12 | 0.08 | 0.08 | 0.07 | 0.08 | 0.12 | 0.17 | 0.08 | 0.10 |
| Li | 7.0 | −0.39 | −0.31 | −0.38 | −0.37 | −0.37 | −0.33 | −0.28 | −0.27 | −0.36 | −0.34 |
| Hg | 5.0 | −0.08 | −0.12 | −0.12 | −0.11 | −0.10 | −0.08 | −0.07 | −0.15 | −0.18 | −0.11 |
| *Site C* | | | | | | | | | | | |
| Mn | 41.0 | −0.09 | −0.11 | −0.20 | −0.11 | −0.15 | −0.38 | −0.12 | −0.07 | −0.20 | −0.16 |
| As | 25.0 | −0.18 | −0.20 | −0.18 | −0.18 | −0.19 | −0.28 | −0.19 | −0.20 | −0.20 | −0.20 |
| Co | 43.3 | 0.39 | 0.44 | 0.42 | 0.38 | 0.40 | 0.31 | 0.43 | 0.50 | 0.40 | 0.41 |
| B | 14.4 | 0.22 | 0.22 | 0.28 | 0.22 | 0.22 | 0.20 | 0.20 | 0.43 | 0.31 | 0.26 |
| Mo | 2.6 | 0.16 | 0.16 | 0.12 | 0.14 | 0.15 | 0.15 | 0.16 | 0.15 | 0.14 | 0.15 |
| Pb | 3.2 | −0.15 | −0.15 | −0.06 | −0.11 | −0.11 | −0.05 | −0.15 | −0.08 | −0.07 | −0.10 |
| Ra | 2.9 | 0.38 | 0.36 | 0.40 | 0.36 | 0.36 | 0.28 | 0.34 | 0.50 | 0.42 | 0.38 |
| Li | 5.6 | 0.43 | 0.50 | 0.42 | 0.42 | 0.42 | 0.35 | 0.48 | 0.47 | 0.40 | 0.43 |
| Be | 8.5 | 0.02 | 0.01 | 0.04 | 0.05 | 0.04 | 0.01 | 0.01 | 0.14 | 0.06 | 0.04 |
| Cr | 1.9 | 0.32 | 0.30 | 0.34 | 0.33 | 0.34 | 0.26 | 0.30 | 0.38 | 0.34 | 0.32 |
| Tl | 1.0 | 0.23 | 0.20 | 0.22 | 0.20 | 0.21 | 0.10 | 0.18 | 0.39 | 0.25 | 0.22 |
| *Site D* | | | | | | | | | | | |
| Mn | 11.0 | −0.33 | −0.40 | −0.41 | −0.46 | −0.45 | −0.46 | −0.43 | −0.38 | −0.41 | −0.41 |
| As | 4.9 | −0.05 | −0.08 | −0.04 | −0.06 | −0.05 | −0.04 | −0.10 | −0.03 | −0.07 | −0.06 |
| Co | 9.6 | −0.50 | −0.43 | −0.55 | −0.51 | −0.51 | −0.42 | −0.41 | −0.53 | −0.59 | −0.49 |
| B | 1.7 | −0.31 | −0.38 | −0.36 | −0.31 | −0.31 | −0.29 | −0.44 | −0.33 | −0.39 | −0.35 |
| Ra | 1.8 | −0.27 | −0.28 | −0.20 | −0.20 | −0.20 | −0.09 | −0.29 | −0.24 | −0.21 | −0.22 |
| Ni | 2.1 | −0.09 | −0.13 | −0.09 | −0.11 | −0.11 | −0.11 | −0.16 | −0.07 | −0.08 | −0.11 |
| Cd | 2.3 | 0.06 | −0.01 | −0.05 | −0.03 | −0.02 | −0.06 | −0.02 | −0.09 | −0.09 | −0.03 |

**Figure S0:** Hazard Index (HI) at each monitoring well against the nine vegetation indices and biophysical parameters (NDVI, SAVI, CIRE, NDRE1, NDRE2, MTCI, LAI, LCC and CCC) averaged within 60 m of the upgradient and downgradient wells at Sites A–D;

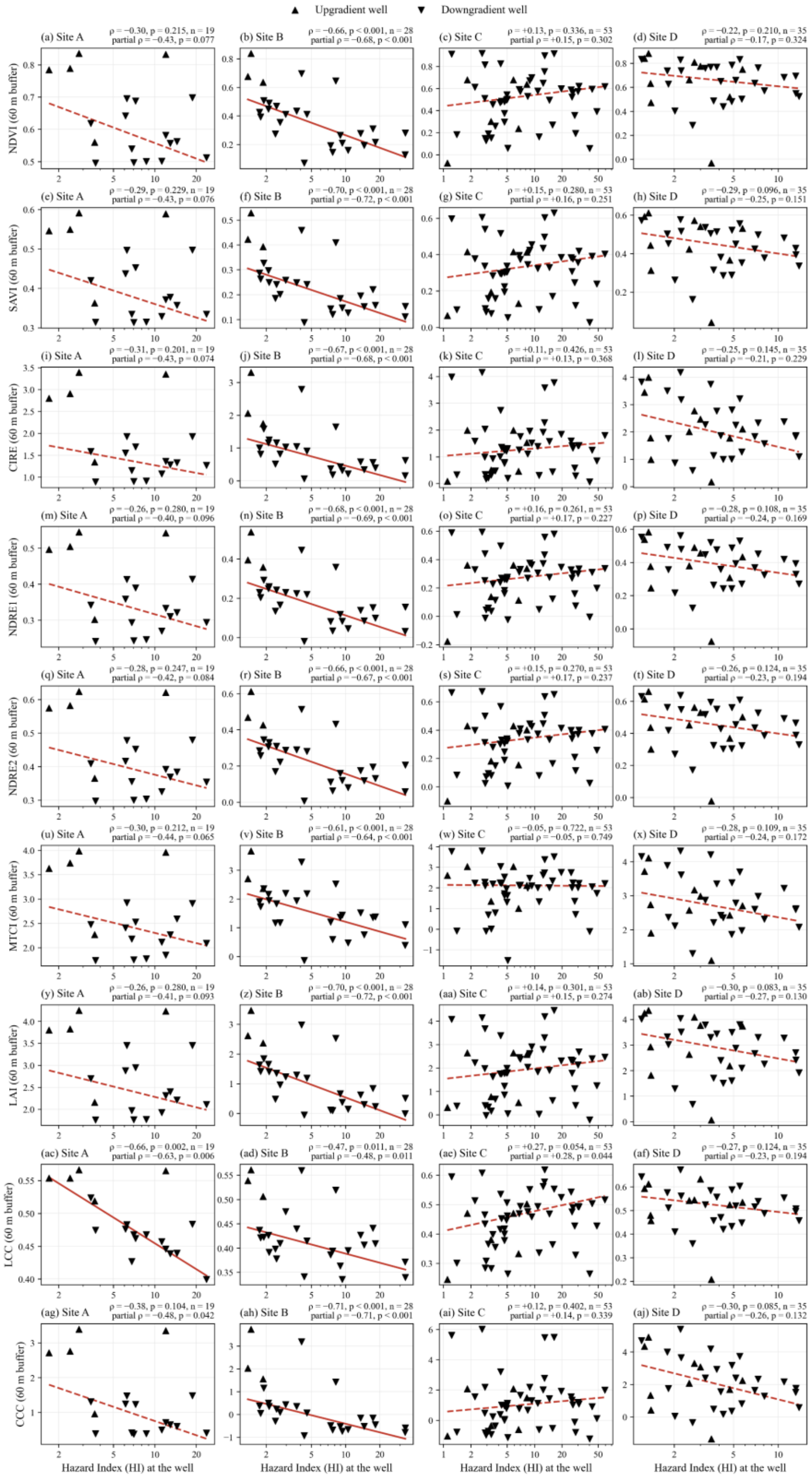

**Figure S1.** Constituent-level Random Forest classification at Site B (rows: metals with HQ ≥ 1; columns: HQ, normalized HQ, RF classification, RF probability, classification outcome).

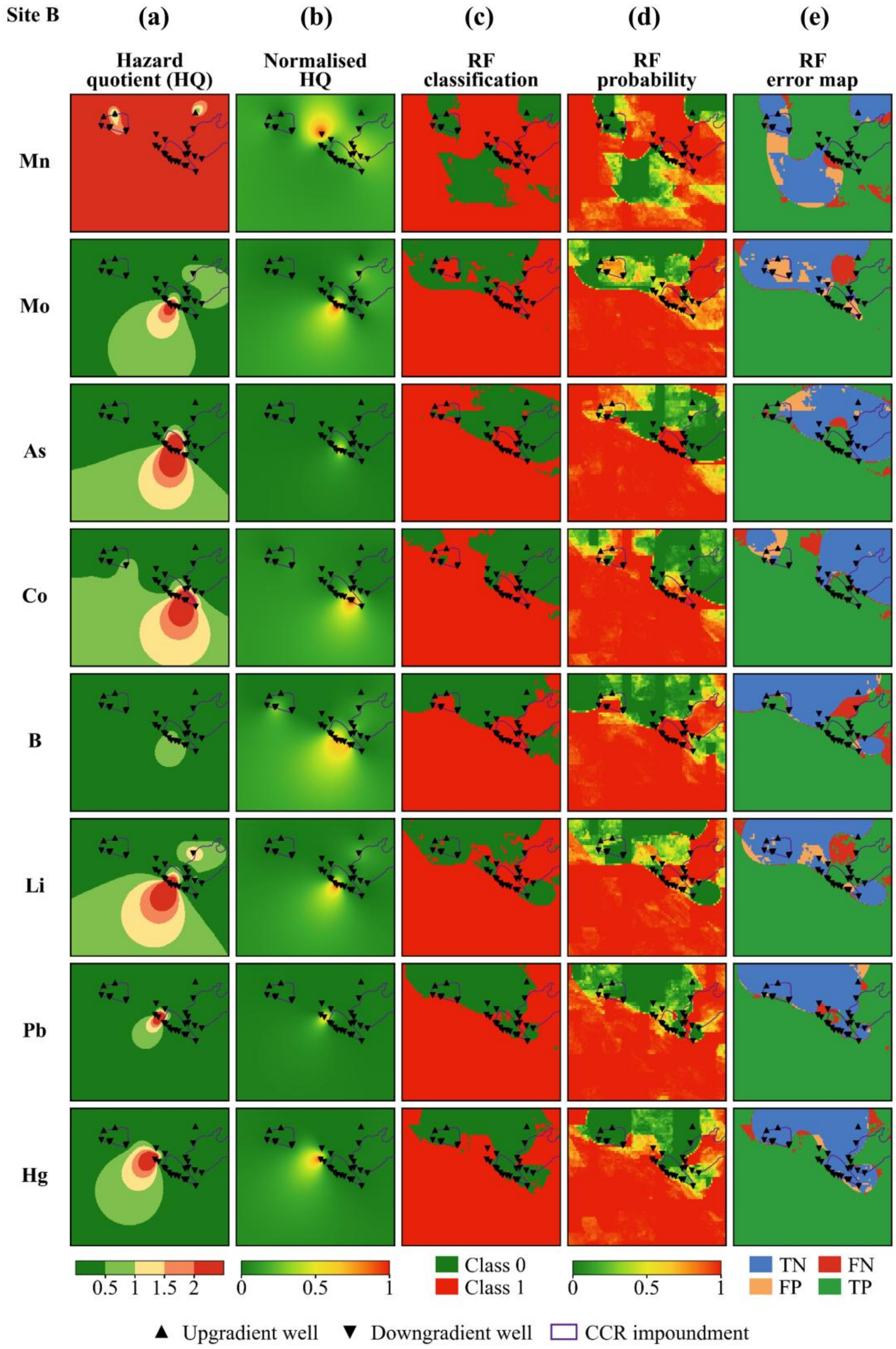

**Figure S2.** As Figure S1, Site C.

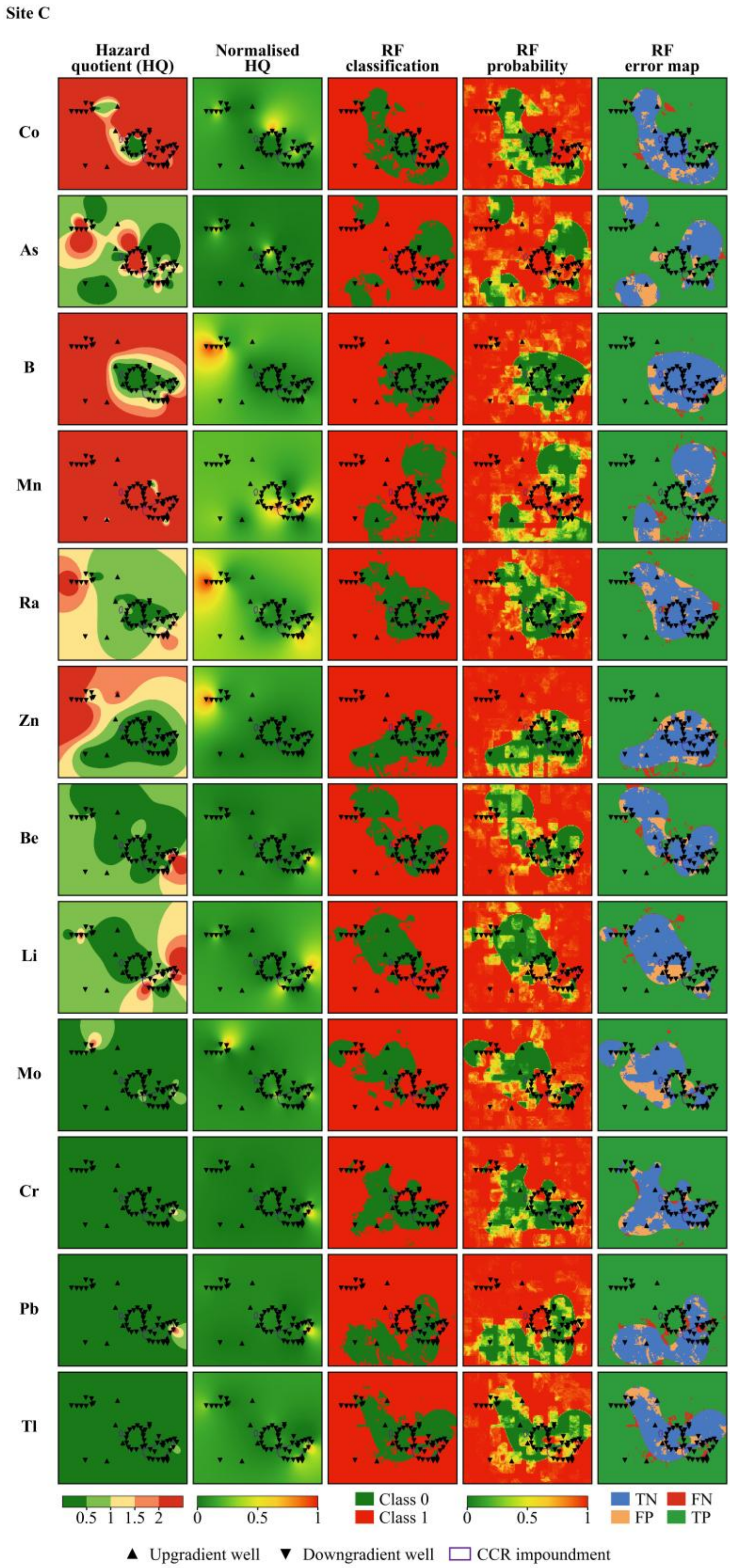

**Figure S3.** As Figure S1, Site D.

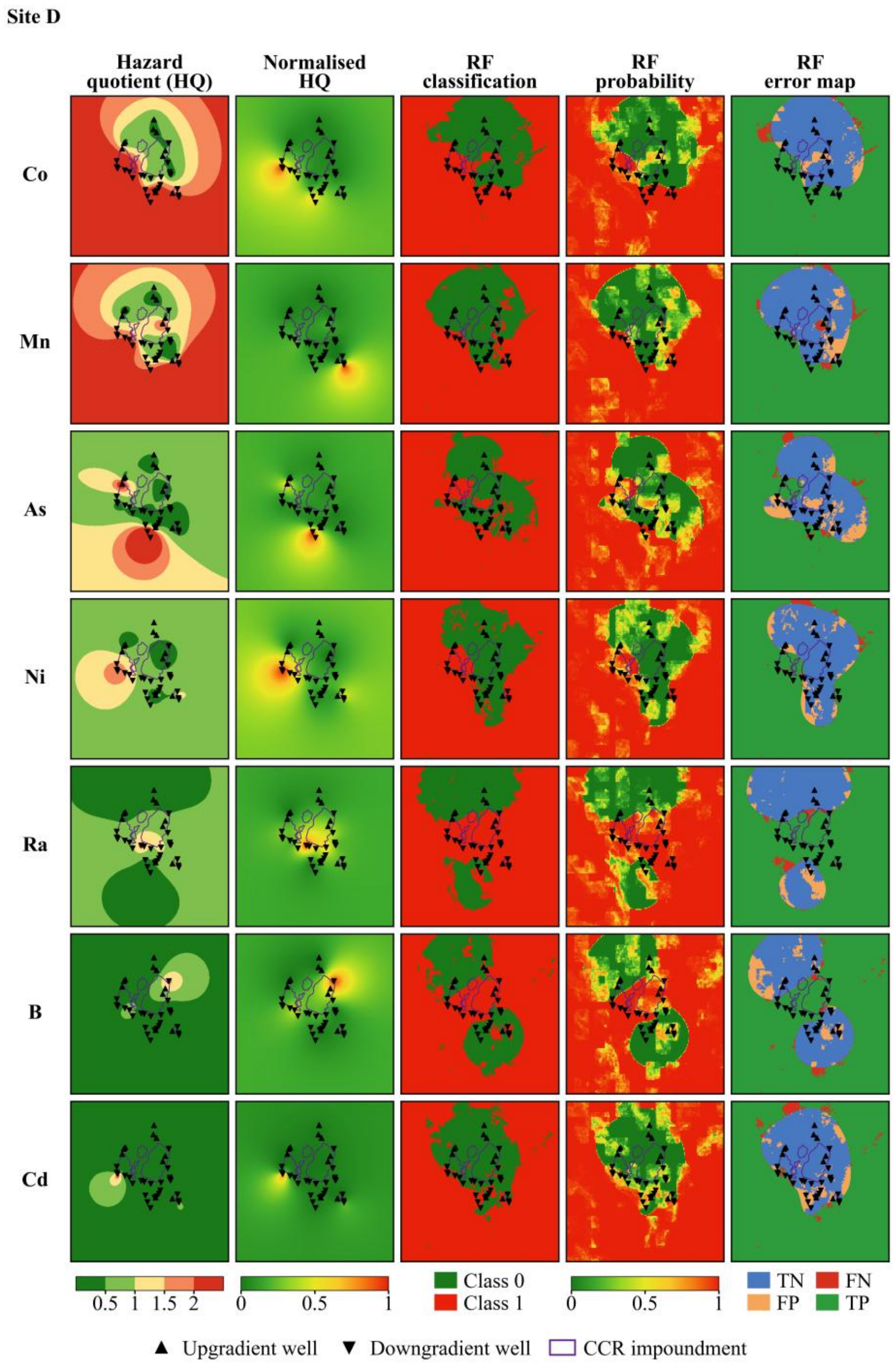

**Figure S4.** Mean absolute SHAP value of the Random Forest summed by feature group at Sites A–D.

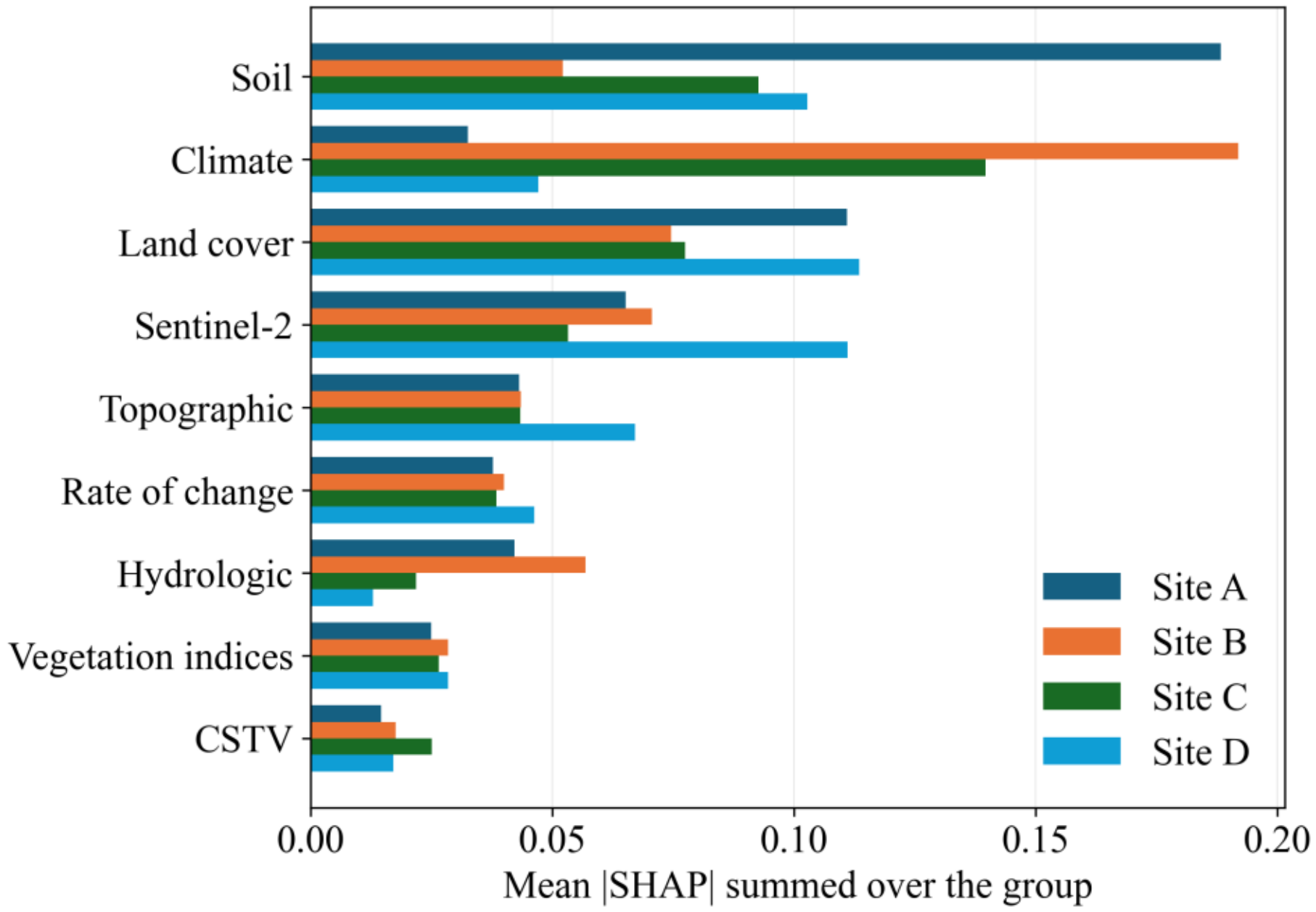

**Figure S5.** Leave-one-site-out performance of the six classifiers at each held-out site.

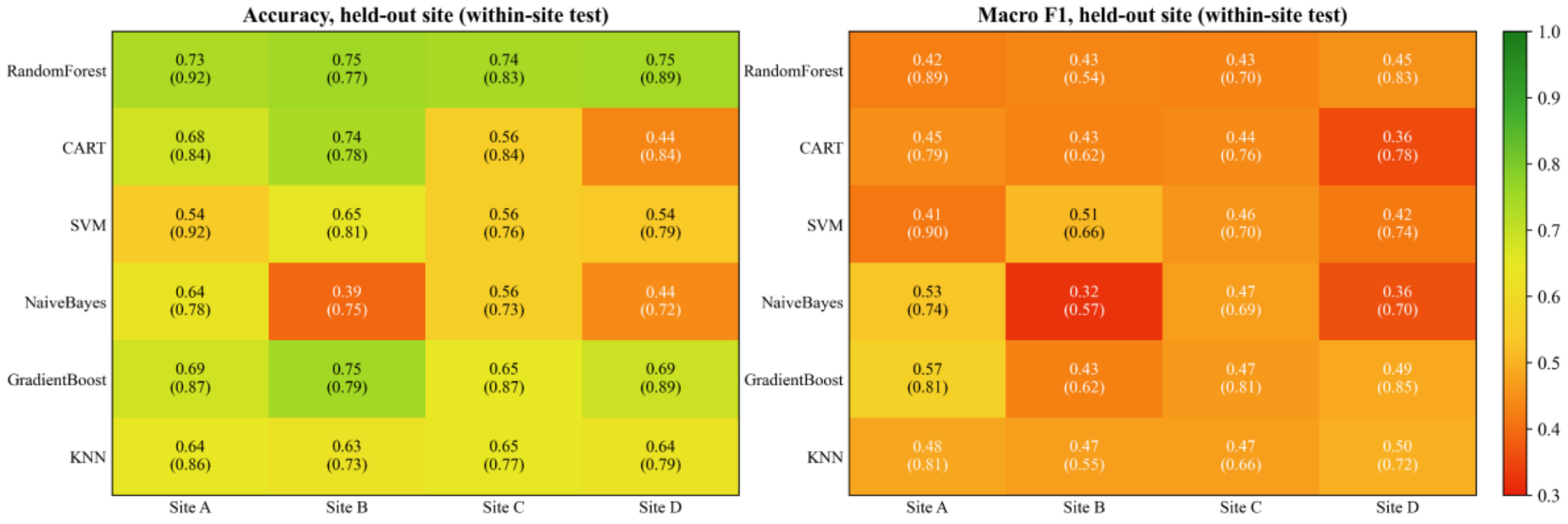

**Figure S6.** Leave-one-site-out Random Forest maps of the held-out sites.

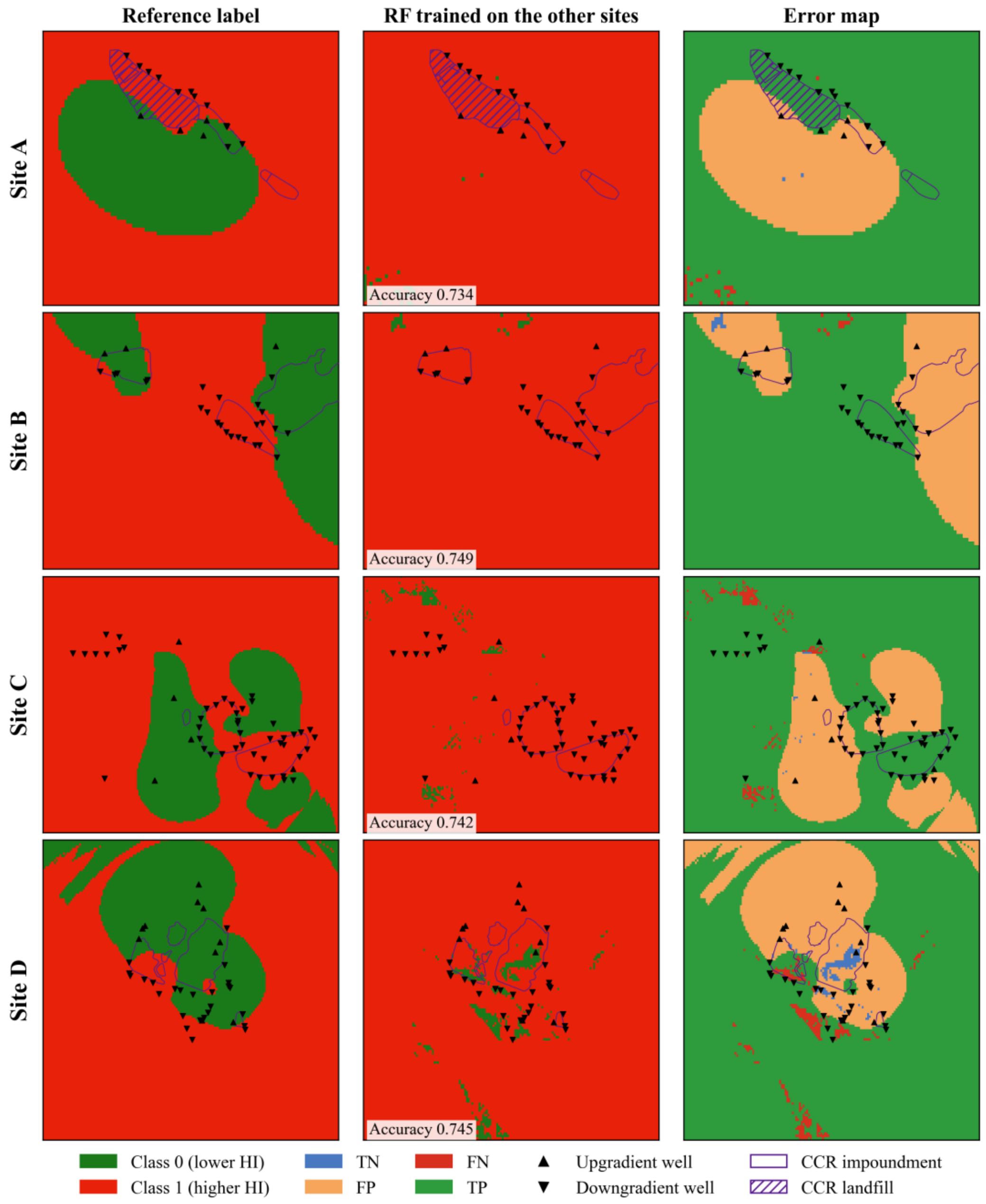

**Figure S7.** Vegetation and biophysical parameter indices at each site.

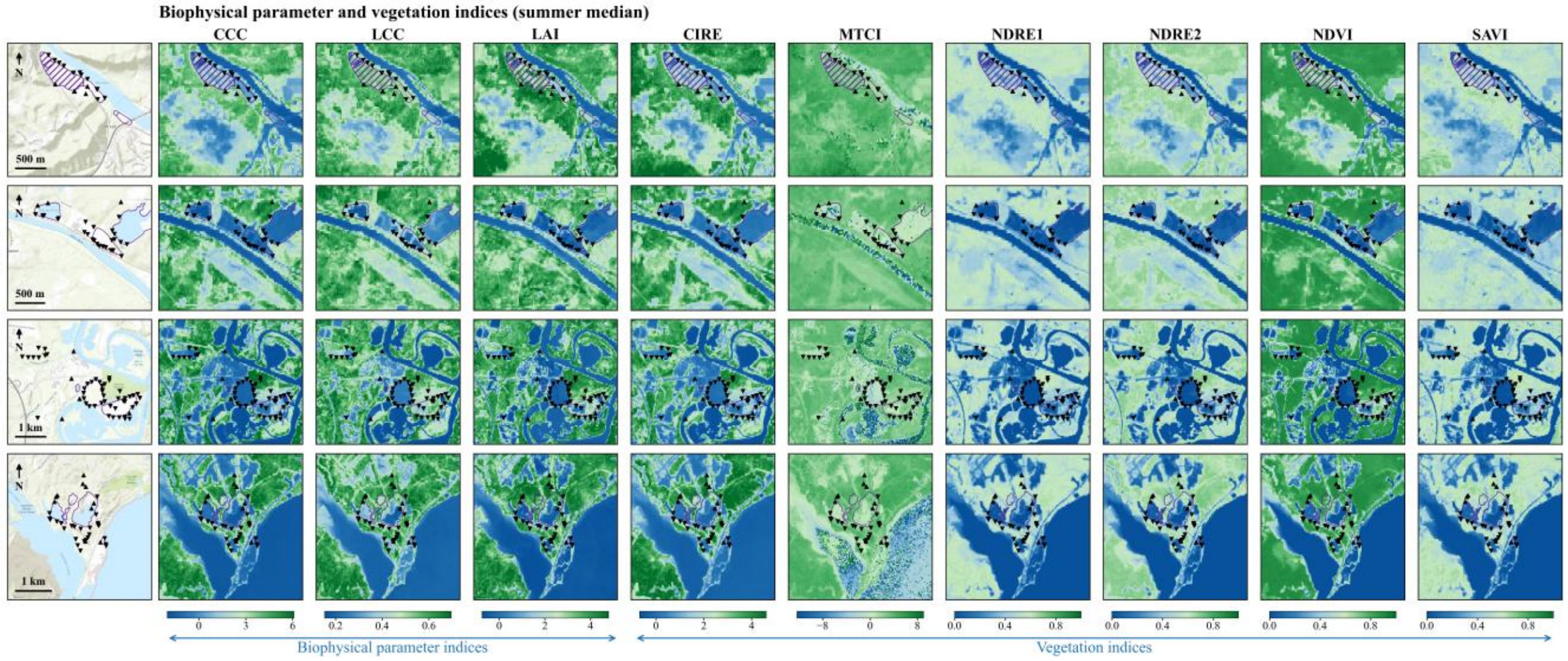

**Figure S8.** Rate of Change of vegetation and biophysical parameter indices at each site.

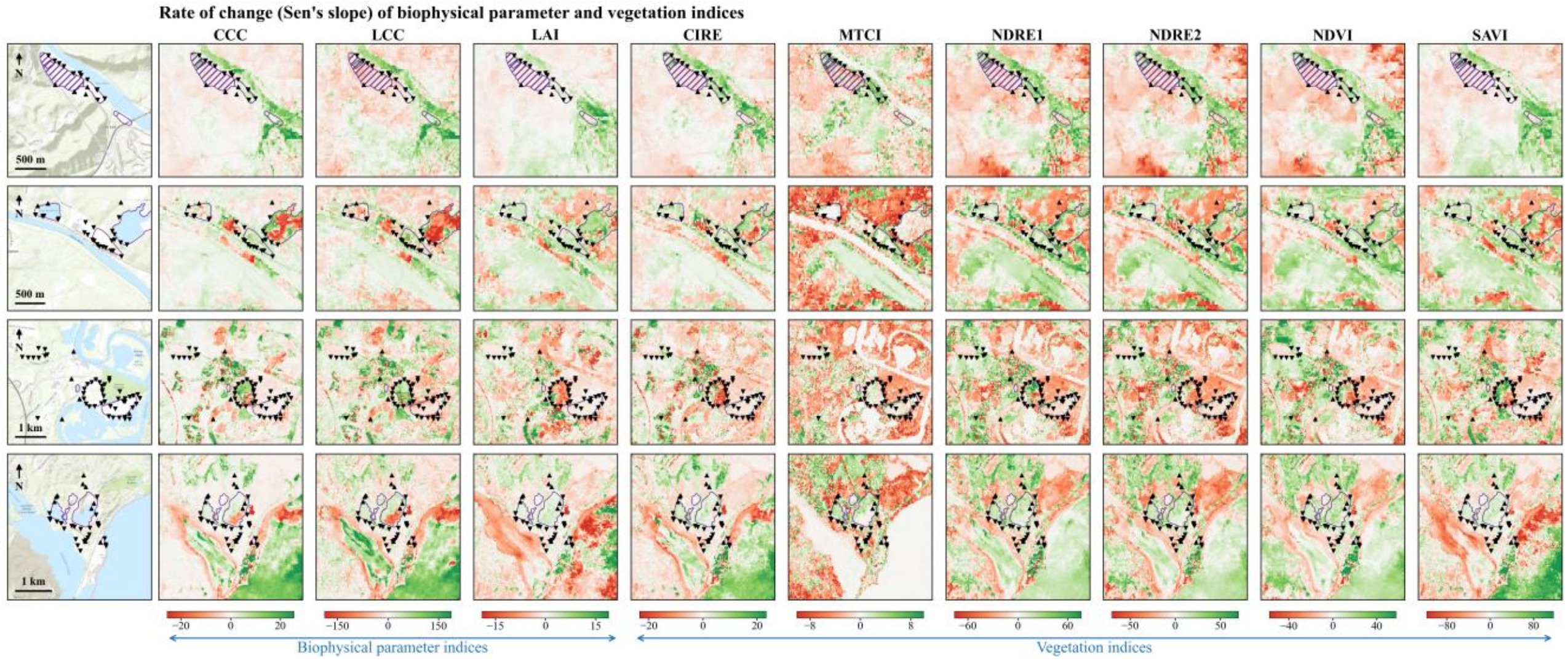

**Figure S9.** Sentinel-2 surface reflectance at each site.

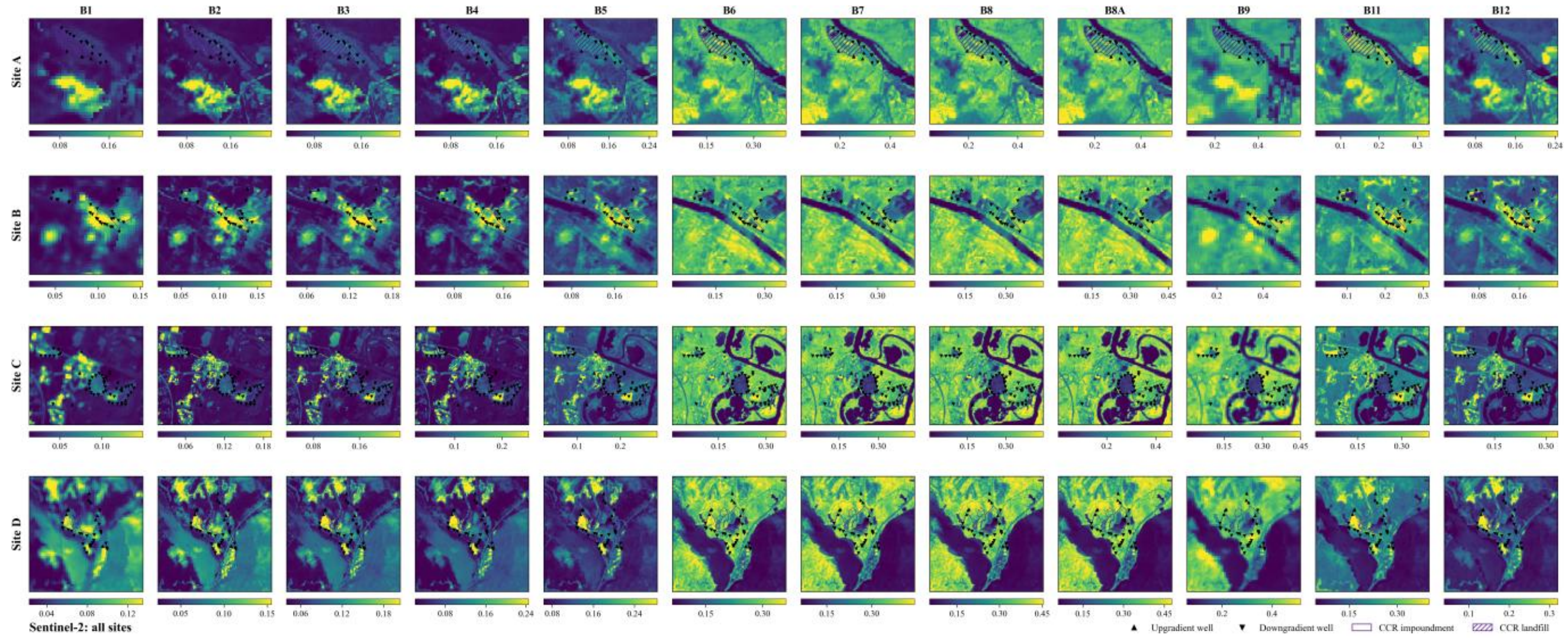

**Figure S9.** Topographic features at each site.

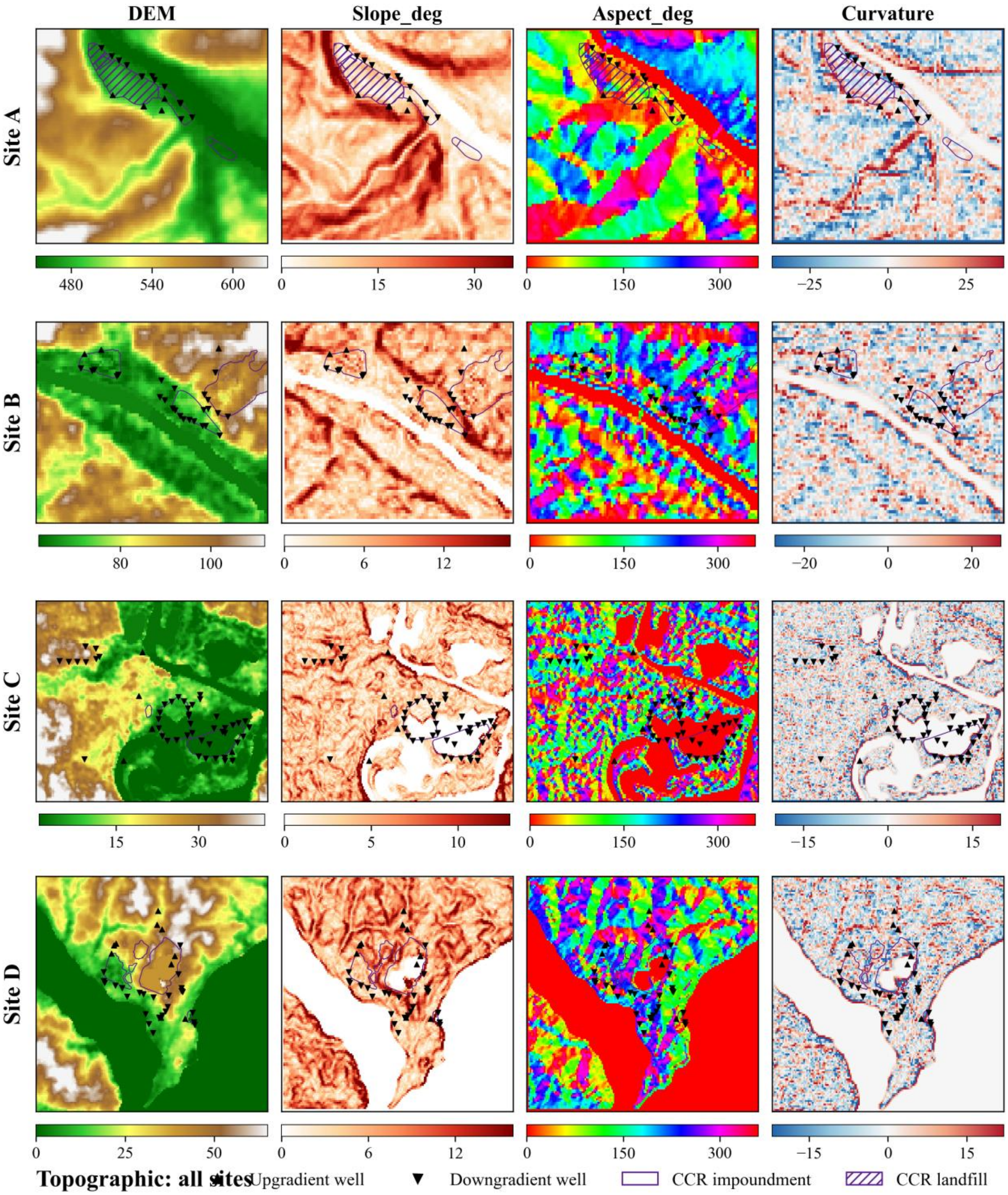

**Figure S10.** Soil features at each site.

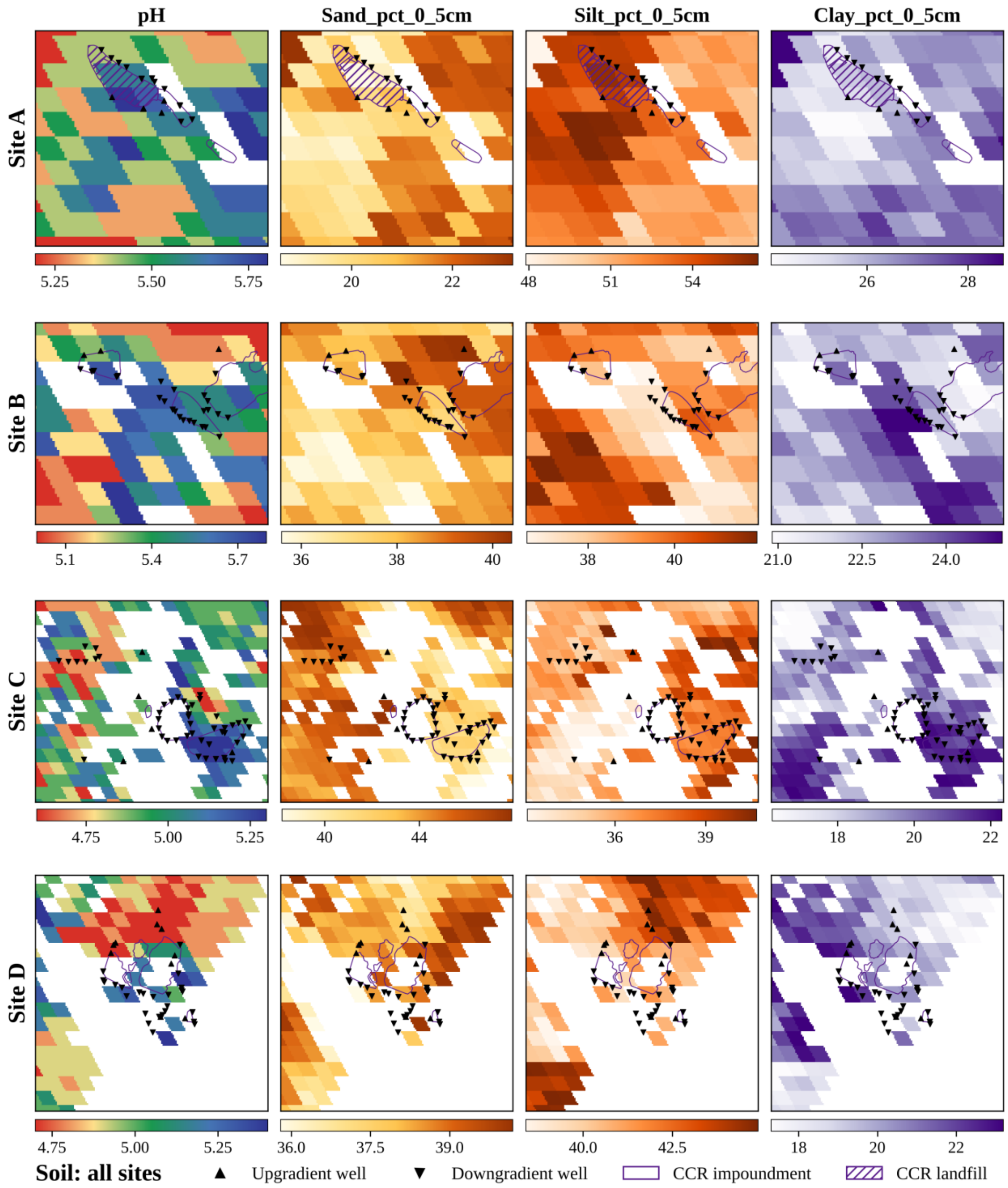

**Figure S11**. Climate features at each site.

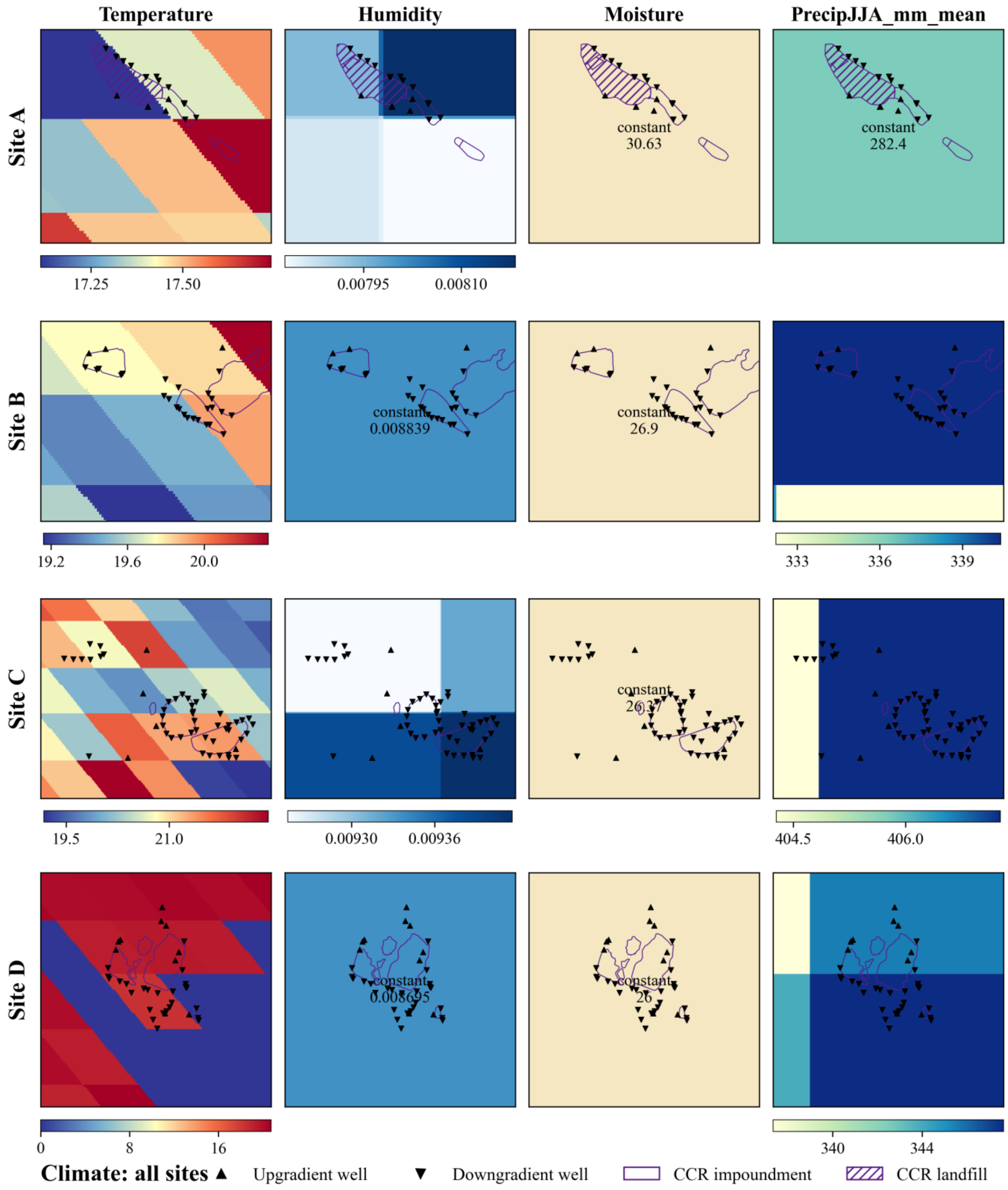

**Figure S12.** Hydrologic features at each site.

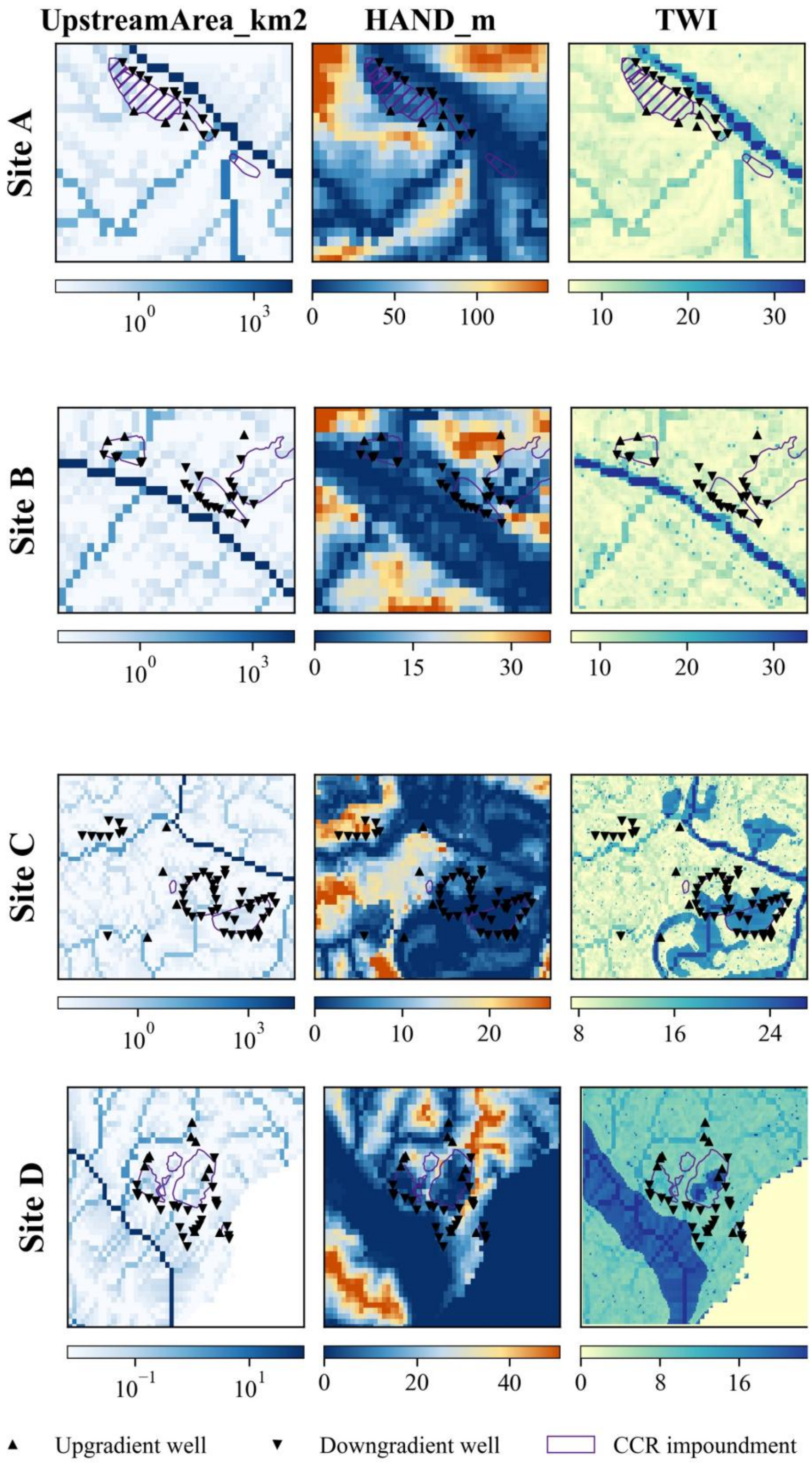

**Figure S13.** Site A monitoring wells with HQ and HI.

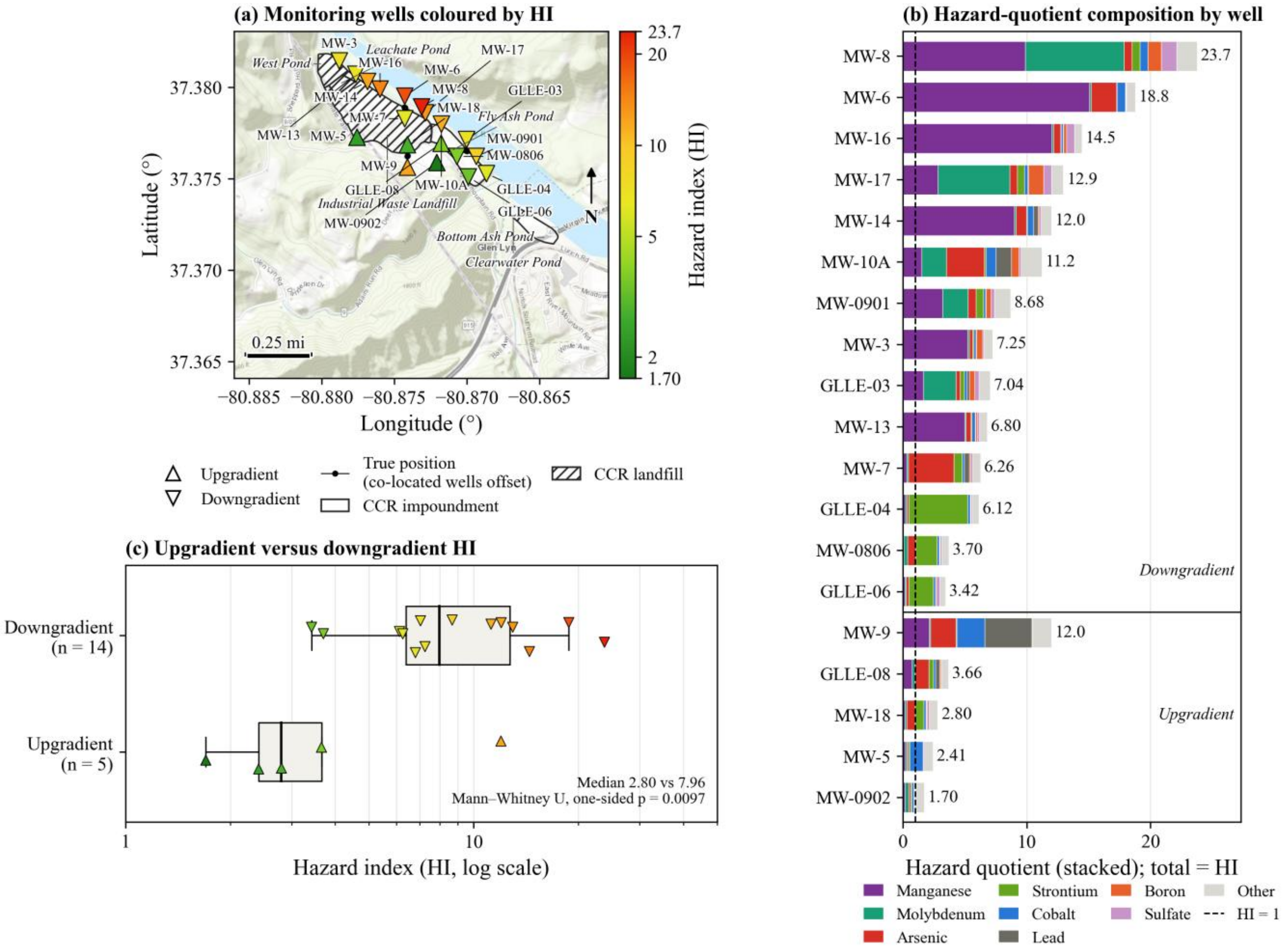

**Figure S14.** Site B monitoring wells with HQ and HI.

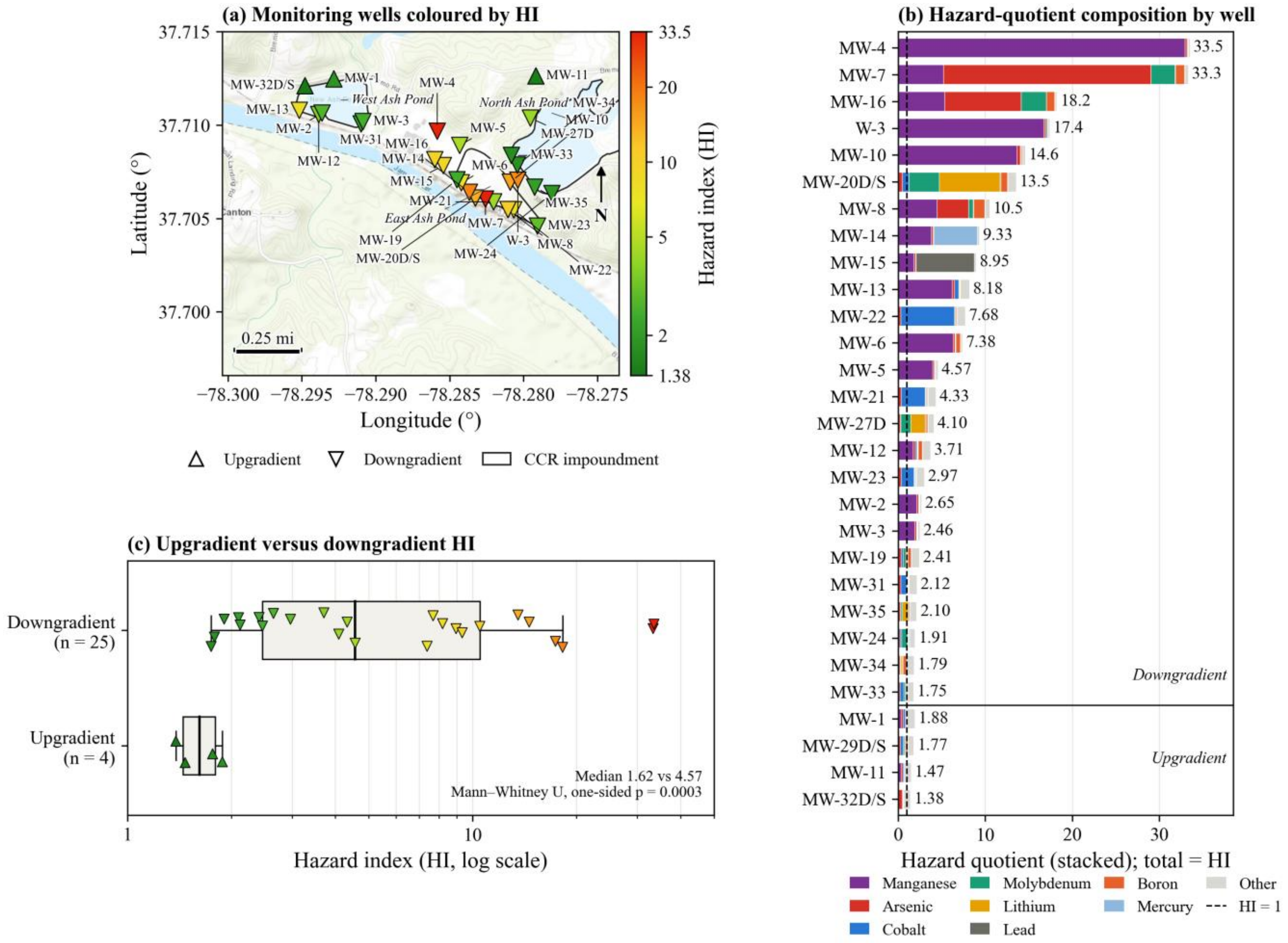

**Figure S15.** Site C monitoring wells with HQ and HI.

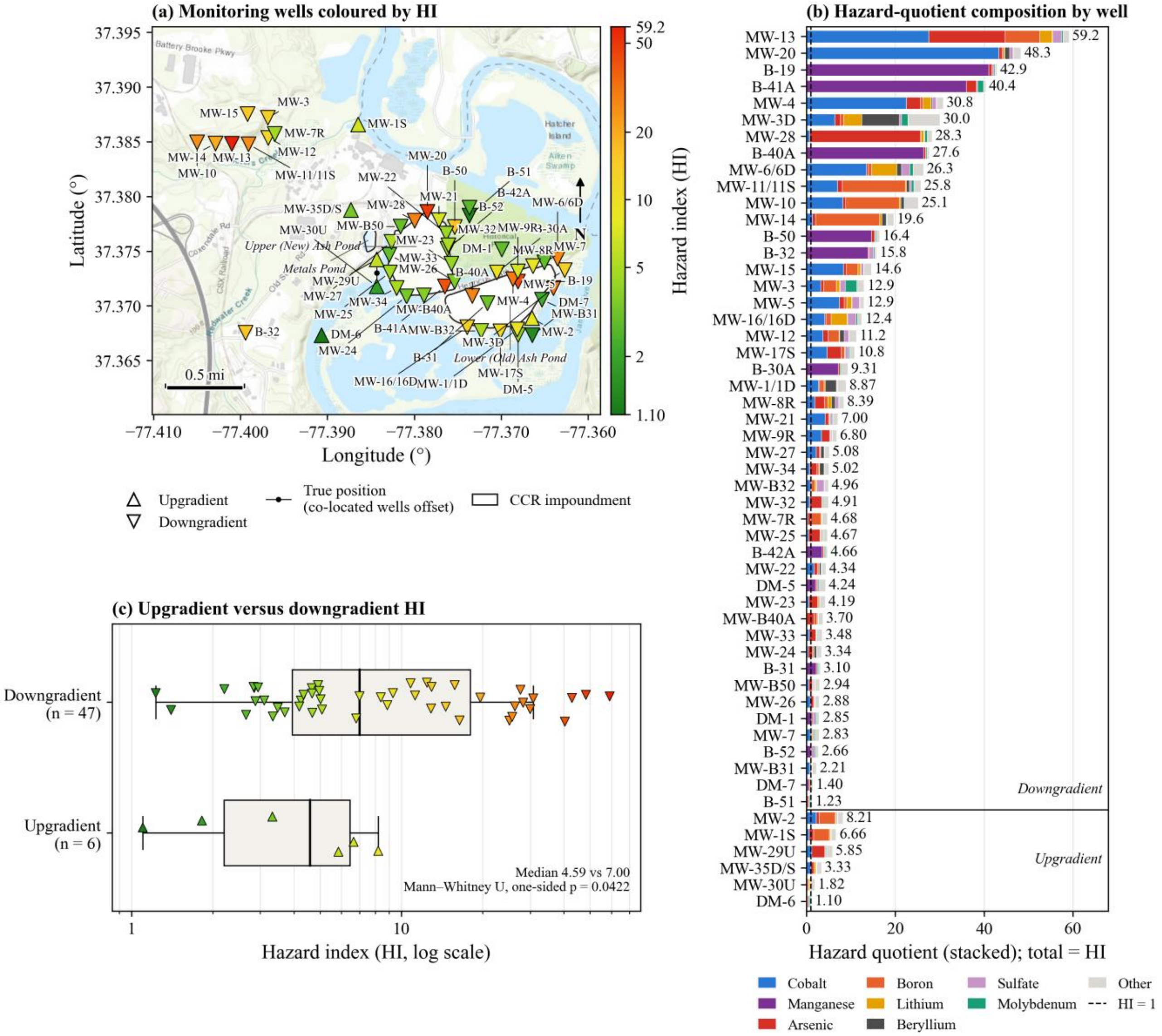

**Figure S16.** Site D monitoring wells with HQ and HI.

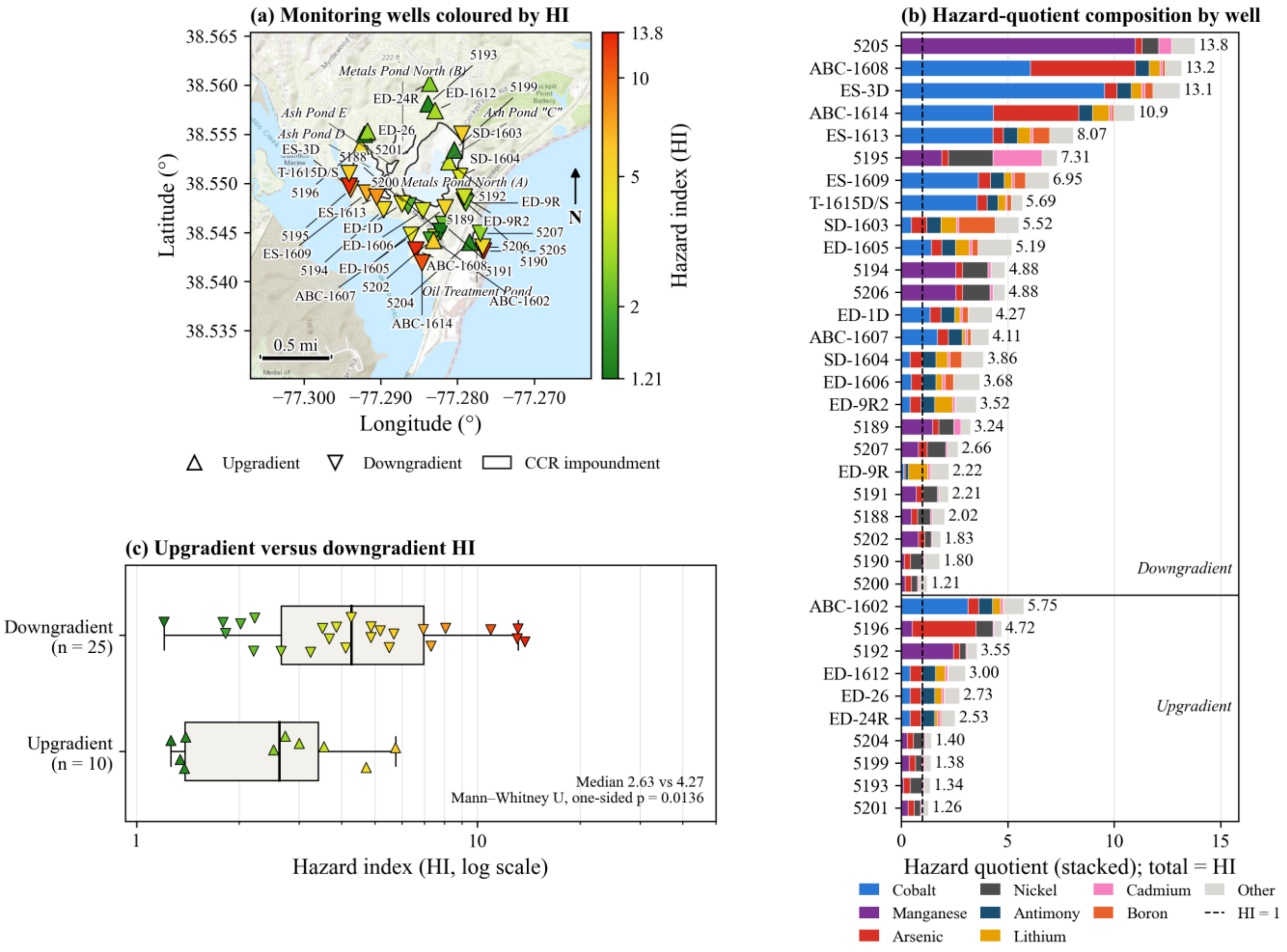